\documentclass{article}
\usepackage{iclr2026_conference,times}
\usepackage{hyperref}
\usepackage{url}
\usepackage{natbib}
\setcitestyle{numbers,sort&compress}
\usepackage{amsmath,amssymb,mathtools}
\usepackage{algorithm}
\usepackage{algpseudocode}
\usepackage{bm}
\usepackage{multirow}
\usepackage{graphicx}
\usepackage{tabularx}
\usepackage{float}
\usepackage{afterpage}
\usepackage{caption}
\usepackage[capitalise]{cleveref}
\Crefname{lstlisting}{Listing}{Listings}
\usepackage{booktabs}
\usepackage{threeparttable}
\usepackage{pifont}
\usepackage[margin=1in]{geometry}
\usepackage{array}
\usepackage[table]{xcolor}
\usepackage{adjustbox}
\usepackage[utf8]{inputenc}
\usepackage[T1]{fontenc}
\usepackage{textcomp}
\usepackage{enumitem}
\usepackage{listings}
\definecolor{codebg}{RGB}{248,248,248}
\definecolor{codegreen}{RGB}{0,128,0}
\definecolor{codegray}{RGB}{128,128,128}
\definecolor{codepurple}{RGB}{128,0,128}
\lstdefinestyle{pythonstyle}{
  language=Python,
  backgroundcolor=\color{codebg},
  basicstyle=\ttfamily\footnotesize,
  keywordstyle=\color{codepurple}\bfseries,
  stringstyle=\color{codegreen},
  commentstyle=\color{codegray}\itshape,
  numbers=left,
  numberstyle=\tiny\color{codegray},
  numbersep=5pt,
  breaklines=true,
  frame=single,
  rulecolor=\color{black!20},
  xleftmargin=1.5em,
  framexleftmargin=1.5em,
  columns=fullflexible,
  keepspaces=true,
  showstringspaces=false,
  tabsize=4,
  aboveskip=0.8em,
  belowskip=0.5em,
}
\newcommand{\unibayes}[1]{Bayes@$\!#1$}
\newcommand{\bayes}[1]{Bayes@$\!#1$}
\newcommand{\pass}[1]{Pass@$\!#1$}
\newcommand{\avg}[1]{avg@$\!#1$}
\newcommand{\aimefour}{AIME'24}
\newcommand{\aimefive}{AIME'25}
\newcommand{\aimeboth}{AIME'24/'25}
\newcommand{\hmmt}{HMMT'25}
\newcommand{\brumo}{BrUMO'25}
\newcommand{\yes}{\textcolor{mygreen}{\ding{51}}}
\newcommand{\no}{\textcolor{myred}{\ding{55}}}

\definecolor{mygreen}{rgb}{0.0, 0.5, 0.0}
\definecolor{myred}{rgb}{0.7, 0.0, 0.0}
\definecolor{myorange}{rgb}{0.8, 0.4, 0.0}
\definecolor{legendblue}{RGB}{31,119,180}
\definecolor{legendorange}{RGB}{255,127,14}
\definecolor{legendred}{RGB}{214,39,40}
\definecolor{legendbrown}{RGB}{140,86,75}
\definecolor{rank1}{rgb}{0.80, 1.00, 0.80}
\definecolor{rank2}{rgb}{0.88, 1.00, 0.82}
\definecolor{rank3}{rgb}{0.94, 1.00, 0.86}
\definecolor{rank4}{rgb}{1.00, 1.00, 0.88}
\definecolor{rank5}{rgb}{1.00, 0.96, 0.82}
\definecolor{rank6}{rgb}{1.00, 0.92, 0.76}
\definecolor{rank7}{rgb}{1.00, 0.88, 0.72}
\definecolor{rank8}{rgb}{1.00, 0.82, 0.70}
\definecolor{rank9}{rgb}{0.98, 0.74, 0.72}
\definecolor{rank10}{rgb}{0.96, 0.64, 0.72}
\definecolor{rank11}{rgb}{0.90, 0.55, 0.70}
\definecolor{rank12}{rgb}{0.84, 0.46, 0.68}
\definecolor{rank13}{rgb}{0.78, 0.37, 0.66}
\definecolor{rank14}{rgb}{0.72, 0.28, 0.64}
\definecolor{rank15}{rgb}{0.66, 0.19, 0.62}
\definecolor{rank16}{rgb}{0.60, 0.10, 0.60}
\definecolor{rank17}{rgb}{0.54, 0.01, 0.58}
\definecolor{rank18}{rgb}{0.48, 0.00, 0.56}
\definecolor{rank19}{rgb}{0.42, 0.00, 0.54}
\definecolor{rank20}{rgb}{0.36, 0.00, 0.52}
\newcommand{\rankcell}[2]{%
  \ifnum#1=1 \cellcolor{rank1}#2%
  \else\ifnum#1=2 \cellcolor{rank2}#2%
  \else\ifnum#1=3 \cellcolor{rank3}#2%
  \else\ifnum#1=4 \cellcolor{rank4}#2%
  \else\ifnum#1=5 \cellcolor{rank5}#2%
  \else\ifnum#1=6 \cellcolor{rank6}#2%
  \else\ifnum#1=7 \cellcolor{rank7}#2%
  \else\ifnum#1=8 \cellcolor{rank8}#2%
  \else\ifnum#1=9 \cellcolor{rank9}#2%
  \else\ifnum#1=10 \cellcolor{rank10}#2%
  \else\ifnum#1=11 \cellcolor{rank11}#2%
  \else\ifnum#1=12 \cellcolor{rank12}#2%
  \else\ifnum#1=13 \cellcolor{rank13}#2%
  \else\ifnum#1=14 \cellcolor{rank14}#2%
  \else\ifnum#1=15 \cellcolor{rank15}#2%
  \else\ifnum#1=16 \cellcolor{rank16}#2%
  \else\ifnum#1=17 \cellcolor{rank17}#2%
  \else\ifnum#1=18 \cellcolor{rank18}#2%
  \else\ifnum#1=19 \cellcolor{rank19}#2%
  \else\cellcolor{rank20}#2%
  \fi\fi\fi\fi\fi\fi\fi\fi\fi\fi\fi\fi\fi\fi\fi\fi\fi\fi\fi%
}
\title{Don’t Pass@k: A Bayesian Framework for \\ Large Language Model Evaluation}
\author{
Mohsen Hariri,$^{*}$\hspace{0.2em} Amirhossein Samandar,$^{\dagger}$\hspace{0.2em} Michael Hinczewski,$^{\dagger}$\hspace{0.2em} Vipin Chaudhary$^{*}$ \\
$^{*}$Department of Computer and Data Sciences, Case Western Reserve University, Cleveland, OH, USA \\
$^{\dagger}$Department of Physics, Case Western Reserve University, Cleveland, OH, USA \\
\texttt{\{mohsen.hariri, amirhossein.samandar, mxh605, vipin\}@case.edu} \\
}

\newcommand{\icon}[1]{\adjustbox{valign=m}{\includegraphics[width=2ex,height=2ex,keepaspectratio]{#1}}}

\newcommand{\skyicon}{\icon{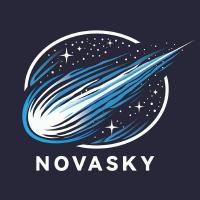}}
\newcommand{\qwenicon}{\icon{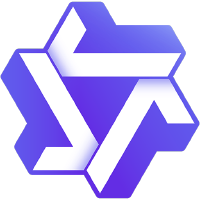}}
\newcommand{\dsicon}{\icon{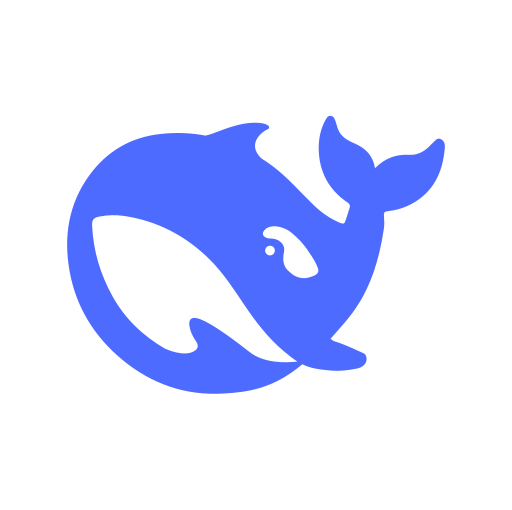}}
\newcommand{\gpticon}{\icon{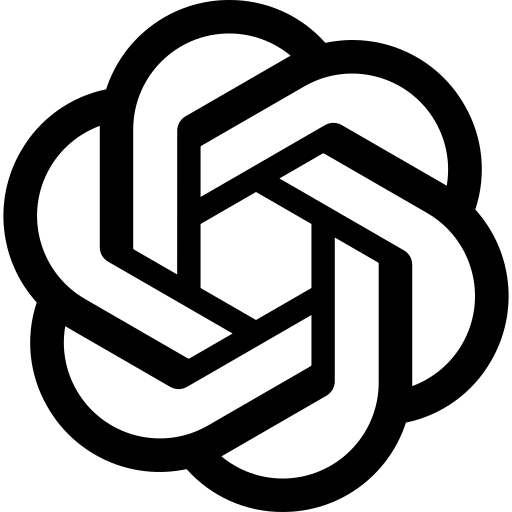}}
\newcommand{\gairicon}{\icon{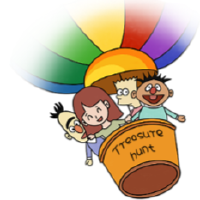}}
\newcommand{\lgicon}{\icon{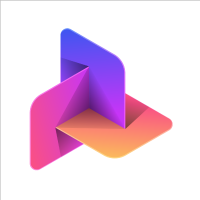}}
\newcommand{\nvidiaicon}{\icon{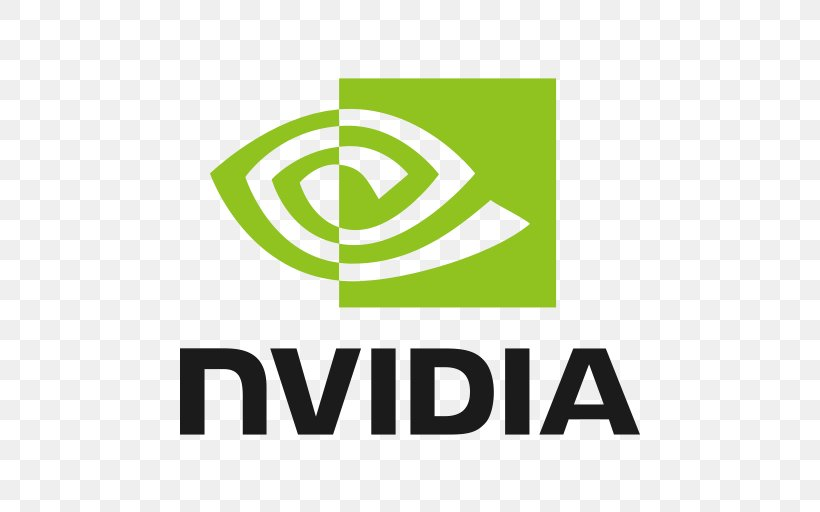}}
\newcommand{\openicon}{\icon{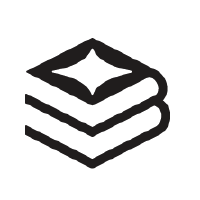}}
\newcommand{\compassicon}{\icon{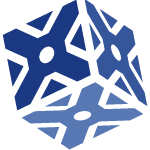}}
\newcommand{\microsoft}{\icon{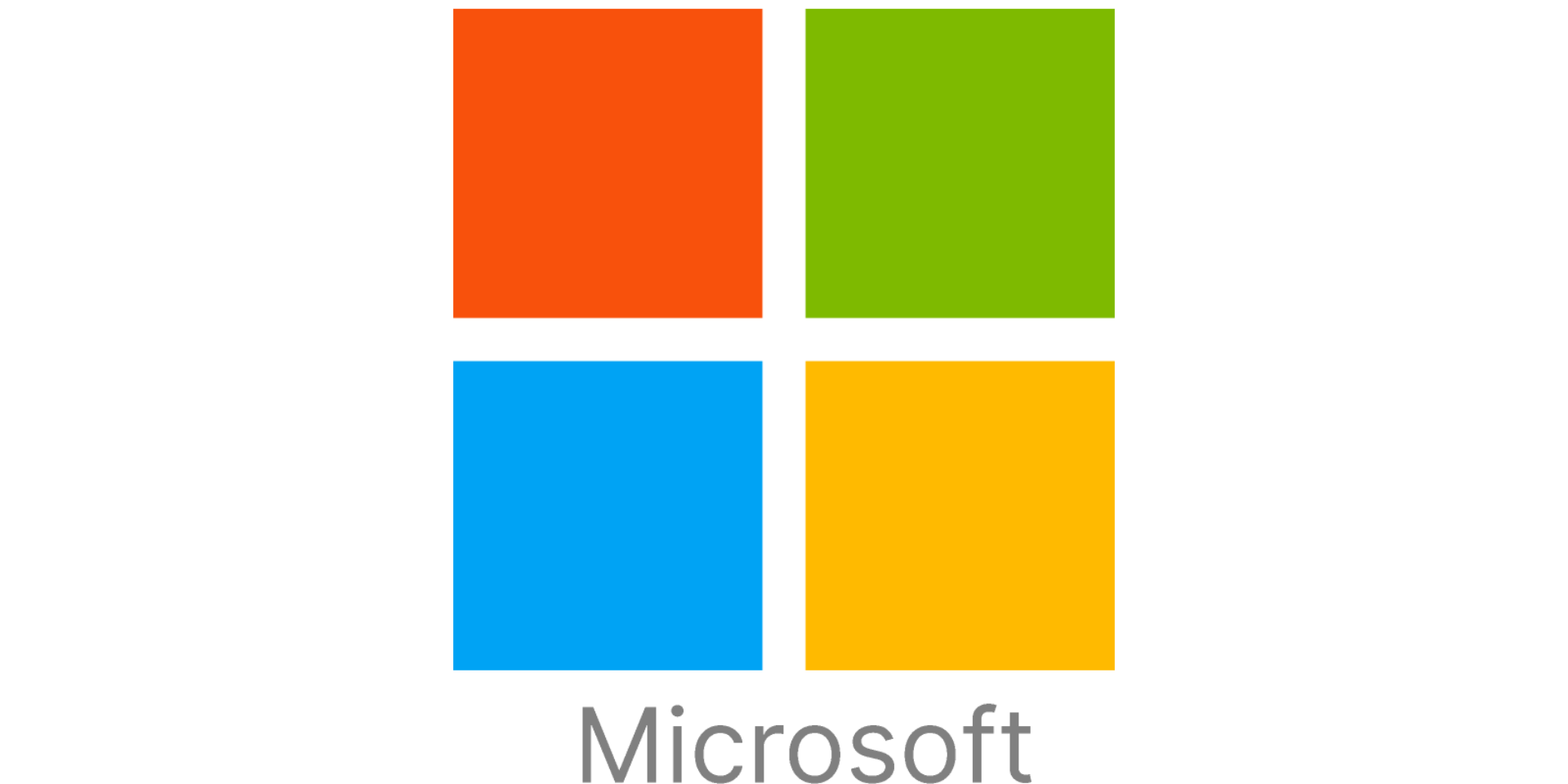}}
\newcommand{\fuse}{\icon{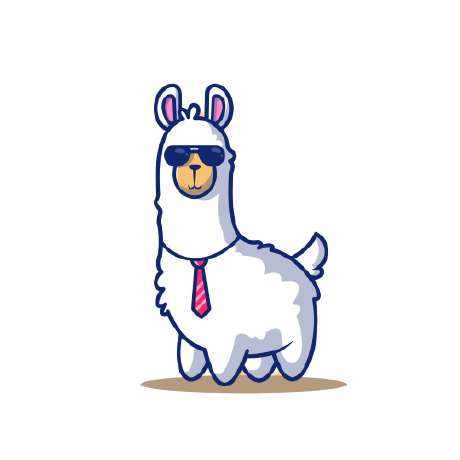}}
\newcommand{\openr}{\icon{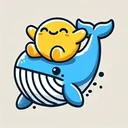}}
\newcommand{\qihoo}{\icon{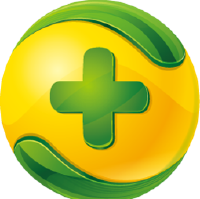}}
\newcommand{\bespoke}{\icon{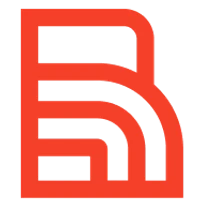}}

\newcommand{\scorio}{\texttt{Scorio}}

\iclrfinalcopy
\begin{document}
\maketitle
\begin{abstract}
\pass{k} is widely used to report the reasoning performance of LLMs, but it often produces unstable and potentially misleading rankings, especially when the number of trials (samples) is limited and computational resources are constrained. We present a principled Bayesian evaluation framework that replaces \pass{k} and average accuracy over $N$ trials (\avg{N}) with posterior estimates of a model's underlying success probability and credible intervals, yielding stable rankings and a transparent decision rule for differences. Evaluation outcomes are modeled as categorical (not just 0/1) with a Dirichlet prior, giving closed-form expressions for the posterior mean and uncertainty of any weighted rubric and enabling the use of prior evidence when appropriate. Theoretically, under a uniform prior, the Bayesian posterior mean is order-equivalent to average accuracy (\pass{1}), explaining its empirical robustness while adding principled uncertainty. Empirically, in simulations with known ground-truth success rates and on \aimeboth, \hmmt, and \brumo, the posterior-based procedure achieves faster convergence and greater rank stability than \pass{k} and recent variants, enabling reliable comparisons at far smaller sample counts. The framework clarifies when observed gaps are statistically meaningful (non-overlapping credible intervals) versus noise, and it naturally extends to graded, rubric-based evaluations. Together, these results recommend replacing \pass{k} for LLM evaluation and ranking with a posterior-based, compute-efficient protocol that unifies binary and non-binary evaluation while making uncertainty explicit. Source code is available at \href{https://github.com/mohsenhariri/scorio}{\adjustbox{valign=m}{\includegraphics[height=3.0ex]{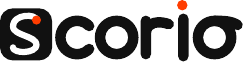}}} \footnote{\url{https://github.com/mohsenhariri/scorio}. See Appendix~\ref{sec:app:scorio} for API documentation.}.

\end{abstract}

\section{Introduction}

\begin{figure}[t]
    \captionsetup{type=figure,font=small}
    \includegraphics[width=\linewidth]{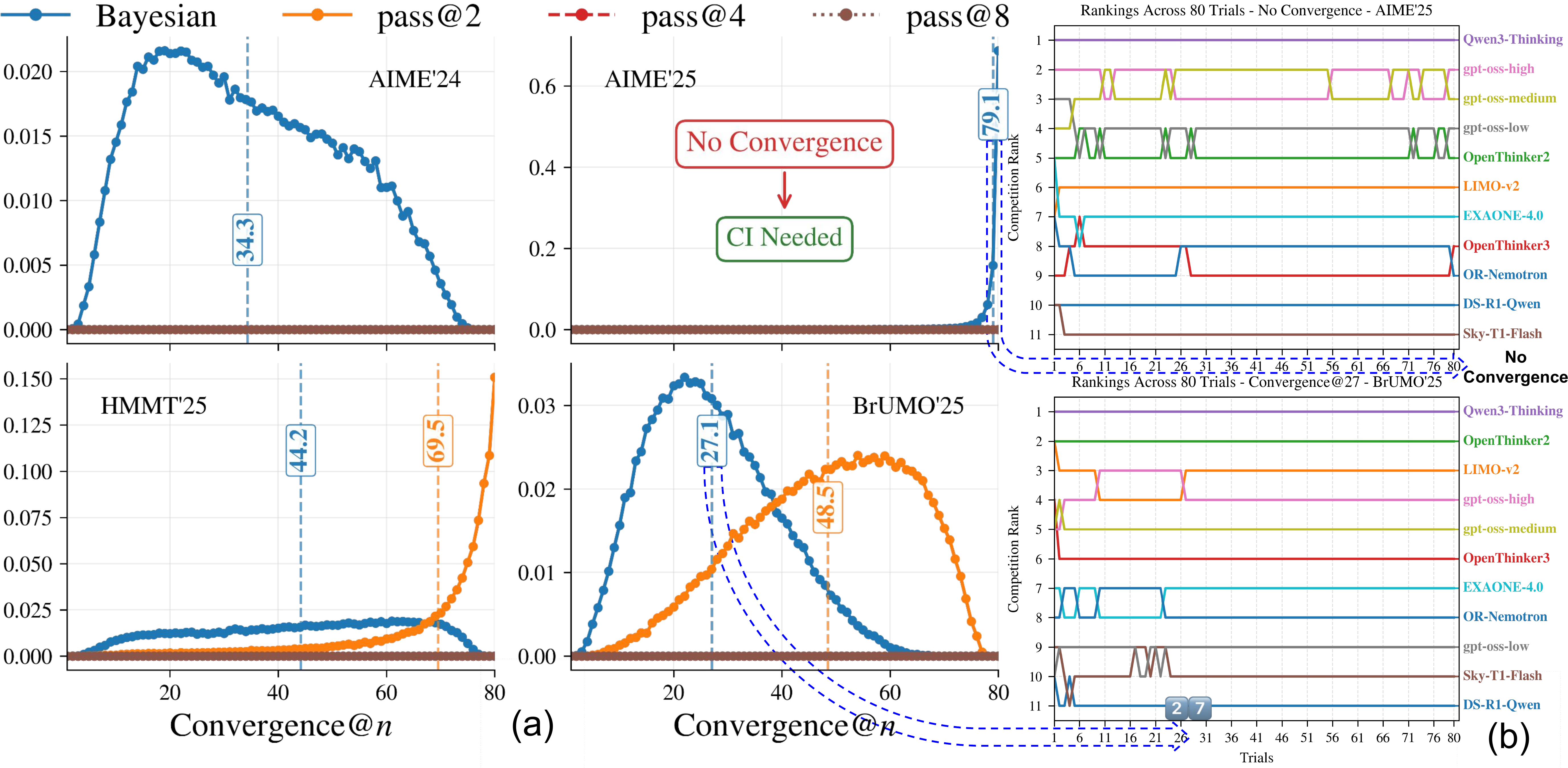}
    \caption{a) Probability mass functions (PMFs) of convergence@$n$, the number of trials $n$ above which a ranking of LLM models consistently matches the ranking using $N_\text{max}=80$ trials. Eleven LLM models (listed on the right) and four math-reasoning datasets are used---\aimefour, \aimefive, \hmmt, and \brumo---comparing \pass{2}/4/8 against our Bayesian posterior evaluation (\bayes{N}). Each PMF is estimated by bootstrapping with $10^5$ samples over the $N_\text{max}$ trials; vertical lines indicate the mean of each convergence distribution. On \aimeboth, the Pass family frequently \emph{fails to converge}, whereas \bayes{N} converges. On HMMT and BrUMO, Pass methods converge more slowly (mean required trials $\approx 69.5$ and $\approx 48.5$) than \bayes{N} ($\approx 44.2$ and $\approx 27.1$), respectively. Right: Example competition-style ranking from a single bootstrap replicate, highlighting the mean convergence for \aimefive\, and \brumo. Per task rankings, including worst-case replicates, are in Appendix~\ref{app:conv} (\Cref{fig:combined_rankings_1}).}
    \label{fig:showcase}
\end{figure}
Large language models (LLMs) have moved rapidly from research artifacts to everyday infrastructure~\cite{vaswani2017attention,brown2020gpt3}. Students use them for homework and exam preparation; developers rely on them for code synthesis and refactoring~\cite{stackoverflow2025AI}; analysts and clinicians use them for decision support; and agents built atop LLMs are increasingly embedded in workflows across industry and government. This demand has catalyzed unprecedented investment: specialized chips, datacenters, and startups dedicated to LLM training, serving, and tooling~\cite{maslej2025artificial}. As deployment accelerates, trust, oversight, and comparability become central: \emph{how we evaluate LLMs} directly shapes which models are adopted, what progress is declared, and how resources are allocated~\cite{liang2022helm,hendrycks2021mmlu,srivastava2022bigbench,kaplan2020scaling,hoffmann2022chinchilla,wei2022cot,ouyang2022instructgpt}.

Evaluation, however, remains the weakest link in the LLM pipeline. Alongside advances in model efficiency and compression\citep{Dettmers2022LLMint8,Frantar2022GPTQ,Han2015Pruning,Hinton2015Distillation,kwon2023efficient,Zhang2024KV1Bit,Zhang2024PQCache,hariri2026quantize}, training and fine-tuning methods such as parameter-efficient fine-tuning (PEFT), low-rank adaptation (LoRA), and reinforcement learning from human feedback (RLHF) \citep{Hu2021LoRA,Christiano2017Prefs,ouyang2022instructgpt}, and inference/decoding (sampling strategies, caching, efficient attention) \citep{Holtzman2019Degeneration,Dao2022FlashAttention}, the community still leans on simple, yet flawed, success rates and \pass{k}-style metrics to summarize capabilities \citep{chen2021evaluating}. These practices are convenient but fragile. On small or costly benchmarks (e.g., math reasoning sets with only tens of problems such as AIME) \citep{MAA_AIME2024,MAA_AIME2025}, \pass{k} or single-run accuracy often produce unstable rankings \citep{liu2024your}, are sensitive to decoding choices and seed effects \citep{Holtzman2019Degeneration,hochlehnert2025sober}, and provide little guidance on whether observed gaps are meaningful or mere noise \citep{Dror2018Significance,Yeh2000Significance}. Averaging across multiple runs (``\avg{N}'') helps but is compute-hungry \citep{Dodge2019ShowYourWork}, offers no unified way to handle graded/rubric outcomes, and lacks a principled decision rule for significance \citep{Dror2018Significance,Zheng2023LLMJudge,Chen2024LLMJudgeBias}.

This paper takes a different approach: we treat evaluation itself as a statistical inference problem. We introduce a \emph{posterior-based} framework that replaces \pass{k} and \avg{N} with estimates of a model’s underlying success probabilities and associated uncertainty~\cite{xiao2025confidence}. Outcomes are modeled as \emph{categorical}~\cite{hayden2025straightforward} rather than purely binary: each item can yield correct, partially correct, formatting-error, refusal, or rubric-defined levels. A Dirichlet prior over these categories yields closed-form posterior means and credible intervals for any \emph{weighted rubric}, allowing the evaluator to report both a point estimate and principled uncertainty with negligible overhead. In the binary special case under a uniform prior, its posterior mean is order-equivalent to average accuracy, explaining the empirical robustness of avg@$N$ while making uncertainty explicit.

The framework addresses \emph{four} persistent pain points. {\large \ding{202}} \emph{Convergence}: as shown in \Cref{fig:showcase}, we ideally want methods that can converge to the true underlying ranking with the smallest number of trials, but different approaches can have significantly different convergence speeds. {\large \ding{203}}
\emph{Credible intervals}: a simple, transparent rule---\textbf{do not declare a winner when intervals overlap}---reduces leaderboard churn and over-interpretation of tiny gaps by introducing a compute-efficient credible interval (CI).  Updates are analytic; one can monitor interval widths online, and allocate additional trials only when needed (no Monte Carlo/bootstrap simulations are required for CI estimation). {\large \ding{204}} \emph{Categorical evaluation}: our approach unifies binary and non-binary evaluation. Graded rubrics are natural in this framework, so one can evaluate step-by-step reasoning, partial credit, or judge categories without ad hoc aggregation. {\large \ding{205}} \emph{Prior information}: we can incorporate prior evidence when appropriate (e.g., reuse of stable rubric distributions across closely related tasks or versions).

We validate the approach in two settings: In controlled simulations with known ground-truth success rates, the posterior procedure converges to correct rankings with fewer samples than \pass{k} and recent variants, and it flags when ties are statistically unresolved. On real math-reasoning benchmarks (\aimeboth~\cite{MAA_AIME2024,MAA_AIME2025}, \hmmt~\cite{HMMT_Feb2025}, and \brumo~\cite{BrUMO_2025}-derived sets), we observe the same pattern: the posterior method achieves greater rank stability at far smaller sample counts than \pass{k}, while clarifying when differences are meaningful versus noise. Practically, this yields a computationally efficient protocol that is easy to implement and audit.

We summarize our contributions as follows:

\begin{itemize}[noitemsep, topsep=0pt]
\item \textbf{A unified Bayesian evaluation framework.} We model per-item outcomes as categorical with a Dirichlet prior, yielding closed-form posterior means and credible intervals for \emph{any} weighted rubric, with binary evaluation as a special case. This unifies 0/1 and graded evaluations and supports reuse of prior evidence when justified.

\item \textbf{A compute-efficient, interval-aware protocol.} We provide a simple recipe: report posterior means with credible intervals; only declare differences when intervals do not overlap; adaptively allocate additional samples until intervals meet pre-specified widths. This protocol naturally supports sequential/online evaluation.
\item \textbf{Empirical evidence on simulations and math benchmarks.} On synthetic data with known ground truth and on AIME’24/’25, \hmmt, and \brumo{} datasets, our method achieves faster convergence and greater rank stability than \pass{k} and recent variants, enabling reliable comparisons with far fewer samples.
\end{itemize}

\section{Bayesian Framework for Evaluating LLM Performance}\label{sec:Bayesian_framework}
\subsection{Background: The \texorpdfstring{\pass{k}}{k} Metric and Its Limitations}

Evaluation metrics for LLMs aim to quantify performance on tasks like reasoning or programming, but they often struggle to provide reliable relative rankings across models. \pass{k}, for instance, estimates the probability of at least one correct answer within $k$ model attempts (see Appendix~\ref{app:extended_related_work} for details). While convenient, this metric exhibits high variance~\cite{dalal2025leveraging}, particularly when $k$ approaches the total number of trials, $N$, resulting in unstable rankings~\cite{chen2021evaluating}. Small fluctuations in correctness can distort comparisons, particularly in benchmarks with few problems or limited computational resources, raising doubts about its suitability for differentiating model capabilities. If a metric cannot consistently distinguish stronger models from weaker ones, its value as a benchmarking tool is undermined~\cite{liu2024your}.

Estimating uncertainty in \pass{k} scores is also challenging, as it lacks closed-form expressions for variance, relying instead on computationally intensive approximations like bootstrapping. A truly effective metric should yield reliable performance rankings with a minimal number of trials, prioritizing both accuracy and efficiency in resource-constrained environments. To address these limitations, we propose a Bayesian evaluation framework that provides more stable estimates of performance, incorporates uncertainty, and facilitates robust relative comparisons across models~\cite{xiao2025confidence,ross2025textual,vashurin2025uncertainty}.

\subsection{Results Matrix}
Consider a results matrix $R$ for an LLM evaluated on a test set comprising $M$ questions. Due to the stochastic nature of LLM sampling, responses may vary across independent trials, so we run the LLM $N$ times per question. The outcomes are captured in the $M \times N$ matrix $R$, where element $R_{\alpha i}$ represents the score in the $i$th trial for the $\alpha$th question. This score is an integer ranging from $0$ to a maximum value $C$, reflecting a rating system with $C+1$ categories. In the binary case ($C=1$), 0 indicates an incorrect answer and 1 a correct one, though we accommodate more nuanced rubrics generally.

\subsection{Weighted Performance Metric}\label{theo:form}

For the $\alpha$th question, $\alpha = 1, \ldots, M$, there is an underlying probability $\pi_{\alpha k}$ that the LLM's answer falls in the $k$th category. We denote $\bm{\pi}_\alpha$ as the $(C+1)$-dimensional vector with elements $\pi_{\alpha k}$, $k=0, \ldots, C$. If all $\bm{\pi}_\alpha$ were known, we could calculate a desired performance metric $\bar{\pi}$ as a weighted average over these probabilities:

\begin{equation}
\label{eqn:gold_distribution}
\bar{\pi} = \frac{1}{M} \sum_{\alpha=1}^M \bm{w} \cdot \bm{\pi}_\alpha = \frac{1}{M} \sum_{\alpha=1}^M \sum_{k=0}^C w_k \pi_{\alpha k},
\end{equation}

where $\bm{w}$ is a $(C+1)$-dimensional vector of constant weights. For example, if $w_k = k$, then $\bar{\pi}$ represents the average category label. In the case where $C=1$, this average corresponds to the mean probability of a correct answer over the entire test set. However, we allow for a general choice of $\bm{w}$ to accommodate a wide range of possible metrics.

\subsection{Bayesian Estimator and Uncertainty for the Performance Metric}
\label{subsec:Bayesian_uncertainty}

In principle, we could estimate $\bm{\pi}_\alpha$ by running an arbitrarily large number of trials with the LLM, yielding an accurate estimate of $\bar{\pi}$. However, we are typically constrained to small $N$ due to limited computational resources. Our goal is to develop a Bayesian approach to estimate $\bar{\pi}$ and its associated uncertainty given a finite $N$. The first step is to construct ${\cal P}(\bm{\pi}_\alpha | \bm{R}_\alpha)$, the posterior probability of $\bm{\pi}_\alpha$ given the $\alpha$th row of the matrix $R$, denoted $\bm{R}_\alpha$. This posterior depends on the data in $\bm{R}_\alpha$ and a chosen prior distribution ${\cal P}(\bm{\pi}_\alpha)$ for the unknown underlying probability vector $\bm{\pi}_\alpha$. The prior could be uniform (assuming no prior information) or incorporate previously gathered evidence about the LLM's performance. The Bayesian framework focuses on two quantities: the first is the mean of $\bar{\pi}$ over the joint posterior for all questions, which we denote as $\mu(R)$. This is a Bayesian optimal estimator, minimizing the quadratic loss function ${\cal L}(\bar{\pi}^\text{est}) = {\mathbb E}_{R,\bm{\pi}_\alpha} (\bar{\pi}^\text{est}(R) - \bar{\pi})^2$ over all possible estimators $\bar{\pi}^\text{est}(R)$, where the expectation value is over all possible $\bm{\pi}_\alpha$ and realizations of $R$ \cite{jaynes2003probability}. The second quantity is the variance $\sigma^2(R)$,
which quantifies the uncertainty of the $\mu$ estimate. Both $\mu(R)$ and $\sigma^2(R)$ have exact closed-form expressions, derived in Appendix~\ref{app:Bayes_derivation}, and can be simply calculated for any $R$ using Algorithm~\ref{summary}.

\begin{algorithm}
    \caption{LLM performance evaluation using the \bayes{N} framework.}
    \label{summary}
    \begin{algorithmic}
        \Function{EvaluatePerformance}{$R$, $[R^0]$, $\bm{w}$}
            \State \textbf{input:} $M \times N$ matrix $R$ of results, with each element $R_{\alpha i} = 0,\ldots,C$
            \State $\qquad\quad$ weight vector $\bm{w} = (w_0,\ldots,w_C)$ defining performance metric $\bar{\pi}$
            \State \textbf{optional input:} $M \times D$ matrix $R^0$ of results for prior; otherwise $D=0$
            \State \textbf{output:} performance metric estimate $\mu$ and associated uncertainty $\sigma$
            \vspace{0.5em}
            \State $T = 1+C+D+N$
            \For{$\alpha = 1$ to $M$} \Comment{Tally results in $R$ and $R^0$}
            \For{$k = 0$ to $C$}
                \State $n_{\alpha k} = \sum_{i=1}^N \delta_{k,R_{\alpha i}}$
                \State $n^0_{\alpha k} = 1+\sum_{i=1}^D \delta_{k,R^0_{\alpha i}}$
                \State $\nu_{\alpha k} = n^0_{\alpha k} + n_{\alpha k}$
            \EndFor
            \EndFor
            \State $\mu = w_0 + \frac{1}{M T} \sum_{\alpha=1}^M \sum_{j=0}^C \nu_{\alpha j}(w_j - w_0)$
            \State $\sigma = \left[\frac{1}{M^2 (T+1)}\sum_{\alpha=1}^M \left\{ \sum_{j=0}^C \frac{\nu_{\alpha j}}{T} (w_j-w_0)^2 - \left(\sum_{j=0}^C \frac{\nu_{\alpha j}}{T} (w_j-w_0) \right)^2\right\}\right]^{1/2}$
            \State \textbf{return} $\mu$, $\sigma$
        \EndFunction
    \end{algorithmic}
\end{algorithm}

\subsection{Using Uncertainty Estimates to Decide Significance of Performance Differences}
\label{subsec:Bayesian_Uncertainty}

In general, the expressions for $\mu(R)$ and $\sigma^2(R)$ are valid for any $M$ and $N$, and do not rely on asymptotic arguments like the central limit theorem (CLT). However, there are useful simplifications that occur in specific limiting cases. For example as the size of the test set $M$ becomes large, we can derive not just the moments of the posterior distribution for $\bar{\pi}$, but also its shape, which becomes approximately Gaussian:
${\cal P}(\bar{\pi} | R) \sim {\cal N}(\mu(R), \sigma^2(R))$. This allows us to assess whether two methods exhibit a statistically significant performance difference. Consider results matrices $R$ and $R'$ from two approaches, with corresponding means $\mu$, $\mu'$ and standard deviations $\sigma$, $\sigma'$. The distribution of the performance difference $\Delta \bar{\pi} \equiv \bar{\pi} - \bar{\pi}'$ is a convolution of the individual posteriors, yielding another normal distribution: ${\cal P}(\Delta\bar{\pi} | R, R') \sim {\cal N}(\tilde{\mu},\tilde\sigma^2)$,
where the mean of the difference is $\tilde{\mu} = \mu - \mu'$, and the standard deviation is $\tilde{\sigma} = \sqrt{\sigma^2 + (\sigma')^2}$. To determine our confidence in the ranking of the two methods, we need to determine the probability that $\text{sign}(\Delta \bar\pi) = \text{sign}(\mu - \mu^\prime)$. This can be done by calculating the absolute $z$-score, $z = |\mu - \mu'|/\sqrt{\sigma^2 + (\sigma')^2}$. The probability that the ranking based on $\mu$ and $\mu^\prime$ is correct (the ranking confidence $\rho$) is given by $\rho = (1/2)(1+\text{erf}(z/\sqrt{2}))$. For example $z = 1.645$ corresponds to $\rho = 0.95$.

\subsection{Equivalence of Bayesian and Average Rankings for Uniform Prior}
\label{subsec:avergae_Bayes_equiv}

In the results below, we will denote ranking based on the Bayesian estimator $\mu$ with a uniform prior as \bayes{N}. Because $\mu$ is related to a naive weighted average accuracy via a positive affine transformation, it turns out the ranking based on the average, denoted as \avg{N}, is identical to \bayes{N} (for the detailed proof, see Appendix~\ref{app:bayes_avg_equiv}). In the large-trial limit $N\to\infty$, the value of $\mu$ approaches the average, as expected, but the ranking equivalence holds at all finite $N$. This relationship also extends to uncertainty quantification, where the standard deviation of the average relates to the Bayesian standard deviation $\sigma$ by a scaling factor, providing a concrete method to compute uncertainty in the average without relying on the Central Limit Theorem. This is particularly advantageous in small-sample regimes common in LLM evaluations, where CLT-based methods often underestimate uncertainty and produce invalid intervals (e.g., extending beyond [0,1] or collapsing to zero) \cite{bowyer2025position}. As highlighted by \cite{bowyer2025position}, Bayesian approaches with uniform priors (e.g., Beta(1,1) in the binary case) yield well-calibrated credible intervals even for datasets with fewer than a few hundred datapoints, outperforming CLT approximations in coverage and handling complex structures like clustered data.

\subsection{Gold Standard for Ranking}
\label{sec:gold_standard}

Strictly speaking, the underlying true ranking of LLMs for a particular performance metric $\bar{\pi}$ is unknown, because it would require determining the infinite trial limit, $\bar{\pi} = \lim_{N \to \infty} \mu$, for each LLM. In practice, we have to settle for an approximation to $\bar{\pi}$, calculated at some large but finite value $N = N_\text{max}$ (for example $N_\text{max} = 80$ in our LLM experiments). Specifically, we use \unibayes{N_{\mathrm{max}}}---which is the same as the ranking based on avg@$N_{\mathrm{max}}$---as our ``gold standard'' or reference ranking \cite{hariri2026ranking}. In other words, rankings using smaller $N$ will be compared to this gold standard to assess their accuracy.

For this comparison, we employ Kendall's $\tau$, a nonparametric correlation coefficient that measures ordinal agreement between two rankings by comparing the number of concordant and discordant pairs of models. The coefficient ranges from $-1$ (perfect inversion) to $+1$ (perfect agreement), with $0$ indicating no association. We specifically use the $\tau_b$ variant, which properly accounts for ties in the rankings (e.g., the intentional tie in our simulation below), ensuring that equivalences do not artificially inflate the correlation. See Appendix~\ref{app:kendall_tau} for further discussion and formal definitions.

\begin{figure}[h!]
    \centering
    \includegraphics[width=\linewidth]{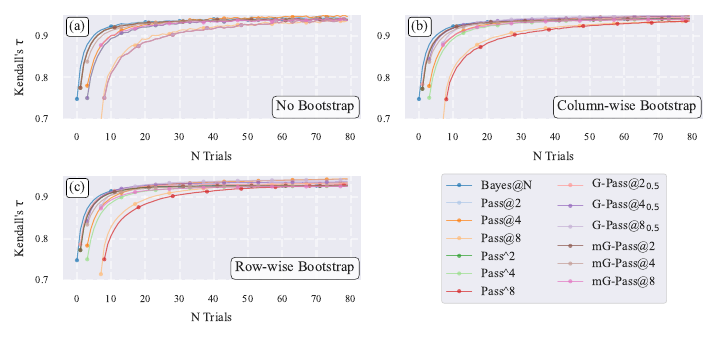}
    \captionsetup{skip=1pt}
    \caption{Kendall's $\tau$ rank correlation for various evaluation methods compared to the true ranking of $11$ sets of biased coins (LLM mimics) with known mean success probabilities $\bar{\pi} = 0.2332$, 0.2545, 0.3604, 0.3642, 0.3642, 0.4466, 0.5418, 0.5276, 0.608 , 0.6213, 0.7327.  The simulation evaluates methods including \pass{k} ($k=2, 4, 8$), \bayes{N},  Pass\textasciicircum$k$, G-Pass@$k_{\tilde\tau}$ ($\tilde\tau=0.5$), and mG-\pass{k} across $1$ to $80$ trials. Panel a) shows $\tau$ results without bootstrapping, while panels b) and c) use two different bootstrapping approaches with $10^4$ samples.}
    \label{fig:biased_coins}
\end{figure}

To validate our claims about the gold standard as \unibayes{N_{\mathrm{max}}}, specifically to determine which evaluation methods converge to the true ranking, we conduct a simulation using biased coins as a metaphor for LLMs. In this setup, we already know the underlying performance distribution (the success probabilities $\bm{\pi}_\alpha$ for each question), allowing us to establish a known ground truth $\bar{\pi}$.
We generate $11$ sets of these $30$ probabilities, with $\bar{\pi}$ values of $[0.2332, 0.2545, 0.3604, 0.3642, 0.3642, 0.4466, 0.5418, 0.5276, 0.608, 0.6213, 0.7327]$, representing different LLMs (note the tie at $0.3642$ to test handling of equivalent performances). We run experiments for $M=30$ questions, where each LLM ``answers'' all the questions in each trial according to its success probabilities $\pi_{\alpha 1}$. Panel (a) of \Cref{fig:biased_coins} shows results without bootstrapping: we generate 1000 independent $R$ matrices, each with $80$ trials; for each step in the number of trials (from $N=1$ to $80$), we compute scores using \pass{k} ($k=2$, $k=4$, and $k=8$ with an unbiased estimator \Cref{eqn:pass_at_k}), \unibayes{N}, Pass\textasciicircum$k$ (\Cref{eqn:pass_to_k}), G-Pass@$k_{\tilde\tau}$ (\Cref{eqn:G_pass_at_k_tau} with $\tilde\tau = 0.5$), and mG-Pass@$k$(\Cref{eqn:mG_pass_at_k}), then derive rankings and compare them to the gold standard using Kendall's $\tau$ as a measure of rank correlation (where $\tau=1$ indicates perfect alignment with the gold standard), and report the average $\tau$ over the $1000$ $R$ matrices. Note that we do not explicitly show average accuracy avg@$N$ because it is equivalent to \unibayes{N}, as discussed in \Cref{subsec:avergae_Bayes_equiv}.
In practice, we are computationally limited to a small number of trials per question. To examine what happens with only $N=80$ trials, we apply two methods of bootstrapping with replacement to the $R$ matrix, allowing us to estimate how results differ from the ideal case with a large number of independent $R$ matrices (panel a). For both methods, we generate $10{,}000$ bootstrap replicates for each of the $N=1$ to $80$ trials, derived from a single $R$ matrix. Panels (b) and (c) of \Cref{fig:biased_coins} illustrate this using two bootstrapping schemes. In the first scheme (panel b, column-wise bootstrapping), we resample trial indices; in the second (panel c, row-wise bootstrapping), we resample answers independently for each question. In both cases, the resulting bootstrap replicates are used to recompute evaluation scores, rankings, and $\tau$ values, which are then averaged to produce smoothed convergence curves. The two bootstrapping approaches yield nearly identical behavior, and both closely match the baseline in panel (a). This demonstrates that the $\tau$ convergence behavior is robust and not sensitive to the ordering of answers in either the rows or columns of $R$. Though in our LLM mimic simulations, we do not have to use bootstrapping (since we can easily generate an arbitrarily large number of $R$ matrices), in actual LLM experiments, we have limited trial data, and these results show that bootstrapping provides a viable way of estimating statistical properties like convergence.

As seen in \Cref{fig:biased_coins}, \unibayes{N} begins with relatively high agreement with the gold standard and converges much faster to $\tau =1$ than Pass@$k$ and its variants, which suffer from greater variance and bias at small $N$. All methods eventually converge to the same ranking, but their rates of convergence differ substantially. This makes the convergence rate a crucial factor when choosing between different LLM evaluation methods.

\subsubsection{Potential benefits of non-uniform priors}
\label{app:non_uni_prior}

While the convergence results in \cref{fig:biased_coins} demonstrate that \bayes{N} with a uniform prior outperforms alternatives like \pass{k} in ranking models, there are scenarios where non-uniform priors can achieve even faster convergence. This is the case when we have data from models that are related or closely correlated to the ones we are ultimately interested in ranking. Potential examples include: i) results from an older version of a model used as a prior for ranking a newer version; ii) a non-quantized version (where running trials is computationally expensive) used to provide prior data for a quantized version (where achieving large $N$ is cheaper); iii) a base model used to provide prior data for a fine-tuned one. Though a full exploration of these kinds of priors will be left to future work, in this section, we will show the potential benefits through our synthetic biased-coin LLM models, introduced in Sec.~\ref{sec:gold_standard}.

\begin{figure}
    \centering
    \includegraphics[width=\linewidth]{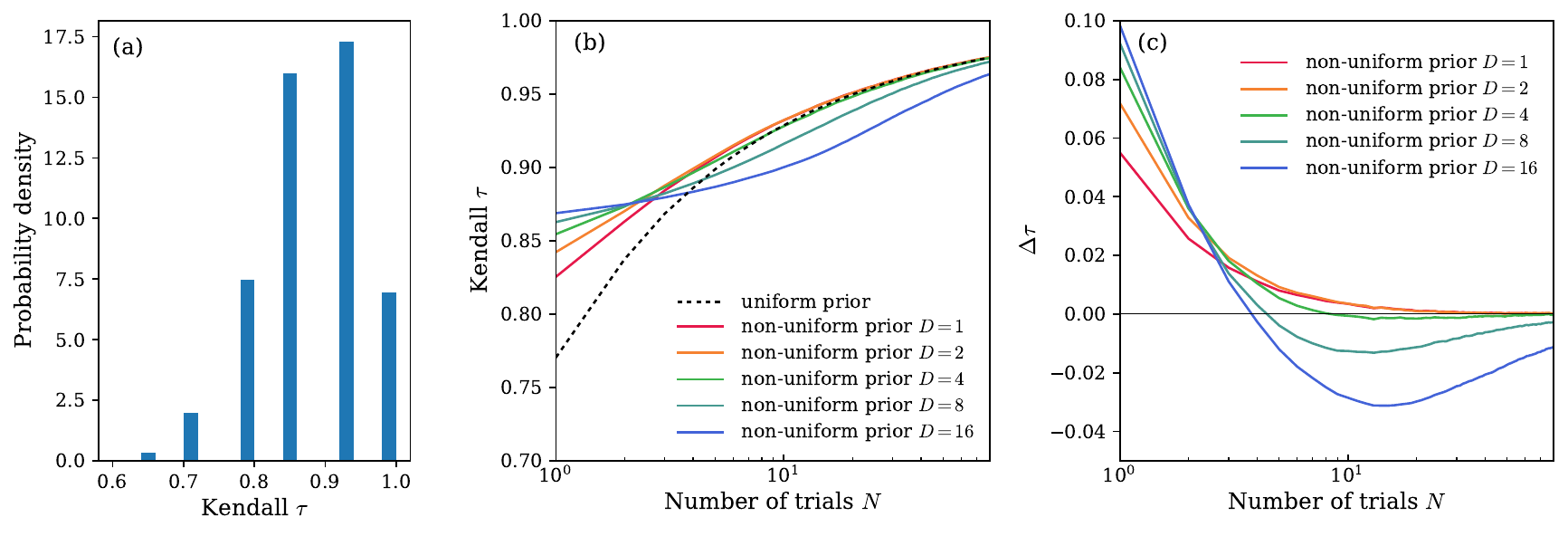}
    \caption{(a) Histogram of Kendall $\tau$ values comparing original ranking of synthetic LLM models and 50k replicates of updated models. (b) Mean Kendall $\tau$ between the estimated and true ranking for the updated models (50k replicates) as a function of $N$, the number of trials. The dashed line corresponds estimates using \bayes{N} with a uniform prior ($D=0$), while the solid lines are \bayes{N} with a non-uniform prior and different choices of $D$. The non-uniform prior is based on results from $D$ trials of the original models. (c) Same as panel (b), except showing the difference $\Delta \tau$ between the non-uniform prior curves and the uniform curve.
    }
    \label{app:nonuniform_prior}
\end{figure}

We start with a set of eight ``original'' models with $C=1$, labeled by $i=1,\ldots,8$. Each model $i$ consists of a set of $M=30$ success probabilities $\pi_{\alpha 1}$ drawn from a distribution Beta$(i+3,12-i)$. We fix these probabilities for all the numerical experiments described below, and their averages for the eight models are: $\bar\pi = \,$[0.3021, 0.3166, 0.4144, 0.4985, 0.5351, 0.5759, 0.6679, 0.7487]. Hence for the original models higher $i$ corresponds to higher overall accuracy. We now imagine an ``update'' of model $i$ that mimics some kind of revision, fine-tuning, or other modification. Because the performance of the updated model should be correlated with the original, we model the update as a stochastic perturbation to the Beta distribution from which success probabilities are drawn: for updated model $i$ the $\pi_{\alpha 1}$ values are drawn from Beta$(i+3+\sigma, 12-i+\sigma^\prime)$, where $\sigma = \pm 1$ and $\sigma^\prime = \pm 1$ are random integers of unit magnitude. For the updated models the value of $\bar\pi$ may not strictly increase with $i$, so the ranking of models could be different than the original. Fig.~\ref{app:nonuniform_prior}(a) shows a histogram of the Kendall's $\tau$ values comparing the original model set (described above) and 50k possible updated sets drawn using this stochastic procedure. A $\tau$ value of 1 corresponds to exactly the same ranking, and we see that the mean $\tau$ over the 50k realizations is 0.88. Hence there is some correlation between the original and updated rankings, but in the vast majority of cases (about 86\% of the updates), the ranking has changed for the updated models.

The question we would like to ask is whether we can use the results from the original models as priors to help speed up convergence when ranking the updated models.  To employ a non-uniform prior for a given model, we follow the procedure described in Appendix~\ref {app:Bayes_derivation}, and incorporate the prior via the $M \times D$ results matrix $R^0$ corresponding to $D$ trial results over $M$ questions using the original model. Combined with $N$ trial results from the updated model, we get the \bayes{N} accuracy estimate $\mu$ for the updated model. These estimates are then used to rank the 8 updated models. Because we know the $\bar\pi$ values for this set, we know the true ranking, and we can compare the estimated and true rankings via Kendall's $\tau$.

For each choice of $N$ and $D$ we run 50k replicates, with each replicate consisting of a set of stochastic updates of the original models. The mean $\tau$ values over all these replicates are shown in Fig.~\ref{app:nonuniform_prior}(b) as a function of $N$ for several different $D$. As expected, the $\tau$ curves increase with $N$, since the ranking becomes more certain with more trials, but the convergence properties vary. The dashed line is the case of a uniform prior ($D=0$), while the solid lines represent five different non-uniform prior scenarios, with $D=1$, 2, 4, 8, and 16. For small $N$ and small $D \le 4$ we see a clear benefit of the non-uniform prior: already at $N=1$, the value of $\tau$ starts higher than the uniform case, and remains so until the latter catches up for $N > 5$. Thus when we have prior data available, we can extract more accurate rankings with just a small number of trials of the updated model, relative to the uniform case. However there is a possibility to over-emphasize the prior: when $D=8$ and 16, the benefit for small $N$ turns into a disadvantage at larger $N$. The $\tau$ curves dip beneath the $D=0$ result, indicating that the prior has impeded accurate ranking. Fig.~\ref{app:nonuniform_prior}(c) shows these trends more clearly by plotting $\Delta \tau$, the difference between the $\tau$ for each $D$ and the uniform $\tau$ with $D=0$. So we see that priors have to be used judiciously, with a large enough $D$ to nudge the ranking in the correct direction, but not too large to outweigh the results from the updated models. One of the goals of our future work will be to establish practical guidelines for $D$ in different real-world use cases.

\subsubsection{Ranking with Uncertainty}
\label{sec:sim_ranking_variance}

In \Cref{subsec:Bayesian_Uncertainty}, we described how uncertainty estimates from the Bayesian approach can be used to evaluate the relative performance of two models. Here, we extend these ideas to incorporate uncertainty into the ranking of multiple models. We do this via our biased-coin LLM mimics, which we denote LLM$_\beta$ for $\beta=1,\ldots,11$, described in the previous section. To incorporate a chosen credible interval in the ranking, we order their $\mu$ values from highest to lowest, choose the appropriate $z$ threshold (for example $z = 1.645$ for 95\% CI in the ranking), and assign two consecutive methods the same ranking if the absolute $z$-score falls below this threshold.

The first row of Table~\ref{tab:sim_ranking} shows the underlying gold standard ranking for all the LLM mimics, since in this case we know the true $\bar{\pi}$ values. Note the tie between LLM$_4$ and LLM$_5$, because their $\bar{\pi} = 0.3642$ is the same. The second row shows the \unibayes{80} ranking without a credible interval (CI) and the third row shows \unibayes{80} incorporating the $95\%$ CI.  The \unibayes{80} ranking without CI aligns with the gold standard, except for two differences: the order of $\mathrm{LLM}_{10}$ and $\mathrm{LLM}_{9}$ is swapped, and the tie between $\mathrm{LLM}_{5}$ and $\mathrm{LLM}_{4}$ is not captured, which is expected since this ranking relies solely on $\mu$ estimates without accounting for uncertainty $\sigma$. In contrast, the third row, which incorporates the CI, reveals multiple ties across several models. Interestingly, $\mathrm{LLM}_{10}$ and $\mathrm{LLM}_{9}$ are now indistinguishable at the $95\%$ CI. Despite the fact that $N=80$ would be an atypically large number of trials for an actual LLM evaluation, it is insufficient to confidently distinguish the small performance difference ($\bar{\pi} = 0.608$ vs. 0.6213) between the two models.

\begin{table}[h!]
\centering
\setlength{\tabcolsep}{3pt}
\caption{Comparison of biased-coin LLM mimic rankings based on the gold standard, \unibayes{80} without credible interval (CI), and \unibayes{80} with CI.}
\label{tab:sim_ranking}
\begin{tabular}{l *{11}{c}}
\toprule
 LLM mimic  & \rotatebox{0}{LLM$_{11}$} & \rotatebox{0}{LLM$_{10}$} & \rotatebox{0}{LLM$_9$} & \rotatebox{0}{LLM$_8$} & \rotatebox{0}{LLM$_7$} & \rotatebox{0}{LLM$_6$} & \rotatebox{0}{LLM$_5$} & \rotatebox{0}{LLM$_4$} & \rotatebox{0}{LLM$_3$} & \rotatebox{0}{LLM$_2$} & \rotatebox{0}{LLM$_1$} \\
\midrule
\textbf{Gold Standard} & \rankcell{1}{1} & \rankcell{2}{2} & \rankcell{3}{3} & \rankcell{5}{5} & \rankcell{4}{4} & \rankcell{6}{6} & \rankcell{7}{7} & \rankcell{7}{7} & \rankcell{8}{8} & \rankcell{9}{9} & \rankcell{10}{10} \\
\textbf{\unibayes{80} (w/o CI)} & \rankcell{1}{1} & \rankcell{3}{3} & \rankcell{2}{2} & \rankcell{5}{5} & \rankcell{4}{4} & \rankcell{6}{6} & \rankcell{7}{7} & \rankcell{8}{8} & \rankcell{9}{9} & \rankcell{10}{10} & \rankcell{11}{11} \\
\textbf{\unibayes{80} (w/ CI)} & \rankcell{1}{1} & \rankcell{2}{2} & \rankcell{2}{2} & \rankcell{3}{3} & \rankcell{3}{3} & \rankcell{4}{4} & \rankcell{5}{5} & \rankcell{5}{5} & \rankcell{5}{5} & \rankcell{6}{6} & \rankcell{7}{7} \\
\bottomrule
\end{tabular}
\end{table}

\begin{figure}[h!]
    \centering
    \includegraphics[width=\linewidth]{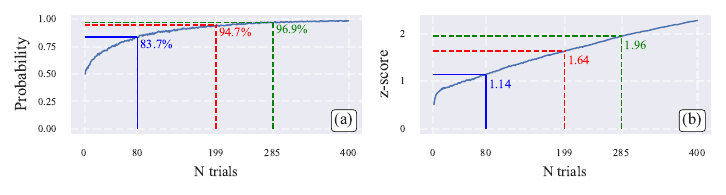}
     \captionsetup{skip=0.1pt}
    \caption{(a) Probability of correctly ranking $\mathrm{LLM}_{10}$ above $\mathrm{LLM}_{9}$ using \unibayes{N} in the biased-coin simulations, shown as a function of trial count $N$. The probability is $83.7\%$ at $N=80$, increases to $\sim 94.7\%$ at $N=199$, and reaches $96.9\%$ at $N=285$. (b) Corresponding absolute $z$-scores as a function of $N$, with values of $\sim 1.14$ at $N=80$, $1.645$ at $N=199$ ($95\%$ confidence), and $1.96$ at $N=285$ ($97.5\%$ confidence).}
    \label{fig:CI_sim}
\end{figure}

To quantify the trials needed to reliably separate models with closely matched performance, we simulated the probability of correctly ranking $\mathrm{LLM}_{10}$ above $\mathrm{LLM}_{9}$ as a function of the number of trials $N$, shown in the left panel of \Cref{fig:CI_sim}. At $N = 80$, the probability of obtaining the correct ranking is $83.7\%$. The right panel plots the absolute $z$-score versus $N$; at $N = 80$, the $z \sim 1.14$, corresponding to approximately $87\%$ confidence (though the plots exhibit some noise due to simulation variability). These values closely align with the empirical probabilities in the left panels.

We also determined the minimum sample size $N$ needed to achieve z-scores of $1.645$ and $1.96$, corresponding to CI of approximately $95\%$ and $97.5\%$, respectively, for distinguishing between models. These thresholds occur at about $N = 199$ and $N = 285$. At these values, the simulated probability of correctly ranking the models is $94.7\%$ and $96.9\%$, respectively, which is closely consistent with expectations given the inherent noise in the results. These results underscore the computational cost of distinguishing models whose true performance metrics differ only slightly. In our biased-coin setup, the underlying success probabilities were $\bar{\pi}_{9} = 0.608$ and $\bar{\pi}_{10} = 0.6213$, yet reliably establishing this distinction requires nearly $200$ trials. Such large sample requirements highlight the importance of considering both uncertainty and convergence rates when interpreting ranking-based evaluations.

\section{Experiments}\label{exps}

In this section, we empirically validate our proposed evaluation methods using real-world datasets, focusing on ranking LLMs for mathematical reasoning tasks. We employ bootstrapping to compute the expected value of each evaluation score at a given $N$. First, we present rankings of LLMs on the \aimefour, \aimefive, \brumo, and \hmmt{} datasets without accounting for variance, based solely on evaluation scores (with ties occurring when scores are identical). Subsequently, we demonstrate how incorporating uncertainty in these scores can alter rankings across different datasets.
Building on the discussion in \Cref{sec:gold_standard}, we adopt the ranking derived from \avg{80} (equivalently, \pass{1} evaluated on the same 80 trials) or \unibayes{80} (uniform-prior Bayesian estimator) as our gold standard for comparing current LLMs, noting their equivalence in rankings (as proven in Section~\ref{subsec:avergae_Bayes_equiv}). For each $N$ from $1$ to $80$ (with \pass{k} and similar methods starting from $N=k$ to avoid computation with insufficient samples), we compare the rankings produced by various evaluation methods against this gold standard, reporting the average Kendall's $\tau$ over $10^4$ bootstrapped resamples to estimate the expected rank correlation at each step (assuming independence among questions and trials).

\subsection{Convergence to Gold Standard}
\label{subsec:ranking_w/o_CI_in_4_datasets}

\begin{figure}[h!]
    \centering
    \includegraphics[width=\linewidth]{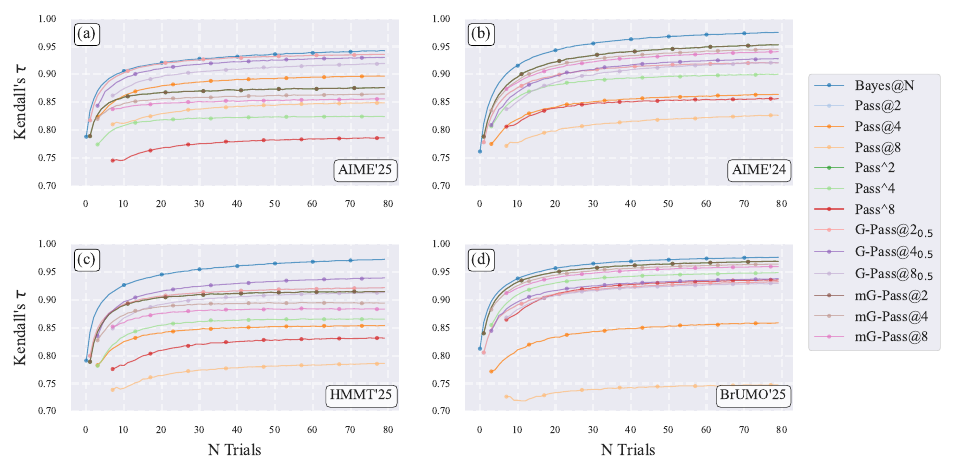}
  \captionsetup{skip=1pt}
    \caption{Average Kendall's $\tau$ correlation between rankings produced by various evaluation methods and the gold standard (derived from \bayes{80}, or equivalently \avg{80}), as a function of the number of trials $N$. Results are averaged over $10^4$ bootstrapped resamples for each dataset: (a) \aimefive, (b) \aimefour, (c) \hmmt, and (d) \brumo. Methods include Bayesian estimation \bayes{N}
    , \pass{k} ($k=2,4,8$),  Pass\textasciicircum$k$, G-Pass@$k_{\tilde\tau}$ ($\tilde\tau=0.5$), and mG-\pass{k}.}
    \label{fig:4_dataset_ranking}
\end{figure}

To assess the ability of different evaluation methods to compare the performance of different LLMs, we plot the average Kendall's $\tau$ against the gold standard as a function of the number of trials $N$ in \Cref{fig:4_dataset_ranking}, combining results from \aimefive{} (panel a), \aimefour{} (panel b), \hmmt{} (panel c), and \brumo{} (panel d). Across all datasets, the \unibayes{N} and \avg{N} curves overlap completely (so we only plot \unibayes{N}) and demonstrate the fastest convergence to high $\tau$ values, indicating robust alignment with the gold standard even in low-sample regimes. In all four datasets, \bayes{N} reaches $\tau > 0.90$ by $N = 10$ and approaches $\tau \approx 1$ at $N \approx 80$. The only exception is \aimefive, where $\tau > 0.90$ is achieved by $N = 10$, but the curve converges to $\tau \approx 0.95$ at $N = 80$.

In contrast, \pass{k} variants ($k=2,4,8$) and their variations (e.g., Pass\textasciicircum$k$, G-Pass@$k_{\tilde\tau}$ with $\tilde\tau=0.5$, mG-\pass{k}) start with lower Kendall's $\tau$ compared to \bayes{N} and converge more slowly in all four datasets. At every $N$, \bayes{N} consistently shows faster convergence and higher agreement with the gold standard. These findings align with our biased-coin simulations in Section~\ref{sec:gold_standard}, demonstrating that the Bayesian method best satisfies the gold-standard criteria---low uncertainty, minimal ties, and rapid convergence---across diverse mathematical reasoning benchmarks.

\subsection{Rankings With Credible Intervals}
\label{subsec:ranking_w_CI_in_4_datasets}
Following the methodology of \Cref{sec:sim_ranking_variance}, we compare model rankings across four datasets (\aimefive, \aimefour, \hmmt, and \brumo) using \unibayes{80} as the gold standard (see \Cref{fig:4_dataset_ranking}). Table~\ref{tab:dataset_ranking} summarizes these comparisons by reporting, for each dataset, two versions of the ranking: the rank \emph{with} a $95\%$ CI and the rank \emph{without} CI. The ``w/ CI'' rank accounts for uncertainty in the Bayes@$80$ scores and therefore allows models with overlapping CIs to share the same rank; the ``w/o CI'' rank is the strict ordering determined by the point estimates of \unibayes{80} for that dataset.

Table~\ref{tab:dataset_ranking} indicates that point-estimate rankings diverge from those accounting for credible intervals. \qwenicon~Qwen3-30B-A3B-Thinking-2507 and \qwenicon~Qwen3-4B-Thinking-2507 consistently secure the top positions across all four datasets; specifically, the dominance of the 30B model is statistically distinguishable at the $95\%$ CI level in every case. Conversely, the relative ordering of the remaining models varies by dataset.

When incorporating $95\%$ CIs, we observe that while all four datasets exhibit five tied groups, the extent of ambiguity varies significantly. \aimefive{} yields the fewest distinct ranks (up to 11), followed by \aimefour{} (up to 13), and both \hmmt{} and \brumo{} (up to 14). This compression of ranks indicates greater uncertainty in the \unibayes{80} gold standard for \aimefive{} (due to more extensive ties) compared to the others under our current trial budget. Intuitively, this higher uncertainty in \aimefive{}'s gold-standard scores implies that more additional trials would be required for that dataset to empirically produce a statistically stable ranking; conversely, we can be more confident in the estimated gold standards for \aimefour{}, \hmmt{}, and \brumo{} given the current number of trials. This distinction also explains why \aimefive{} reaches a Kendall’s $\tau$ of 0.95 at $N=80$, whereas the other three datasets converge to $\sim 1$ at the same sample size in \Cref{fig:4_dataset_ranking}.

\begin{table}[h!]
\centering
\setlength{\tabcolsep}{4pt}
\caption{Rankings for four datasets. Models are listed in the order of their gold-standard ranking (\bayes{80} point estimates, i.e., without uncertainty) for \aimefive. Each dataset column gives the rank with a $95\%$ CI (left) and the rank without CI (right).}
\label{tab:dataset_ranking}
\begin{tabular}{lcccccccc}
\toprule
\multirow{2}{*}{\textbf{Model}} & \multicolumn{2}{c}{\textbf{\aimefive}} & \multicolumn{2}{c}{\textbf{\aimefour}} & \multicolumn{2}{c}{\textbf{\hmmt}} & \multicolumn{2}{c}{\textbf{\brumo}} \\
\cmidrule(lr){2-3} \cmidrule(lr){4-5} \cmidrule(lr){6-7} \cmidrule(lr){8-9}
& \rotatebox{0}{w/ CI} & \rotatebox{0}{w/o CI} & \rotatebox{0}{w/ CI} & \rotatebox{0}{w/o CI} & \rotatebox{0}{w/ CI} & \rotatebox{0}{w/o CI} & \rotatebox{0}{w/ CI} & \rotatebox{0}{w/o CI} \\
\midrule
\qwenicon~{}Qwen3-30B-A3B-Thinking-2507 & \rankcell{1}{1} & \rankcell{1}{1} & \rankcell{1}{1} & \rankcell{1}{1} & \rankcell{1}{1} & \rankcell{1}{1} & \rankcell{1}{1} & \rankcell{1}{1} \\
\qwenicon~{}Qwen3-4B-Thinking-2507 & \rankcell{2}{2} & \rankcell{2}{2} & \rankcell{2}{2} & \rankcell{2}{2} & \rankcell{2}{2} & \rankcell{2}{2} & \rankcell{2}{2} & \rankcell{2}{2} \\
\gpticon~{}gpt-oss-20b-high & \rankcell{3}{3} & \rankcell{3}{3} & \rankcell{3}{3} & \rankcell{5}{5} & \rankcell{3}{3} & \rankcell{4}{4} & \rankcell{6}{6} & \rankcell{11}{11} \\
\gpticon~{}gpt-oss-20b-medium & \rankcell{3}{3} & \rankcell{4}{4} & \rankcell{3}{3} & \rankcell{3}{3} & \rankcell{2}{2} & \rankcell{3}{3} & \rankcell{7}{7} & \rankcell{12}{12} \\
\microsoft~{}Phi-4-reasoning-plus & \rankcell{3}{3} & \rankcell{5}{5} & \rankcell{3}{3} & \rankcell{4}{4} & \rankcell{3}{3} & \rankcell{5}{5} & \rankcell{3}{3} & \rankcell{5}{5} \\
\nvidiaicon~{}AceReason-Nemotron-1.1-7B & \rankcell{4}{4} & \rankcell{6}{6} & \rankcell{5}{5} & \rankcell{9}{9} & \rankcell{4}{4} & \rankcell{6}{6} & \rankcell{3}{3} & \rankcell{4}{4} \\
\microsoft~{}Phi-4-reasoning & \rankcell{5}{5} & \rankcell{7}{7} & \rankcell{5}{5} & \rankcell{10}{10} & \rankcell{5}{5} & \rankcell{8}{8} & \rankcell{4}{4} & \rankcell{7}{7} \\
\gpticon~{}gpt-oss-20b-low & \rankcell{5}{5} & \rankcell{8}{8} & \rankcell{6}{6} & \rankcell{12}{12} & \rankcell{11}{11} & \rankcell{17}{17} & \rankcell{11}{11} & \rankcell{17}{17} \\
\openicon~{}OpenThinker2-32B & \rankcell{5}{5} & \rankcell{9}{9} & \rankcell{4}{4} & \rankcell{8}{8} & \rankcell{5}{5} & \rankcell{7}{7} & \rankcell{2}{2} & \rankcell{3}{3} \\
\qihoo~{}Light-R1-14B-DS & \rankcell{5}{5} & \rankcell{10}{10} & \rankcell{4}{4} & \rankcell{6}{6} & \rankcell{6}{6} & \rankcell{11}{11} & \rankcell{4}{4} & \rankcell{8}{8} \\
\fuse~{}FuseO1-DeepSeekR1-QwQ-SkyT1-Flash-32B & \rankcell{5}{5} & \rankcell{11}{11} & \rankcell{4}{4} & \rankcell{7}{7} & \rankcell{6}{6} & \rankcell{9}{9} & \rankcell{3}{3} & \rankcell{6}{6} \\
\nvidiaicon~{}NVIDIA-Nemotron-Nano-9B-v2 & \rankcell{6}{6} & \rankcell{12}{12} & \rankcell{6}{6} & \rankcell{11}{11} & \rankcell{6}{6} & \rankcell{10}{10} & \rankcell{5}{5} & \rankcell{10}{10} \\
\gairicon~{}LIMO-v2 & \rankcell{6}{6} & \rankcell{13}{13} & \rankcell{7}{7} & \rankcell{13}{13} & \rankcell{7}{7} & \rankcell{12}{12} & \rankcell{5}{5} & \rankcell{9}{9} \\
\lgicon~{}EXAONE-4.0-1.2B & \rankcell{7}{7} & \rankcell{14}{14} & \rankcell{8}{8} & \rankcell{14}{14} & \rankcell{7}{7} & \rankcell{13}{13} & \rankcell{10}{10} & \rankcell{15}{15} \\
\openr~{}OpenR1-Distill-7B & \rankcell{7}{7} & \rankcell{15}{15} & \rankcell{9}{9} & \rankcell{15}{15} & \rankcell{10}{10} & \rankcell{16}{16} & \rankcell{8}{8} & \rankcell{13}{13} \\
\openicon~{}OpenThinker3-1.5B & \rankcell{8}{8} & \rankcell{16}{16} & \rankcell{10}{10} & \rankcell{16}{16} & \rankcell{8}{8} & \rankcell{14}{14} & \rankcell{9}{9} & \rankcell{14}{14} \\
\nvidiaicon~{}OpenReasoning-Nemotron-1.5B & \rankcell{8}{8} & \rankcell{17}{17} & \rankcell{11}{11} & \rankcell{17}{17} & \rankcell{9}{9} & \rankcell{15}{15} & \rankcell{10}{10} & \rankcell{16}{16} \\
\dsicon~{}DeepSeek-R1-Distill-Qwen-1.5B & \rankcell{9}{9} & \rankcell{18}{18} & \rankcell{12}{12} & \rankcell{19}{19} & \rankcell{12}{12} & \rankcell{18}{18} & \rankcell{13}{13} & \rankcell{19}{19} \\
\skyicon~{}Sky-T1-32B-Flash & \rankcell{10}{10} & \rankcell{19}{19} & \rankcell{12}{12} & \rankcell{18}{18} & \rankcell{13}{13} & \rankcell{19}{19} & \rankcell{12}{12} & \rankcell{18}{18} \\
\bespoke~{}Bespoke-Stratos-7B & \rankcell{11}{11} & \rankcell{20}{20} & \rankcell{13}{13} & \rankcell{20}{20} & \rankcell{14}{14} & \rankcell{20}{20} & \rankcell{14}{14} & \rankcell{20}{20} \\
\bottomrule
\end{tabular}
\end{table}

\subsection{Convergence}
\label{sec:convergence}

In this section, we investigate the convergence of model rankings, building on the showcase figure (\Cref{fig:showcase}). We define convergence@$n$ as the smallest trial $n$ at which the ranking induced by the first $n$ trials matches the \emph{gold standard} ranking from all 80 trials (without bootstrapping) and remains unchanged thereafter.

Lower convergence@$n$ values indicate that fewer trials are sufficient to achieve stable rankings. As detailed in the caption of \Cref{fig:showcase}, the figure displays the probability mass functions (PMFs) of convergence@$n$ for each method across the datasets. These PMFs are empirically estimated by generating $10^5$ column-wise bootstrap replicates through resampling the $N_{\max}$ trials, then for each replicate, cumulatively evaluating the ranking at every $N$ (from 1 to 80) and identifying the minimal $n$ where the ranking stabilizes to the \emph{gold standard}. This process captures the distribution of convergence points under repeated sampling, reflecting the inherent uncertainty in finite-sample rankings due to stochastic trial outcomes.

This bootstrapping approach provides a distribution over possible convergence points ($n$), offering insights into the variability and reliability of each evaluation method: \pass{k} (for $k=2, 4, 8$) versus our \unibayes{N}. A lower mean convergence@$n$ signifies more cost-effective convergence, while failure to converge within $80$ trials (as seen in \aimefive) indicates more trials are needed to \emph{confidently} rank LLMs or we must include CI for a reliable ranking.

The key takeaways from \Cref{fig:showcase}, as summarized in its caption, highlight the advantages of \unibayes{N}: it converges reliably on all datasets except \aimefive, often with fewer trials than \pass{k}. For instance, on \hmmt{} and \brumo{}, \bayes{N} achieves mean convergence at approximately $44.2$ and $27.1$ trials, respectively, compared to around $69.5$ and $48.5$ for the best-performing \pass{k} scores. The right panel of the figure further illustrates this through an example ranking from a bootstrap replicate, emphasizing differences in convergence for \aimefive{} and \brumo{}. See Appendix~\ref{app:conv} (\Cref{fig:app:cdf}) for the corresponding cumulative distribution functions (CDFs).

\paragraph{Worst-case scenarios}\label{par:wc}To further distinguish the \bayes{N} framework from \avg{N}, we analyze the \emph{worst-case} bootstrap replicates, i.e., those that either require the maximum number of trials to stabilize the rankings or fail to converge. For 11 LLMs, \Cref{fig:combined_rankings_1} shows these trajectories as competition rankings, with each line tracing a model’s rank as trials accumulate; convergence is defined as the point at which the ranking order remains unchanged for all subsequent trials. In \aimefour{} the ranking converges at trial~75, in \hmmt{} at trial~78, and in \brumo{} at trial~68, whereas in \aimefive{} no convergence is observed within 80 trials, underscoring persistent instability and the need for additional trials or \bayes{N}'s credible intervals. When a ranking does not converge within the trial budget (as for \aimefive{} in \Cref{fig:showcase}) only \bayes{N} can be used to quantify uncertainty and estimate the minimum \(N\) required for a reliable ranking (see \Cref{sec:sim_ranking_variance}).

\begin{figure}
    \centering
    \includegraphics[width=1\linewidth]{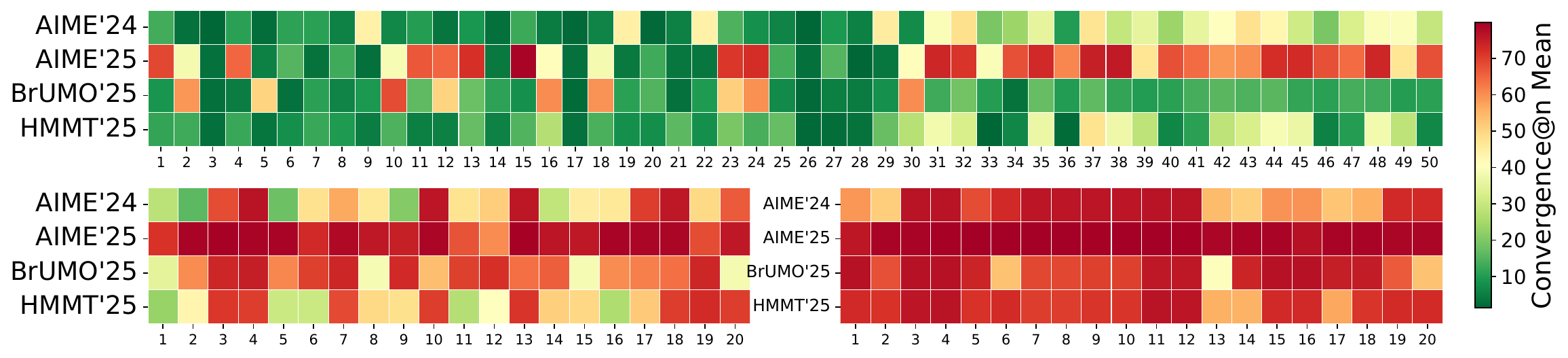}
    \caption{\textbf{Convergence@$n$ without CI}. Mean convergence@$n$ across model combinations for \aimefour{}, \aimefive{}, \hmmt{}, and \brumo{}. \textbf{Top}: 50 combinations of 5 models. \textbf{Bottom-left}: 20 combinations of 10 models. \textbf{Bottom-right}: 20 combinations of 15 models. Color indicates the mean convergence@$n$ over $10^5$ bootstrap replicates (green: fast convergence; red: slow convergence).}
    \label{fig:conv-comb}
\end{figure}

This situation becomes even more severe as more models are included. As shown in \Cref{fig:combined_rankings_2}, when the number of models is increased to \(L = 20\), none of the datasets exhibit convergence. To examine convergence as a function of \(L\) more systematically, we consider a pool of 20 LLMs (\Cref{app:tab:modelid}) and construct 50 subsets of 5 models (\Cref{app:tab:comb5}), 20 subsets of 10 models (\Cref{app:tab:comb10}), and 20 subsets of 15 models (\Cref{app:tab:comb15}). For each subset, we generate \(10^{5}\) bootstrap replicates to estimate convergence@$n$. \Cref{fig:conv-comb} reports the resulting convergence@$n$ values across all subsets and replicates, showing that as the number of models increases, evaluation methods such as \avg{N} and the \pass{k} family become unreliable for estimating model abilities and producing stable rankings.

\subsection{Rubric-Aware Categorical Evaluation}
\label{sec:exp:cat}

While evaluation is often reduced to binary correctness, this simplification discards valuable signals that capture other aspects of model behavior. For instance, LLM outputs can be assessed not only on correctness but also on whether they are well-structured, coherent, or exhibit step-by-step reasoning in mathematical tasks. In practice, evaluators could record richer dimensions such as format compliance, calibration of confidence, degenerate outputs, out-of-distribution (OOD) behavior, and verifier scores. This limitation is especially important for reasoning models, where overthinking ~\cite{chen2024not} inflates token usage without corresponding gains in reliability. \bayes{N} provides a principled way to capture these richer outcomes. By treating per-item results as categorical rather than binary, the approach aligns more closely with actual goals while preserving statistical rigor and transparency.

Concretely, for each question and trial, a set of \emph{base signals} is logged (e.g., correctness, presence of a boxed final answer, length, and perplexity features). These base signals are then augmented with probabilistic labels from a lightweight reward model (e.g., calibrated probabilities of correct, wrong, or off-task) \cite{CompassVerifier}. From these signals, \emph{rubric variables} are defined (Table~\ref{app:tab:var}) and different \emph{categorical schemata} are instantiated (Table~\ref{tab:cat_schema}), mapping each attempt into one of \(C{+}1\) categories. Under \bayes{N}, the resulting category counts induce a Dirichlet posterior, and a rubric is specified by the weight vector \(w\) in \cref{summary}. Different choices of schema and \(w\) encode different evaluation preferences (e.g., stricter compliance, stronger penalties for confidently wrong answers, or efficiency-adjusted scoring). This procedure yields posterior means and credible intervals for each rubric.

\cref{fig:cat:rank} summarizes aggregated results across tasks. The leader \qwenicon~Qwen3-30B-A3B-Thinking-2507 ranks first under all selected schemata, but the gap to rank~2 depends on the rubric: it is largest under \emph{Conf-Wrong Penalty} and smallest under \emph{Verifier-Only}. Mid-pack reorderings are rubric-sensitive: under \emph{Verifier Prob}, \openicon~\emph{OpenThinker2-32B} edges \gpticon~\emph{gpt-oss-20b\_medium}; under calibration-heavy schemata (e.g., \emph{Conf-Calibrated} and \emph{Format+Confidence}), \gpticon~\emph{gpt-oss-20b\_high} overtakes \openicon~\emph{OpenThinker2-32B}; and \emph{OOD Robustness} narrows the gap between ranks~2 and~3. Several categories (\emph{Format Aware}, \emph{Length-Robust}, and \emph{Strict Compliance}) agree closely, indicating that once correctness is accounted for, formatting and length rarely flip the top ranks. In contrast, calibration-focused categories emphasize and penalize confidently wrong behavior, and efficiency-oriented categories favor concision. The lower tier is stable across categories (\lgicon~\emph{EXAONE-4.0-1.2B}, \openicon~\emph{OpenThinker3-1.5B}, \nvidiaicon~\emph{OpenReasoning-Nemotron-1.5B}, \skyicon~\emph{Sky-T1-32B-Flash}, \dsicon~\emph{DeepSeek-R1-Distill-Qwen-1.5B}), suggesting rubric choice primarily reshuffles the middle while preserving extremes. Overall, the categorical schemata surface complementary facets---format compliance, calibration, efficiency, OOD robustness, and verifier alignment---making rubric-dependent differences explicit and enabling compute-efficient, uncertainty-aware comparisons aligned with evaluation goals. For a comprehensive discussion of the categorical Bayesian evaluation framework, including base signals, schema definitions, and their impact on model rankings, see Appendix~\ref{app:sec:cat}.

\begin{table}[t]
\centering
\setlength{\tabcolsep}{4pt}
\small
\begin{threeparttable}
\caption{Comparison of the Bayesian framework and other evaluation methods.}
\label{tab:eval-compare}
\begin{tabular}{lcccc}
\toprule
\textbf{Methods ($N$ trials)} & \textbf{Convergence} & \textbf{Credible interval} & \textbf{Prior knowledge} & \textbf{Categorical} \\
\midrule
\pass{k} and alternatives & \no & \no & \no & \no \\
\avg{N} & \yes & Limited (via bootstrap/binomial CIs) & \no & \no \\
\bayes{N} & \yes~(Sec. \ref{sec:convergence}, \cref{fig:showcase,fig:conv-comb}) & \yes~(\cref{fig:CI_sim}, \cref{tab:sim_ranking,tab:dataset_ranking}) & \yes~(Sec. \ref{app:non_uni_prior}) & \yes~(Sec. \ref{sec:exp:cat}) \\
\bottomrule
\end{tabular}
\end{threeparttable}
\end{table}

\section{Related Work}
Functional-correctness evaluation with \pass{k} became standard in code generation with HumanEval (OpenAI Codex): generate $k$ samples, a task is solved if any sample passes unit tests, and estimate the overall rate with an unbiased estimator that requires producing $n \ge k$ samples per task \cite{chen2021evaluating}. Although \pass{k} was initially introduced in the context of coding, it later became the de facto choice to evaluate LLMs not only on math reasoning tasks~\cite{guo2025deepseek,shao2024deepseekmath,tong2024dartmath,liu2023tinygsmachieving80gsm8k,hwang2024selfexplore,yang2024weaktostrong,muennighoff2025s1simpletesttimescaling,chen2025rethinkingttc,liu2024your,lg2025exaone} but also on safety evaluations spanning agent red-teaming, jailbreaks, and backdoor analyses~\cite{nakash2025redteaming,aghakhani2024trojanpuzzle,liu2024loratk,yan2024codebreaker,rtlbreaker2024,schaeffer2025powerlaws}. For a broader review of these metrics and their variants, see Appendix~\ref{app:extended_related_work}. Beyond standard \pass{k}, \emph{pass\textasciicircum$k$} quantifies reliability across $k$ i.i.d. trials for agents, while the generalized \emph{G-pass@k$_{\tau}$} continuum (and its area-under-$\tau$ summary \emph{mG-Pass}) jointly assess potential and stability in reasoning outputs~\cite{yao2024taubench,liu2024your}.

Efforts like HELM advance holistic, transparent evaluation across scenarios and metrics \cite{liang2022helm}, while practice guidelines distill reproducibility pitfalls and prescribe multi-run, uncertainty-aware reporting with fixed prompts, decoding, and dataset \cite{biderman2024lessons}. The LM Evaluation Harness offers standardized, reproducible frameworks to implement these recommendations \cite{biderman2024lessons}. It supports uncertainty reporting through binomial-style uncertainty estimates for binary mean metrics and bootstrap estimates for others.

The last category of related work focuses on measuring uncertainty in LLM evaluation. These works converge on interval-aware, small-sample-valid reporting rather than CLT/Wald error bars. Bowyer et al. show that CLT-based intervals \emph{miscalibrate} on small benchmarks and advocate small-$n$-appropriate frequentist or Bayesian intervals for reliable comparisons~\cite{bowyer2025position}. A Bayesian alternative models capability as a latent success probability and reports posterior uncertainty that remains informative with limited trials, yielding more stable rankings~\cite{xiao2025confidence}. In judge-based settings, \emph{Judging LLMs on a Simplex} places model and judge behavior on the probability simplex, enabling uncertainty-aware comparisons and highlighting how distributional structure matters for evaluation~\cite{vossler2025judging}. Beyond bespoke LLM metrics, prediction-powered inference supplies general procedures for valid confidence intervals that leverage model predictions to reduce labeled-sample requirements~\cite{angelopoulos2023prediction}. Finally, in adjacent retrieval evaluation with LLM-generated assessments, Oosterhuis et al. construct reliable confidence intervals and demonstrate that calibrated uncertainty, rather than point estimates, should guide decisions, reinforcing this shift for LLM evaluation more broadly~\cite{oosterhuis2024reliable}.
\section{Conclusion: Strengths, Limitations \& Future Directions}

The overall benefits of the Bayesian framework are summarized in Table~\ref{tab:eval-compare}: it provides fast convergence, analytical uncertainty estimates, and the incorporation of prior knowledge and categorical results. However, it is worth noting that our approach quantifies \emph{statistical} uncertainty from finite samples; it does not fix dataset bias, distribution shift, or rubric misspecification. Results therefore depend on the chosen benchmark, prompts, and inference settings (hardware).  Although we have validated our approach with biased-coin LLM mimic simulations, together with experiments using actual LLMs (up to $N_\text{max} = 80$ trials across four tasks and 20 models), more extensive evaluations may be constrained by computing and academic budgets.

The focus of the current work was the simplest version of the Bayesian approach, using a uniform prior, which provides a conservative and reproducible starting point. But the theory allows for more complex, informative priors, and this opens up a rich vein of future directions that should be systematically explored: for example priors from past runs, domain- or task-conditioned priors, and expert-elicited priors. These have the potential of accelerating convergence even further, but must be chosen and reported carefully.  Clear guidance and tools for prior elicitation will hopefully ensure that gains in sample efficiency do not come at the cost of hidden bias.

\newpage

\section*{Ethics Statement}
This research relies only on publicly available, non-personal benchmarks; no human subjects, user data, or PII are involved. Potential misuse includes cherry-picking priors, rubrics, or samples to exaggerate performance. To prevent this, use of \bayes{N} with user-defined priors requires clear documentation and reporting of posterior credible intervals.

\section*{Reproducibility Statement}
To ensure reproducibility, detailed implementation instructions are provided in Appendix~\ref{app:experiment}.
\section*{Acknowledgments}
This research was supported in part by NSF awards 2117439, 2112606, and 2320952.
\bibliography{iclr2026_conference}

\begin{thebibliography}{94}
\providecommand{\natexlab}[1]{#1}
\providecommand{\url}[1]{\texttt{#1}}
\expandafter\ifx\csname urlstyle\endcsname\relax
  \providecommand{\doi}[1]{doi: #1}\else
  \providecommand{\doi}{doi: \begingroup \urlstyle{rm}\Url}\fi

\bibitem[Vaswani et~al.(2017)Vaswani, Shazeer, Parmar, Uszkoreit, Jones, Gomez, Kaiser, and Polosukhin]{vaswani2017attention}
Ashish Vaswani, Noam Shazeer, Niki Parmar, Jakob Uszkoreit, Llion Jones, Aidan~N. Gomez, Lukasz Kaiser, and Illia Polosukhin.
\newblock Attention is all you need.
\newblock In \emph{Advances in Neural Information Processing Systems}, 2017.
\newblock URL \url{https://arxiv.org/abs/1706.03762}.

\bibitem[Brown et~al.(2020)Brown, Mann, Ryder, Subbiah, Kaplan, Dhariwal, Neelakantan, Shyam, Sastry, Askell, Agarwal, Herbert-Voss, Krueger, Henighan, Child, Ramesh, Ziegler, Wu, Winter, Hesse, Chen, Sigler, Litwin, Gray, Chess, Clark, Berner, McCandlish, Radford, Sutskever, and Amodei]{brown2020gpt3}
Tom Brown, Benjamin Mann, Nick Ryder, Melanie Subbiah, Jared~D. Kaplan, Prafulla Dhariwal, Arvind Neelakantan, Pranav Shyam, Girish Sastry, Amanda Askell, Sandhini Agarwal, Ariel Herbert-Voss, Gretchen Krueger, Tom Henighan, Rewon Child, Aditya Ramesh, Daniel Ziegler, Jeffrey Wu, Clemens Winter, Chris Hesse, Mark Chen, Eric Sigler, Mateusz Litwin, Scott Gray, Benjamin Chess, Jack Clark, Christopher Berner, Sam McCandlish, Alec Radford, Ilya Sutskever, and Dario Amodei.
\newblock Language models are few-shot learners.
\newblock In \emph{Advances in Neural Information Processing Systems}, volume~33, pages 1877--1901. Curran Associates, Inc., 2020.
\newblock URL \url{https://proceedings.neurips.cc/paper/2020/file/1457c0d6bfcb4967418bfb8ac142f64a-Paper.pdf}.

\bibitem[{Stack Overflow}(2025)]{stackoverflow2025AI}
{Stack Overflow}.
\newblock {Stack Overflow Developer Survey 2025: AI and Developer Tools}, 2025.
\newblock URL \url{https://survey.stackoverflow.co/2025/ai}.
\newblock Accessed: 2025-09-24.

\bibitem[Maslej et~al.(2025)Maslej, Fattorini, Perrault, Gil, Parli, Kariuki, Capstick, Reuel, Brynjolfsson, Etchemendy, Ligett, Lyons, Manyika, Niebles, Shoham, Wald, Walsh, Hamrah, Santarlasci, Betts~Lotufo, Rome, Shi, and Oak]{maslej2025artificial}
Nestor Maslej, Loredana Fattorini, Raymond Perrault, Yolanda Gil, Vanessa Parli, Njenga Kariuki, Emily Capstick, Anka Reuel, Erik Brynjolfsson, John Etchemendy, Katrina Ligett, Terah Lyons, James Manyika, Juan~Carlos Niebles, Yoav Shoham, Russell Wald, Toby Walsh, Armin Hamrah, Lapo Santarlasci, Julia Betts~Lotufo, Alexandra Rome, Andrew Shi, and Sukrut Oak.
\newblock Artificial intelligence index report 2025.
\newblock \emph{arXiv preprint arXiv:2504.07139}, 2025.
\newblock URL \url{https://arxiv.org/abs/2504.07139}.

\bibitem[Liang et~al.(2022)Liang, Bommasani, Lee, Tsipras, Soylu, Yasunaga, Zhang, Narayanan, Wu, Kumar, et~al.]{liang2022helm}
Percy Liang, Rishi Bommasani, Tony Lee, Dimitris Tsipras, Dilara Soylu, Michihiro Yasunaga, Yian Zhang, Deepak Narayanan, Yuhuai Wu, Ananya Kumar, et~al.
\newblock Holistic evaluation of language models.
\newblock \emph{arXiv preprint arXiv:2211.09110}, 2022.
\newblock URL \url{https://arxiv.org/abs/2211.09110}.

\bibitem[Hendrycks et~al.(2021{\natexlab{a}})Hendrycks, Burns, Basart, Zou, Mazeika, Song, and Steinhardt]{hendrycks2021mmlu}
Dan Hendrycks, Collin Burns, Steven Basart, Andy Zou, Mantas Mazeika, Dawn Song, and Jacob Steinhardt.
\newblock Measuring massive multitask language understanding.
\newblock In \emph{International Conference on Learning Representations (ICLR)}, 2021{\natexlab{a}}.
\newblock URL \url{https://arxiv.org/abs/2009.03300}.

\bibitem[Srivastava et~al.(2022)Srivastava, Rastogi, Rao, Shoeb, Abid, Fisch, Brown, Santoro, Gupta, Garriga-Alonso, et~al.]{srivastava2022bigbench}
Aarohi Srivastava, Abhinav Rastogi, Abhishek Rao, Abu Awal~Md Shoeb, Abubakar Abid, Adam Fisch, Adam~R. Brown, Adam Santoro, Aditya Gupta, Adri{\`a} Garriga-Alonso, et~al.
\newblock {Beyond the Imitation Game: Quantifying and Extrapolating the Capabilities of Language Models (BIG-bench)}.
\newblock \emph{arXiv preprint arXiv:2206.04615}, 2022.
\newblock URL \url{https://arxiv.org/abs/2206.04615}.

\bibitem[Kaplan et~al.(2020)Kaplan, McCandlish, Henighan, Brown, Chess, Child, Gray, Radford, Wu, and Amodei]{kaplan2020scaling}
Jared Kaplan, Sam McCandlish, Tom Henighan, Tom~B. Brown, Benjamin Chess, Rewon Child, Scott Gray, Alec Radford, Jeffrey Wu, and Dario Amodei.
\newblock Scaling laws for neural language models.
\newblock \emph{arXiv preprint arXiv:2001.08361}, 2020.
\newblock URL \url{https://arxiv.org/abs/2001.08361}.

\bibitem[Hoffmann et~al.(2022)Hoffmann, Borgeaud, Mensch, Buchatskaya, Cai, Rutherford, de~Las~Casas, Hendricks, Welbl, Clark, Hennigan, Noland, Millican, van~den Driessche, Damoc, Guy, Osindero, Simonyan, Elsen, Rae, Vinyals, and Sifre]{hoffmann2022chinchilla}
Jordan Hoffmann, Sebastian Borgeaud, Arthur Mensch, Elena Buchatskaya, Trevor Cai, Eliza Rutherford, Diego de~Las~Casas, Lisa~Anne Hendricks, Johannes Welbl, Aidan Clark, Tom Hennigan, Eric Noland, Katie Millican, George van~den Driessche, Bogdan Damoc, Aurelia Guy, Simon Osindero, Karen Simonyan, Erich Elsen, Jack~W. Rae, Oriol Vinyals, and Laurent Sifre.
\newblock Training compute-optimal large language models.
\newblock \emph{arXiv preprint arXiv:2203.15556}, 2022.
\newblock URL \url{https://arxiv.org/abs/2203.15556}.

\bibitem[Wei et~al.(2022)Wei, Wang, Schuurmans, Bosma, Ichter, Xia, Chi, Le, and Zhou]{wei2022cot}
Jason Wei, Xuezhi Wang, Dale Schuurmans, Maarten Bosma, Brian Ichter, Fei Xia, Ed~H. Chi, Quoc~V. Le, and Denny Zhou.
\newblock Chain-of-thought prompting elicits reasoning in large language models.
\newblock In \emph{Advances in Neural Information Processing Systems}, 2022.
\newblock URL \url{https://openreview.net/forum?id=\%5FVjQlMeSB\%5FJ}.

\bibitem[Ouyang et~al.(2022)Ouyang, Wu, Jiang, Almeida, Wainwright, Mishkin, Zhang, Agarwal, Slama, Ray, Schulman, Hilton, Kelton, Miller, Simens, Askell, Welinder, Christiano, Leike, and Lowe]{ouyang2022instructgpt}
Long Ouyang, Jeffrey Wu, Xu~Jiang, Diogo Almeida, Carroll Wainwright, Pamela Mishkin, Chong Zhang, Sandhini Agarwal, Katarina Slama, Alex Ray, John Schulman, Jacob Hilton, Fraser Kelton, Luke Miller, Maddie Simens, Amanda Askell, Peter Welinder, Paul~F. Christiano, Jan Leike, and Ryan Lowe.
\newblock Training language models to follow instructions with human feedback.
\newblock In \emph{Advances in Neural Information Processing Systems}, volume~35, pages 27730--27744. Curran Associates, Inc., 2022.
\newblock URL \url{https://proceedings.neurips.cc/paper\%5Ffiles/paper/2022/file/b1efde53be364a73914f58805a001731-Paper-Conference.pdf}.

\bibitem[Dettmers et~al.(2022)Dettmers, Lewis, Belkada, and Zettlemoyer]{Dettmers2022LLMint8}
Tim Dettmers, Mike Lewis, Younes Belkada, and Luke Zettlemoyer.
\newblock {LLM.int8(): 8-bit Matrix Multiplication for Transformers at Scale}, 2022.
\newblock URL \url{https://arxiv.org/abs/2208.07339}.

\bibitem[Frantar et~al.(2022)Frantar, Ashkboos, Hoefler, and Alistarh]{Frantar2022GPTQ}
Elias Frantar, Saleh Ashkboos, Torsten Hoefler, and Dan Alistarh.
\newblock {GPTQ}: Accurate post-training quantization for generative pre-trained transformers, 2022.
\newblock URL \url{https://arxiv.org/abs/2210.17323}.

\bibitem[Han et~al.(2015)Han, Pool, Tran, and Dally]{Han2015Pruning}
Song Han, Jeff Pool, John Tran, and William Dally.
\newblock Learning both weights and connections for efficient neural networks.
\newblock In \emph{NeurIPS}, 2015.
\newblock URL \url{https://papers.nips.cc/paper/5784-learning-both-weights-and-connections-for-efficient-neural-network}.

\bibitem[Hinton et~al.(2015)Hinton, Vinyals, and Dean]{Hinton2015Distillation}
Geoffrey Hinton, Oriol Vinyals, and Jeff Dean.
\newblock Distilling the knowledge in a neural network, 2015.
\newblock URL \url{https://arxiv.org/abs/1503.02531}.

\bibitem[Kwon et~al.(2023)Kwon, Li, Zhuang, Sheng, Zheng, Yu, Gonzalez, Zhang, and Stoica]{kwon2023efficient}
Woosuk Kwon, Zhuohan Li, Siyuan Zhuang, Ying Sheng, Lianmin Zheng, Cody~Hao Yu, Joseph Gonzalez, Hao Zhang, and Ion Stoica.
\newblock Efficient memory management for large language model serving with pagedattention.
\newblock In \emph{Proceedings of the 29th symposium on operating systems principles}, pages 611--626, 2023.
\newblock URL \url{https://arxiv.org/abs/2309.06180}.

\bibitem[Zhang et~al.(2024)Zhang, Yi, Xu, and Shrivastava]{Zhang2024KV1Bit}
Tianyi Zhang, Jonah Yi, Zhaozhuo Xu, and Anshumali Shrivastava.
\newblock Kv cache is 1 bit per channel: Efficient large language model inference with coupled quantization.
\newblock In A.~Globerson, L.~Mackey, D.~Belgrave, A.~Fan, U.~Paquet, J.~Tomczak, and C.~Zhang, editors, \emph{Advances in Neural Information Processing Systems}, volume~37, pages 3304--3331. Curran Associates, Inc., 2024.
\newblock \doi{10.52202/079017-0109}.
\newblock URL \url{https://proceedings.neurips.cc/paper\%5Ffiles/paper/2024/file/05d6b5b6901fb57d2c287e1d3ce6d63c-Paper-Conference.pdf}.

\bibitem[Zhang et~al.(2025{\natexlab{a}})Zhang, Ji, Chen, Fu, Miao, Nie, Chen, and Cui]{Zhang2024PQCache}
Hailin Zhang, Xiaodong Ji, Yilin Chen, Fangcheng Fu, Xupeng Miao, Xiaonan Nie, Weipeng Chen, and Bin Cui.
\newblock Pqcache: Product quantization-based kvcache for long context llm inference.
\newblock \emph{Proc. ACM Manag. Data}, 3\penalty0 (3), June 2025{\natexlab{a}}.
\newblock \doi{10.1145/3725338}.
\newblock URL \url{https://doi.org/10.1145/3725338}.

\bibitem[Hariri et~al.(2026{\natexlab{a}})Hariri, Luo, Chen, Zhong, Zhang, Wang, Hu, Han, and Chaudhary]{hariri2026quantize}
Mohsen Hariri, Alan Luo, Weicong Chen, Shaochen Zhong, Tianyi Zhang, Qifan Wang, Xia Hu, Xiaotian Han, and Vipin Chaudhary.
\newblock Quantize what counts: More for keys, less for values.
\newblock In \emph{The 64th Annual Meeting of the Association for Computational Linguistics (Findings)}, 2026{\natexlab{a}}.
\newblock \doi{10.48550/arXiv.2502.15075}.
\newblock URL \url{https://openreview.net/forum?id=vMIlB97WV1}.

\bibitem[Hu et~al.(2021)Hu, Shen, Wallis, Allen-Zhu, Li, Wang, Wang, and Chen]{Hu2021LoRA}
Edward~J. Hu, Yelong Shen, Phillip Wallis, Zeyuan Allen-Zhu, Yuanzhi Li, Shean Wang, Lu~Wang, and Weizhu Chen.
\newblock {LoRA: Low-Rank Adaptation of Large Language Models}.
\newblock \emph{arXiv preprint arXiv:2106.09685}, 2021.
\newblock URL \url{https://arxiv.org/abs/2106.09685}.

\bibitem[Christiano et~al.(2017)Christiano, Leike, Brown, Martic, Legg, and Amodei]{Christiano2017Prefs}
Paul~F. Christiano, Jan Leike, Tom~B. Brown, Miljan Martic, Shane Legg, and Dario Amodei.
\newblock Deep reinforcement learning from human preferences.
\newblock In \emph{NeurIPS}, 2017.
\newblock URL \url{https://arxiv.org/abs/1706.03741}.

\bibitem[Holtzman et~al.(2020)Holtzman, Buys, Du, Forbes, and Choi]{Holtzman2019Degeneration}
Ari Holtzman, Jan Buys, Li~Du, Maxwell Forbes, and Yejin Choi.
\newblock The curious case of neural text degeneration.
\newblock In \emph{ICLR}, 2020.
\newblock URL \url{https://openreview.net/forum?id=rygGQyrFvH}.
\newblock arXiv:1904.09751 (2019).

\bibitem[Dao(2023)]{Dao2022FlashAttention}
Tri Dao.
\newblock Flashattention-2: Faster attention with better parallelism and work partitioning.
\newblock \emph{arXiv preprint arXiv:2307.08691}, 2023.
\newblock URL \url{https://arxiv.org/abs/2307.08691}.

\bibitem[Chen et~al.(2021)Chen, Tworek, Jun, Yuan, de~Oliveira~Pinto, Kaplan, Edwards, Burda, Joseph, Brockman, Ray, Puri, Krueger, Petrov, Khlaaf, Sastry, Mishkin, Chan, Gray, Ryder, Pavlov, Power, Kaiser, Bavarian, Winter, Tillet, Such, Cummings, Plappert, Chantzis, Barnes, Herbert-Voss, Guss, Nichol, Paino, Tezak, Tang, Babuschkin, Balaji, Jain, Saunders, Hesse, Carr, Leike, Achiam, Misra, Morikawa, Radford, Knight, Brundage, Murati, Mayer, Welinder, McGrew, Amodei, McCandlish, Sutskever, and Zaremba]{chen2021evaluating}
Mark Chen, Jerry Tworek, Heewoo Jun, Qiming Yuan, Henrique~Ponde de~Oliveira~Pinto, Jared Kaplan, Harri Edwards, Yuri Burda, Nicholas Joseph, Greg Brockman, Alex Ray, Raul Puri, Gretchen Krueger, Michael Petrov, Heidy Khlaaf, Girish Sastry, Pamela Mishkin, Brooke Chan, Scott Gray, Nick Ryder, Mikhail Pavlov, Alethea Power, Lukasz Kaiser, Mohammad Bavarian, Clemens Winter, Philippe Tillet, Felipe~Petroski Such, Dave Cummings, Matthias Plappert, Fotios Chantzis, Elizabeth Barnes, Ariel Herbert-Voss, William~Hebgen Guss, Alex Nichol, Alex Paino, Nikolas Tezak, Jie Tang, Igor Babuschkin, Suchir Balaji, Shantanu Jain, William Saunders, Christopher Hesse, Andrew~N. Carr, Jan Leike, Josh Achiam, Vedant Misra, Evan Morikawa, Alec Radford, Matthew Knight, Miles Brundage, Mira Murati, Katie Mayer, Peter Welinder, Bob McGrew, Dario Amodei, Sam McCandlish, Ilya Sutskever, and Wojciech Zaremba.
\newblock Evaluating large language models trained on code.
\newblock \emph{arXiv preprint arXiv:2107.03374}, 2021.
\newblock URL \url{https://arxiv.org/abs/2107.03374}.

\bibitem[{Mathematical Association of America}(2024)]{MAA_AIME2024}
{Mathematical Association of America}.
\newblock American invitational mathematics examination (aime).
\newblock \url{https://maa.org/maa-invitational-competitions/}, 2024.
\newblock URL \url{https://maa.org/maa-invitational-competitions/}.
\newblock Official MAA page for the AIME competition (covers AIME 2024).

\bibitem[{Mathematical Association of America}(2025)]{MAA_AIME2025}
{Mathematical Association of America}.
\newblock American invitational mathematics examination (aime).
\newblock \url{https://maa.org/maa-invitational-competitions/}, 2025.
\newblock URL \url{https://maa.org/maa-invitational-competitions/}.
\newblock Official MAA page for the AIME competition (covers AIME 2025).

\bibitem[Liu et~al.(2024{\natexlab{a}})Liu, Liu, Xiao, Wang, Liu, Gao, Zhang, Zhang, and Chen]{liu2024your}
Junnan Liu, Hongwei Liu, Linchen Xiao, Ziyi Wang, Kuikun Liu, Songyang Gao, Wenwei Zhang, Songyang Zhang, and Kai Chen.
\newblock Are your llms capable of stable reasoning?
\newblock \emph{arXiv preprint arXiv:2412.13147}, December 2024{\natexlab{a}}.
\newblock \doi{10.48550/arXiv.2412.13147}.
\newblock URL \url{https://arxiv.org/abs/2412.13147}.
\newblock ACL 2025 camera-ready version.

\bibitem[Hochlehnert et~al.(2025)Hochlehnert, Bhatnagar, Udandarao, Albanie, Prabhu, and Bethge]{hochlehnert2025sober}
Andreas Hochlehnert, Hardik Bhatnagar, Vishaal Udandarao, Samuel Albanie, Ameya Prabhu, and Matthias Bethge.
\newblock A sober look at progress in language model reasoning: Pitfalls and paths to reproducibility.
\newblock \emph{arXiv preprint arXiv:2504.07086}, 2025.
\newblock URL \url{https://arxiv.org/abs/2504.07086}.

\bibitem[Dror et~al.(2018)Dror, Baumer, Shlomov, and Reichart]{Dror2018Significance}
Rotem Dror, Gili Baumer, Segev Shlomov, and Roi Reichart.
\newblock The hitchhiker's guide to testing statistical significance in natural language processing.
\newblock In \emph{Proceedings of the 56th Annual Meeting of the Association for Computational Linguistics (Volume 1: Long Papers)}, pages 1383--1392, Melbourne, Australia, July 2018. Association for Computational Linguistics.
\newblock \doi{10.18653/v1/P18-1128}.
\newblock URL \url{https://aclanthology.org/P18-1128/}.

\bibitem[Yeh(2000)]{Yeh2000Significance}
Alexander Yeh.
\newblock More accurate tests for the statistical significance of result differences.
\newblock In \emph{COLING}, 2000.
\newblock URL \url{https://aclanthology.org/C00-2137/}.

\bibitem[Dodge et~al.(2019)Dodge, Gururangan, Card, Schwartz, and Smith]{Dodge2019ShowYourWork}
Jesse Dodge, Suchin Gururangan, Dallas Card, Roy Schwartz, and Noah~A. Smith.
\newblock Show your work: Improved reporting of experimental results.
\newblock In \emph{Proceedings of the 2019 Conference on Empirical Methods in Natural Language Processing and the 9th International Joint Conference on Natural Language Processing (EMNLP-IJCNLP)}, pages 2185--2194, Hong Kong, China, November 2019. Association for Computational Linguistics.
\newblock \doi{10.18653/v1/D19-1224}.
\newblock URL \url{https://aclanthology.org/D19-1224/}.

\bibitem[Zheng et~al.(2023)Zheng, Chiang, Sheng, Zhuang, Wu, Zhuang, Lin, Li, Li, Xing, Zhang, Gonzalez, and Stoica]{Zheng2023LLMJudge}
Lianmin Zheng, Wei-Lin Chiang, Ying Sheng, Siyuan Zhuang, Zhanghao Wu, Yonghao Zhuang, Zi~Lin, Zhuohan Li, Dacheng Li, Eric~P. Xing, Hao Zhang, Joseph~E. Gonzalez, and Ion Stoica.
\newblock {Judging LLM-as-a-Judge with MT-Bench and Chatbot Arena}.
\newblock \emph{arXiv preprint arXiv:2306.05685}, 2023.
\newblock URL \url{https://arxiv.org/abs/2306.05685}.

\bibitem[Chen et~al.(2024{\natexlab{a}})Chen, Chen, Liu, Jiang, and Wang]{Chen2024LLMJudgeBias}
Guiming~Hardy Chen, Shunian Chen, Ziche Liu, Feng Jiang, and Benyou Wang.
\newblock Humans or {LLM}s as the judge? a study on judgement bias.
\newblock In \emph{Proceedings of the 2024 Conference on Empirical Methods in Natural Language Processing}, pages 8301--8327, Miami, Florida, USA, November 2024{\natexlab{a}}. Association for Computational Linguistics.
\newblock \doi{10.18653/v1/2024.emnlp-main.474}.
\newblock URL \url{https://aclanthology.org/2024.emnlp-main.474/}.

\bibitem[Xiao et~al.(2025)Xiao, Su, Zhang, Chen, Chen, and Liu]{xiao2025confidence}
Xiao Xiao, Yu~Su, Sijing Zhang, Zhang Chen, Yadong Chen, and Tian Liu.
\newblock Confidence in large language model evaluation: A bayesian approach to limited-sample challenges.
\newblock \emph{arXiv preprint arXiv:2504.21303}, 2025.
\newblock URL \url{https://arxiv.org/abs/2504.21303}.

\bibitem[Hayden and Armitage(2025)]{hayden2025straightforward}
Dustin Hayden and Thomas Armitage.
\newblock Straightforward bayesian a/b testing with dirichlet posteriors.
\newblock \emph{arXiv preprint arXiv:2508.08077}, 2025.
\newblock URL \url{https://arxiv.org/abs/2508.08077}.

\bibitem[{Harvard--MIT Mathematics Tournament}(2025)]{HMMT_Feb2025}
{Harvard--MIT Mathematics Tournament}.
\newblock Hmmt february 2025 archive (problems and solutions).
\newblock \url{https://www.hmmt.org/www/archive/282}, 2025.
\newblock URL \url{https://www.hmmt.org/www/archive/282}.
\newblock Official HMMT archive page for February 2025 competition.

\bibitem[{Brown University Math Olympiad Organizers}(2025)]{BrUMO_2025}
{Brown University Math Olympiad Organizers}.
\newblock Brown university math olympiad (brumo).
\newblock \url{https://www.brumo.org/tournament-info}, 2025.
\newblock URL \url{https://www.brumo.org/tournament-info}.
\newblock Official BrUMO website with tournament information (Apr 4--5, 2025).

\bibitem[Dalal et~al.(2025)Dalal, Segal, Ben-Haim, Lahav, and Nevo]{dalal2025leveraging}
Uri Dalal, Meirav Segal, Zvika Ben-Haim, Dan Lahav, and Omer Nevo.
\newblock {Leveraging LLM Inconsistency to Boost Pass@ k Performance}.
\newblock \emph{arXiv preprint arXiv:2505.12938}, 2025.
\newblock URL \url{https://arxiv.org/abs/2505.12938}.

\bibitem[Ross et~al.(2025)Ross, Vouitsis, Ghomi, Hosseinzadeh, Xin, Liu, Sui, Hou, Leung, Loaiza-Ganem, and Cresswell]{ross2025textual}
Brendan~Leigh Ross, No{\"e}l Vouitsis, Atiyeh~Ashari Ghomi, Rasa Hosseinzadeh, Ji~Xin, Zhaoyan Liu, Yi~Sui, Shiyi Hou, Kin~Kwan Leung, Gabriel Loaiza-Ganem, and Jesse~C. Cresswell.
\newblock {Textual Bayes: Quantifying Prompt Uncertainty in LLM-Based Systems}.
\newblock \emph{arXiv preprint arXiv:2506.10060}, 2025.
\newblock URL \url{https://arxiv.org/abs/2506.10060}.

\bibitem[Vashurin et~al.(2025)Vashurin, Goloburda, Ilina, Rubashevskii, Nakov, Shelmanov, and Panov]{vashurin2025uncertainty}
Roman Vashurin, Maiya Goloburda, Albina Ilina, Aleksandr Rubashevskii, Preslav Nakov, Artem Shelmanov, and Maxim Panov.
\newblock {Uncertainty Quantification for LLMs through Minimum Bayes Risk: Bridging Confidence and Consistency}.
\newblock \emph{arXiv preprint arXiv:2502.04964}, 2025.
\newblock URL \url{https://arxiv.org/abs/2502.04964}.

\bibitem[Jaynes(2003)]{jaynes2003probability}
Edwin~T Jaynes.
\newblock \emph{Probability Theory: The Logic of Science}.
\newblock Cambridge University Press, 2003.
\newblock \doi{10.1017/CBO9780511790423}.
\newblock URL \url{https://doi.org/10.1017/CBO9780511790423}.

\bibitem[Bowyer et~al.(2025)Bowyer, Aitchison, and Ivanova]{bowyer2025position}
Sam Bowyer, Laurence Aitchison, and Desi~R Ivanova.
\newblock {Position: Don't Use the CLT in LLM Evals With Fewer Than a Few Hundred Datapoints}.
\newblock \emph{arXiv preprint arXiv:2503.01747}, 2025.
\newblock URL \url{https://arxiv.org/abs/2503.01747}.

\bibitem[Hariri et~al.(2026{\natexlab{b}})Hariri, Hinczewski, Ma, and Chaudhary]{hariri2026ranking}
Mohsen Hariri, Michael Hinczewski, Jing Ma, and Vipin Chaudhary.
\newblock Ranking reasoning {LLM}s under test-time scaling.
\newblock In \emph{The 64th Annual Meeting of the Association for Computational Linguistics}, 2026{\natexlab{b}}.
\newblock \doi{10.48550/arXiv.2603.10960}.
\newblock URL \url{https://openreview.net/forum?id=DjRkQvirQL}.

\bibitem[Chen et~al.(2024{\natexlab{b}})Chen, Xu, Liang, He, Pang, Yu, Song, Liu, Zhou, Zhang, Wang, Tu, Mi, and Yu]{chen2024not}
Xingyu Chen, Jiahao Xu, Tian Liang, Zhiwei He, Jianhui Pang, Dian Yu, Linfeng Song, Qiuzhi Liu, Mengfei Zhou, Zhuosheng Zhang, Rui Wang, Zhaopeng Tu, Haitao Mi, and Dong Yu.
\newblock Do {NOT} think that much for 2+3=? on the overthinking of o1-like {LLMs}.
\newblock \emph{arXiv preprint arXiv:2412.21187}, 2024{\natexlab{b}}.
\newblock URL \url{https://arxiv.org/abs/2412.21187}.

\bibitem[Liu et~al.(2025{\natexlab{a}})Liu, Liu, Liu, Xiao, Gao, Lyu, Gu, Zhang, Wong, Zhang, and Chen]{CompassVerifier}
Shudong Liu, Hongwei Liu, Junnan Liu, Linchen Xiao, Songyang Gao, Chengqi Lyu, Yuzhe Gu, Wenwei Zhang, Derek~F. Wong, Songyang Zhang, and Kai Chen.
\newblock Compassverifier: A unified and robust verifier for llms evaluation and outcome reward.
\newblock \emph{arXiv preprint arXiv:2508.03686}, 2025{\natexlab{a}}.
\newblock URL \url{https://arxiv.org/abs/2508.03686}.

\bibitem[Guo et~al.(2025)Guo, Yang, Zhang, Song, Zhang, Xu, Zhu, Ma, Wang, Bi, et~al.]{guo2025deepseek}
Daya Guo, Dejian Yang, Haowei Zhang, Junxiao Song, Ruoyu Zhang, Runxin Xu, Qihao Zhu, Shirong Ma, Peiyi Wang, Xiao Bi, et~al.
\newblock {DeepSeek-R1}: Incentivizing reasoning capability in {LLMs} via reinforcement learning.
\newblock \emph{arXiv preprint arXiv:2501.12948}, 2025.
\newblock URL \url{https://arxiv.org/abs/2501.12948}.

\bibitem[Shao et~al.(2024)Shao, Wang, Zhu, Xu, Song, Bi, Zhang, Zhang, Li, Wu, and Guo]{shao2024deepseekmath}
Zhihong Shao, Peiyi Wang, Qihao Zhu, Runxin Xu, Junxiao Song, Xiao Bi, Haowei Zhang, Mingchuan Zhang, Y.~K. Li, Y.~Wu, and Daya Guo.
\newblock {DeepSeekMath}: Pushing the limits of mathematical reasoning in open language models.
\newblock \emph{arXiv preprint arXiv:2402.03300}, 2024.
\newblock URL \url{https://arxiv.org/abs/2402.03300}.

\bibitem[Tong et~al.(2024)Tong, Zhang, Wang, Wu, and He]{tong2024dartmath}
Yuxuan Tong, Xiwen Zhang, Rui Wang, Ruidong Wu, and Junxian He.
\newblock Dart-math: Difficulty-aware rejection tuning for mathematical problem-solving.
\newblock In A.~Globerson, L.~Mackey, D.~Belgrave, A.~Fan, U.~Paquet, J.~Tomczak, and C.~Zhang, editors, \emph{Advances in Neural Information Processing Systems}, volume~37, pages 7821--7846. Curran Associates, Inc., 2024.
\newblock \doi{10.52202/079017-0251}.
\newblock URL \url{https://proceedings.neurips.cc/paper\%5Ffiles/paper/2024/file/0ef1afa0daa888d695dcd5e9513bafa3-Paper-Conference.pdf}.

\bibitem[Liu et~al.(2023)Liu, Bubeck, Eldan, Kulkarni, Li, Nguyen, Ward, and Zhang]{liu2023tinygsmachieving80gsm8k}
Bingbin Liu, Sebastien Bubeck, Ronen Eldan, Janardhan Kulkarni, Yuanzhi Li, Anh Nguyen, Rachel Ward, and Yi~Zhang.
\newblock {TinyGSM}: Achieving 80\% on {GSM8K} with small language models, 2023.
\newblock URL \url{https://arxiv.org/abs/2312.09241}.

\bibitem[Hwang et~al.(2024)Hwang, Kim, Kim, Ye, and Seo]{hwang2024selfexplore}
Hyeonbin Hwang, Doyoung Kim, Seungone Kim, Seonghyeon Ye, and Minjoon Seo.
\newblock Self-explore: Enhancing mathematical reasoning in language models with fine-grained rewards.
\newblock In \emph{Findings of the Association for Computational Linguistics: EMNLP 2024}, pages 1444--1466, Miami, Florida, USA, November 2024. Association for Computational Linguistics.
\newblock \doi{10.18653/v1/2024.findings-emnlp.78}.
\newblock URL \url{https://aclanthology.org/2024.findings-emnlp.78/}.

\bibitem[Yang et~al.(2024)Yang, Ma, and Liu]{yang2024weaktostrong}
Yuqing Yang, Yan Ma, and Pengfei Liu.
\newblock Weak-to-strong reasoning.
\newblock In \emph{Findings of the Association for Computational Linguistics: EMNLP 2024}, pages 8350--8367, Miami, Florida, USA, November 2024. Association for Computational Linguistics.
\newblock \doi{10.18653/v1/2024.findings-emnlp.490}.
\newblock URL \url{https://aclanthology.org/2024.findings-emnlp.490/}.

\bibitem[Muennighoff et~al.(2025)Muennighoff, Yang, Shi, Li, Fei-Fei, Hajishirzi, Zettlemoyer, Liang, Cand{\`e}s, and Hashimoto]{muennighoff2025s1simpletesttimescaling}
Niklas Muennighoff, Zitong Yang, Weijia Shi, Xiang~Lisa Li, Li~Fei-Fei, Hannaneh Hajishirzi, Luke Zettlemoyer, Percy Liang, Emmanuel Cand{\`e}s, and Tatsunori Hashimoto.
\newblock s1: Simple test-time scaling, 2025.
\newblock URL \url{https://arxiv.org/abs/2501.19393}.

\bibitem[Chen et~al.(2025)Chen, Ravent{\'o}s, Cheng, Ganguli, and Druckmann]{chen2025rethinkingttc}
Feng Chen, Allan Ravent{\'o}s, Nan Cheng, Surya Ganguli, and Shaul Druckmann.
\newblock Rethinking fine-tuning when scaling test-time compute: Limiting confidence improves mathematical reasoning, 2025.
\newblock URL \url{https://arxiv.org/abs/2502.07154}.

\bibitem[Bae et~al.(2025{\natexlab{a}})Bae, Choi, Choi, Choi, Choi, Hong, Hwang, Jeon, Jeon, Jo, Jo, Jung, Kim, Kim, Kim, Kim, Kim, Kim, Kim, Kim, Lee, Lee, Lee, Lee, Lee, Park, Park, Yang, Yeen, Yi, and Yun]{lg2025exaone}
Kyunghoon Bae, Eunbi Choi, Kibong Choi, Stanley~Jungkyu Choi, Yemuk Choi, Seokhee Hong, Junwon Hwang, Hyojin Jeon, Kijeong Jeon, Gerrard~Jeongwon Jo, Hyunjik Jo, Jiyeon Jung, Hyosang Kim, Joonkee Kim, Seonghwan Kim, Soyeon Kim, Sunkyoung Kim, Yireun Kim, Yongil Kim, Youchul Kim, Edward~Hwayoung Lee, Haeju Lee, Honglak Lee, Jinsik Lee, Kyungmin Lee, Sangha Park, Yongmin Park, Sihoon Yang, Heuiyeen Yeen, Sihyuk Yi, and Hyeongu Yun.
\newblock {EXAONE} deep: Reasoning enhanced language models.
\newblock \emph{arXiv preprint arXiv:2503.12524}, March 2025{\natexlab{a}}.
\newblock \doi{10.48550/arXiv.2503.12524}.
\newblock URL \url{https://arxiv.org/abs/2503.12524}.

\bibitem[Nakash et~al.(2025)Nakash, Kour, Lazar, Vetzler, Uziel, and Tavor]{nakash2025redteaming}
Itay Nakash, George Kour, Koren Lazar, Matan Vetzler, Guy Uziel, and Ateret~Anaby Tavor.
\newblock Effective red-teaming of policy-adherent agents.
\newblock In \emph{Proceedings of the 2025 Conference on Empirical Methods in Natural Language Processing}, pages 2250--2268, Suzhou, China, November 2025. Association for Computational Linguistics.
\newblock ISBN 979-8-89176-332-6.
\newblock \doi{10.18653/v1/2025.emnlp-main.114}.
\newblock URL \url{https://aclanthology.org/2025.emnlp-main.114/}.

\bibitem[Aghakhani et~al.(2024)Aghakhani, Dai, Manoel, Fernandes, Kharkar, Kruegel, Vigna, Evans, Zorn, and Sim]{aghakhani2024trojanpuzzle}
Hojjat Aghakhani, Wei Dai, Andre Manoel, Xavier Fernandes, Anant Kharkar, Christopher Kruegel, Giovanni Vigna, David Evans, Ben Zorn, and Robert Sim.
\newblock {TrojanPuzzle}: Covertly poisoning code-suggestion models.
\newblock In \emph{{IEEE} Symposium on Security and Privacy, {SP} 2024, San Francisco, CA, USA, May 19--23, 2024}, pages 1122--1140. {IEEE}, 2024.
\newblock \doi{10.1109/SP54263.2024.00140}.
\newblock URL \url{https://doi.org/10.1109/SP54263.2024.00140}.

\bibitem[Liu et~al.(2024{\natexlab{b}})Liu, Zhong, Sun, Tian, Hariri, Liu, Tang, Jiang, Yuan, Chuang, Li, Choi, Chen, Chaudhary, and Hu]{liu2024loratk}
Hongyi Liu, Shaochen Zhong, Xintong Sun, Minghao Tian, Mohsen Hariri, Zirui Liu, Ruixiang Tang, Zhimeng Jiang, Jiayi Yuan, Yu-Neng Chuang, Li~Li, Soo-Hyun Choi, Rui Chen, Vipin Chaudhary, and Xia Hu.
\newblock {LoRATK: LoRA Once, Backdoor Everywhere in the Share-and-Play Ecosystem}.
\newblock \emph{arXiv preprint arXiv:2403.00108}, 2024{\natexlab{b}}.
\newblock URL \url{https://arxiv.org/abs/2403.00108}.

\bibitem[Yan et~al.(2024)Yan, Wang, Duan, Hong, Lee, Kim, and Hong]{yan2024codebreaker}
Shenao Yan, Shen Wang, Yue Duan, Hanbin Hong, Kiho Lee, Doowon Kim, and Yuan Hong.
\newblock An {LLM-Assisted} {Easy-to-Trigger} backdoor attack on code completion models: Injecting disguised vulnerabilities against strong detection.
\newblock In \emph{33rd USENIX Security Symposium (USENIX Security 24)}, pages 1795--1812, Philadelphia, PA, August 2024. USENIX Association.
\newblock ISBN 978-1-939133-44-1.
\newblock URL \url{https://www.usenix.org/conference/usenixsecurity24/presentation/yan}.

\bibitem[Mankali et~al.(2024)Mankali, Bhandari, Alam, Karri, Maniatakos, Sinanoglu, and Knechtel]{rtlbreaker2024}
Lakshmi~Likhitha Mankali, Jitendra Bhandari, Manaar Alam, Ramesh Karri, Michail Maniatakos, Ozgur Sinanoglu, and Johann Knechtel.
\newblock {{RTL}-Breaker}: Assessing the security of {LLM}s against backdoor attacks on {HDL} code generation.
\newblock \emph{arXiv preprint arXiv:2411.17569}, November 2024.
\newblock \doi{10.48550/arXiv.2411.17569}.
\newblock URL \url{https://arxiv.org/abs/2411.17569}.
\newblock Accepted at DATE 2025.

\bibitem[Schaeffer et~al.(2025)Schaeffer, Kazdan, Hughes, Juravsky, Price, Lynch, Jones, Kirk, Mirhoseini, and Koyejo]{schaeffer2025powerlaws}
Rylan Schaeffer, Joshua Kazdan, John Hughes, Jordan Juravsky, Sara Price, Aengus Lynch, Erik Jones, Robert Kirk, Azalia Mirhoseini, and Sanmi Koyejo.
\newblock How do large language monkeys get their power ({L}aws)?
\newblock In Aarti Singh, Maryam Fazel, Daniel Hsu, Simon Lacoste-Julien, Felix Berkenkamp, Tegan Maharaj, Kiri Wagstaff, and Jerry Zhu, editors, \emph{Proceedings of the 42nd International Conference on Machine Learning}, volume 267 of \emph{Proceedings of Machine Learning Research}, pages 53132--53176. PMLR, 13--19 Jul 2025.
\newblock \doi{10.48550/arXiv.2502.17578}.
\newblock URL \url{https://proceedings.mlr.press/v267/schaeffer25a.html}.
\newblock Oral.

\bibitem[Yao et~al.(2024)Yao, Shinn, Razavi, and Narasimhan]{yao2024taubench}
Shunyu Yao, Noah Shinn, Pedram Razavi, and Karthik Narasimhan.
\newblock ${\tau}$-bench: A benchmark for tool-agent-user interaction in real-world domains.
\newblock \emph{arXiv preprint}, 2024.
\newblock \doi{10.48550/arXiv.2406.12045}.
\newblock URL \url{https://doi.org/10.48550/arXiv.2406.12045}.
\newblock Introduces the pass$^{k}$ metric.

\bibitem[Biderman et~al.(2024)Biderman, Schoelkopf, Sutawika, Gao, Tow, Abbasi, Aji, Ammanamanchi, Black, Clive, DiPofi, Etxaniz, Fattori, Forde, Foster, Hsu, Jaiswal, Lee, Li, Lovering, Muennighoff, Pavlick, Phang, Skowron, Tan, Tang, Wang, Winata, Yvon, and Zou]{biderman2024lessons}
Stella Biderman, Hailey Schoelkopf, Lintang Sutawika, Leo Gao, Jonathan Tow, Baber Abbasi, Alham~Fikri Aji, Pawan~Sasanka Ammanamanchi, Sidney Black, Jordan Clive, Anthony DiPofi, Julen Etxaniz, Benjamin Fattori, Jessica~Zosa Forde, Charles Foster, Jeffrey Hsu, Mimansa Jaiswal, Wilson~Y. Lee, Haonan Li, Charles Lovering, Niklas Muennighoff, Ellie Pavlick, Jason Phang, Aviya Skowron, Samson Tan, Xiangru Tang, Kevin~A. Wang, Genta~Indra Winata, Fran\c{c}ois Yvon, and Andy Zou.
\newblock Lessons from the trenches on reproducible evaluation of language models.
\newblock \emph{arXiv preprint arXiv:2405.14782}, 2024.
\newblock URL \url{https://arxiv.org/abs/2405.14782}.

\bibitem[Vossler et~al.(2025)Vossler, Xia, Mai, Subbaswamy, and Feng]{vossler2025judging}
Patrick Vossler, Fan Xia, Yifan Mai, Adarsh Subbaswamy, and Jean Feng.
\newblock {LLMs Judging LLMs: A Simplex Perspective}.
\newblock \emph{arXiv preprint arXiv:2505.21972}, 2025.
\newblock URL \url{https://arxiv.org/abs/2505.21972}.

\bibitem[Angelopoulos et~al.(2023)Angelopoulos, Bates, Fannjiang, Jordan, and Zrnic]{angelopoulos2023prediction}
Anastasios~N Angelopoulos, Stephen Bates, Clara Fannjiang, Michael~I Jordan, and Tijana Zrnic.
\newblock Prediction-powered inference.
\newblock \emph{Science}, 382\penalty0 (6671):\penalty0 669--674, 2023.
\newblock \doi{10.1126/science.adi6000}.
\newblock URL \url{https://www.science.org/doi/10.1126/science.adi6000}.

\bibitem[Oosterhuis et~al.(2024)Oosterhuis, Jagerman, Qin, Wang, and Bendersky]{oosterhuis2024reliable}
Harrie Oosterhuis, Rolf Jagerman, Zhen Qin, Xuanhui Wang, and Michael Bendersky.
\newblock Reliable confidence intervals for information retrieval evaluation using generative a.i.
\newblock In \emph{Proceedings of the 30th ACM SIGKDD Conference on Knowledge Discovery and Data Mining}, pages 2307--2317. Association for Computing Machinery, 2024.
\newblock \doi{10.1145/3637528.3671883}.
\newblock URL \url{https://doi.org/10.1145/3637528.3671883}.

\bibitem[Wolf et~al.(2019)Wolf, Debut, Sanh, Chaumond, Delangue, Moi, Cistac, Rault, Louf, Funtowicz, Davison, Shleifer, von Platen, Ma, Jernite, Plu, Xu, Le~Scao, Gugger, Drame, Lhoest, and Rush]{wolf2019huggingface}
Thomas Wolf, Lysandre Debut, Victor Sanh, Julien Chaumond, Clement Delangue, Anthony Moi, Pierric Cistac, Tim Rault, R{\'e}mi Louf, Morgan Funtowicz, Joe Davison, Sam Shleifer, Patrick von Platen, Clara Ma, Yacine Jernite, Julien Plu, Canwen Xu, Teven Le~Scao, Sylvain Gugger, Mariama Drame, Quentin Lhoest, and Alexander~M. Rush.
\newblock {HuggingFace}'s transformers: State-of-the-art natural language processing.
\newblock \emph{arXiv preprint arXiv:1910.03771}, 2019.
\newblock URL \url{https://arxiv.org/abs/1910.03771}.

\bibitem[Brown et~al.(2024)Brown, Juravsky, Ehrlich, Clark, Le, R{\'e}, and Mirhoseini]{brown2024large}
Bradley Brown, Jordan Juravsky, Ryan Ehrlich, Ronald Clark, Quoc~V Le, Christopher R{\'e}, and Azalia Mirhoseini.
\newblock Large language monkeys: Scaling inference compute with repeated sampling.
\newblock \emph{arXiv preprint arXiv:2407.21787}, 2024.
\newblock URL \url{https://arxiv.org/abs/2407.21787}.

\bibitem[Papineni et~al.(2002)Papineni, Roukos, Ward, and Zhu]{papineni-etal-2002-bleu}
Kishore Papineni, Salim Roukos, Todd Ward, and Wei-Jing Zhu.
\newblock {B}leu: a method for automatic evaluation of machine translation.
\newblock In Pierre Isabelle, Eugene Charniak, and Dekang Lin, editors, \emph{Proceedings of the 40th Annual Meeting of the Association for Computational Linguistics}, pages 311--318, Philadelphia, Pennsylvania, USA, July 2002. Association for Computational Linguistics.
\newblock \doi{10.3115/1073083.1073135}.
\newblock URL \url{https://aclanthology.org/P02-1040/}.

\bibitem[Ren et~al.(2020)Ren, Guo, Lu, Zhou, Liu, Tang, Sundaresan, Zhou, Blanco, and Ma]{ren2020codebleu}
Shuo Ren, Daya Guo, Shuai Lu, Long Zhou, Shujie Liu, Duyu Tang, Neel Sundaresan, Ming Zhou, Ambrosio Blanco, and Shuai Ma.
\newblock {CodeBLEU}: A method for automatic evaluation of code synthesis.
\newblock \emph{arXiv preprint arXiv:2009.10297}, 2020.
\newblock URL \url{https://arxiv.org/abs/2009.10297}.

\bibitem[Kulal et~al.(2019)Kulal, Pasupat, Chandra, Lee, Padon, Aiken, and Liang]{kulal2019spoc}
Sumith Kulal, Panupong Pasupat, Kartik Chandra, Mina Lee, Oded Padon, Alex Aiken, and Percy~S Liang.
\newblock {SPoC}: Search-based pseudocode to code.
\newblock \emph{Advances in Neural Information Processing Systems}, 32, 2019.
\newblock URL \url{https://arxiv.org/abs/1906.04908}.

\bibitem[Hendrycks et~al.(2021{\natexlab{b}})Hendrycks, Burns, Kadavath, Arora, Basart, Tang, Song, and Steinhardt]{hendrycks2021measuring}
Dan Hendrycks, Collin Burns, Saurav Kadavath, Akul Arora, Steven Basart, Eric Tang, Dawn Song, and Jacob Steinhardt.
\newblock Measuring mathematical problem solving with the {MATH} dataset.
\newblock \emph{arXiv preprint arXiv:2103.03874}, 2021{\natexlab{b}}.
\newblock URL \url{https://arxiv.org/abs/2103.03874}.

\bibitem[Cobbe et~al.(2021)Cobbe, Kosaraju, Bavarian, Chen, Jun, Kaiser, Plappert, Tworek, Hilton, Nakano, Hesse, and Schulman]{cobbe2021training}
Karl Cobbe, Vineet Kosaraju, Mohammad Bavarian, Mark Chen, Heewoo Jun, Lukasz Kaiser, Matthias Plappert, Jerry Tworek, Jacob Hilton, Reiichiro Nakano, Christopher Hesse, and John Schulman.
\newblock Training verifiers to solve math word problems.
\newblock \emph{arXiv preprint arXiv:2110.14168}, 2021.
\newblock URL \url{https://arxiv.org/abs/2110.14168}.

\bibitem[Wang et~al.(2022)Wang, Wei, Schuurmans, Le, Chi, Narang, Chowdhery, and Zhou]{wang2022self}
Xuezhi Wang, Jason Wei, Dale Schuurmans, Quoc Le, Ed~Chi, Sharan Narang, Aakanksha Chowdhery, and Denny Zhou.
\newblock Self-consistency improves chain of thought reasoning in language models.
\newblock \emph{arXiv preprint arXiv:2203.11171}, 2022.
\newblock URL \url{https://arxiv.org/abs/2203.11171}.

\bibitem[Lewkowycz et~al.(2022)Lewkowycz, Andreassen, Dohan, Dyer, Michalewski, Ramasesh, Slone, Anil, Schlag, Gutman-Solo, Wu, Neyshabur, Gur-Ari, and Misra]{lewkowycz2022solving}
Aitor Lewkowycz, Anders Andreassen, David Dohan, Ethan Dyer, Henryk Michalewski, Vinay Ramasesh, Ambrose Slone, Cem Anil, Imanol Schlag, Theo Gutman-Solo, Yuhuai Wu, Behnam Neyshabur, Guy Gur-Ari, and Vedant Misra.
\newblock Solving quantitative reasoning problems with language models.
\newblock \emph{Advances in Neural Information Processing Systems}, 35:\penalty0 3843--3857, 2022.
\newblock URL \url{https://proceedings.neurips.cc/paper\%5Ffiles/paper/2022/file/18abbeef8cfe9203fdf9053c9c4fe191-Paper-Conference.pdf}.

\bibitem[KENDALL(1938)]{10.1093/biomet/30.1-2.81}
M.~G. KENDALL.
\newblock A new measure of rank correlation.
\newblock \emph{Biometrika}, 30\penalty0 (1-2):\penalty0 81--93, 06 1938.
\newblock ISSN 0006-3444.
\newblock \doi{10.1093/biomet/30.1-2.81}.
\newblock URL \url{https://doi.org/10.1093/biomet/30.1-2.81}.

\bibitem[Team(2025)]{reduce_overthinking_2025}
NovaSky Team.
\newblock Think less, achieve more: Cut reasoning costs by 50
\newblock https://novasky-ai.github.io/posts/reduce-overthinking, 2025.
\newblock URL \url{https://novasky-ai.github.io/posts/reduce-overthinking}.
\newblock Accessed: 2025-01-23.

\bibitem[Yang et~al.(2025)Yang, Li, Yang, Zhang, Hui, Zheng, Yu, Gao, Huang, Lv, Zheng, Liu, Zhou, Huang, Hu, Ge, Wei, Lin, Tang, Yang, Tu, Zhang, Yang, Yang, Zhou, Zhou, Lin, Dang, Bao, Yang, Yu, Deng, Li, Xue, Li, Zhang, Wang, Zhu, Men, Gao, Liu, Luo, Li, Tang, Yin, Ren, Wang, Zhang, Ren, Fan, Su, Zhang, Zhang, Wan, Liu, Wang, Cui, Zhang, Zhou, and Qiu]{qwen3technicalreport}
An~Yang, Anfeng Li, Baosong Yang, Beichen Zhang, Binyuan Hui, Bo~Zheng, Bowen Yu, Chang Gao, Chengen Huang, Chenxu Lv, Chujie Zheng, Dayiheng Liu, Fan Zhou, Fei Huang, Feng Hu, Hao Ge, Haoran Wei, Huan Lin, Jialong Tang, Jian Yang, Jianhong Tu, Jianwei Zhang, Jianxin Yang, Jiaxi Yang, Jing Zhou, Jingren Zhou, Junyang Lin, Kai Dang, Keqin Bao, Kexin Yang, Le~Yu, Lianghao Deng, Mei Li, Mingfeng Xue, Mingze Li, Pei Zhang, Peng Wang, Qin Zhu, Rui Men, Ruize Gao, Shixuan Liu, Shuang Luo, Tianhao Li, Tianyi Tang, Wenbiao Yin, Xingzhang Ren, Xinyu Wang, Xinyu Zhang, Xuancheng Ren, Yang Fan, Yang Su, Yichang Zhang, Yinger Zhang, Yu~Wan, Yuqiong Liu, Zekun Wang, Zeyu Cui, Zhenru Zhang, Zhipeng Zhou, and Zihan Qiu.
\newblock Qwen3 technical report, 2025.
\newblock URL \url{https://arxiv.org/abs/2505.09388}.

\bibitem[OpenAI(2025)]{openai2025gptoss120bgptoss20bmodel}
OpenAI.
\newblock gpt-oss-120b \& gpt-oss-20b model card, 2025.
\newblock URL \url{https://arxiv.org/abs/2508.10925}.

\bibitem[Ye et~al.(2025)Ye, Huang, Xiao, Chern, Xia, and Liu]{ye2025limoreasoning}
Yixin Ye, Zhen Huang, Yang Xiao, Ethan Chern, Shijie Xia, and Pengfei Liu.
\newblock {LIMO: Less is More for Reasoning}, 2025.
\newblock URL \url{https://arxiv.org/abs/2502.03387}.

\bibitem[Bae et~al.(2025{\natexlab{b}})Bae, Choi, Choi, Choi, Choi, Han, Hong, Hwang, Hwang, Jang, Jeon, Jeon, Jo, Jo, Jung, Kim, Kim, Kim, Kim, Kim, Kim, Kim, Kim, Kim, Kim, Lee, Lee, Lee, Lee, Lee, Lee, Park, Paik, Park, Park, Seo, Yang, Yeen, Yi, and Yun]{exaone-4.0}
Kyunghoon Bae, Eunbi Choi, Kibong Choi, Stanley~Jungkyu Choi, Yemuk Choi, Kyubeen Han, Seokhee Hong, Junwon Hwang, Taewan Hwang, Joonwon Jang, Hyojin Jeon, Kijeong Jeon, Gerrard~Jeongwon Jo, Hyunjik Jo, Jiyeon Jung, Euisoon Kim, Hyosang Kim, Jihoon Kim, Joonkee Kim, Seonghwan Kim, Soyeon Kim, Sunkyoung Kim, Yireun Kim, Yongil Kim, Youchul Kim, Edward~Hwayoung Lee, Gwangho Lee, Haeju Lee, Honglak Lee, Jinsik Lee, Kyungmin Lee, Sangha Park, Young~Min Paik, Yongmin Park, Youngyong Park, Sanghyun Seo, Sihoon Yang, Heuiyeen Yeen, Sihyuk Yi, and Hyeongu Yun.
\newblock Exaone 4.0: Unified large language models integrating non-reasoning and reasoning modes.
\newblock \emph{arXiv preprint arXiv:2507.11407}, 2025{\natexlab{b}}.
\newblock URL \url{https://arxiv.org/abs/2507.11407}.

\bibitem[Toshniwal et~al.(2025)Toshniwal, Sorokin, Ficek, Moshkov, and Gitman]{toshniwal2025genselect}
Shubham Toshniwal, Ivan Sorokin, Aleksander Ficek, Ivan Moshkov, and Igor Gitman.
\newblock {GenSelect: A Generative Approach to Best-of-N}.
\newblock In \emph{2nd AI for Math Workshop @ ICML 2025}, 2025.
\newblock URL \url{https://openreview.net/forum?id=8LhnmNmUDb}.

\bibitem[Moshkov et~al.(2025)Moshkov, Hanley, Sorokin, Toshniwal, Henkel, Schifferer, Du, and Gitman]{moshkov2025aimo2winningsolutionbuilding}
Ivan Moshkov, Darragh Hanley, Ivan Sorokin, Shubham Toshniwal, Christof Henkel, Benedikt Schifferer, Wei Du, and Igor Gitman.
\newblock {AIMO-2 Winning Solution: Building State-of-the-Art Mathematical Reasoning Models with OpenMathReasoning dataset}, 2025.
\newblock URL \url{https://arxiv.org/abs/2504.16891}.

\bibitem[Ahmad et~al.(2025{\natexlab{a}})Ahmad, Majumdar, Ficek, Narenthiran, Samadi, Huang, Jain, Noroozi, and Ginsburg]{ahmad2025opencodereasoningiisimpletesttime}
Wasi~Uddin Ahmad, Somshubra Majumdar, Aleksander Ficek, Sean Narenthiran, Mehrzad Samadi, Jocelyn Huang, Siddhartha Jain, Vahid Noroozi, and Boris Ginsburg.
\newblock {OpenCodeReasoning-II: A Simple Test Time Scaling Approach via Self-Critique}, 2025{\natexlab{a}}.
\newblock URL \url{https://arxiv.org/abs/2507.09075}.

\bibitem[Ahmad et~al.(2025{\natexlab{b}})Ahmad, Narenthiran, Majumdar, Ficek, Jain, Huang, Noroozi, and Ginsburg]{ahmad2025opencodereasoning}
Wasi~Uddin Ahmad, Sean Narenthiran, Somshubra Majumdar, Aleksander Ficek, Siddhartha Jain, Jocelyn Huang, Vahid Noroozi, and Boris Ginsburg.
\newblock {OpenCodeReasoning: Advancing Data Distillation for Competitive Coding}.
\newblock \emph{arXiv preprint arXiv:2504.01943}, 2025{\natexlab{b}}.
\newblock URL \url{https://arxiv.org/abs/2504.01943}.

\bibitem[Guha et~al.(2025)Guha, Marten, Keh, Raoof, Smyrnis, Bansal, Nezhurina, Mercat, Vu, Sprague, Suvarna, Feuer, Chen, Khan, Frankel, Grover, Choi, Muennighoff, Su, Zhao, Yang, Pimpalgaonkar, Sharma, Ji, Deng, Pratt, Ramanujan, Saad-Falcon, Li, Dave, Albalak, Arora, Wulfe, Hegde, Durrett, Oh, Bansal, Gabriel, Grover, Chang, Shankar, Gokaslan, Merrill, Hashimoto, Choi, Jitsev, Heckel, Sathiamoorthy, Dimakis, and Schmidt]{guha2025openthoughtsdatarecipesreasoning}
Etash Guha, Ryan Marten, Sedrick Keh, Negin Raoof, Georgios Smyrnis, Hritik Bansal, Marianna Nezhurina, Jean Mercat, Trung Vu, Zayne Sprague, Ashima Suvarna, Benjamin Feuer, Liangyu Chen, Zaid Khan, Eric Frankel, Sachin Grover, Caroline Choi, Niklas Muennighoff, Shiye Su, Wanjia Zhao, John Yang, Shreyas Pimpalgaonkar, Kartik Sharma, Charlie Cheng-Jie Ji, Yichuan Deng, Sarah Pratt, Vivek Ramanujan, Jon Saad-Falcon, Jeffrey Li, Achal Dave, Alon Albalak, Kushal Arora, Blake Wulfe, Chinmay Hegde, Greg Durrett, Sewoong Oh, Mohit Bansal, Saadia Gabriel, Aditya Grover, Kai-Wei Chang, Vaishaal Shankar, Aaron Gokaslan, Mike~A. Merrill, Tatsunori Hashimoto, Yejin Choi, Jenia Jitsev, Reinhard Heckel, Maheswaran Sathiamoorthy, Alexandros~G. Dimakis, and Ludwig Schmidt.
\newblock {OpenThoughts: Data Recipes for Reasoning Models}, 2025.
\newblock URL \url{https://arxiv.org/abs/2506.04178}.

\bibitem[Abdin et~al.(2025)Abdin, Agarwal, Awadallah, Balachandran, Behl, Chen, de~Rosa, Gunasekar, Javaheripi, Joshi, Kauffmann, Lara, Mendes, Mitra, Nushi, Papailiopoulos, Saarikivi, Shah, Shrivastava, Vineet, Wu, Yousefi, and Zheng]{abdin2025phi4reasoning}
Marah Abdin, Sahaj Agarwal, Ahmed Awadallah, Vidhisha Balachandran, Harkirat Behl, Lingjiao Chen, Gustavo de~Rosa, Suriya Gunasekar, Mojan Javaheripi, Neel Joshi, Piero Kauffmann, Yash Lara, Caio C{\'e}sar~Teodoro Mendes, Arindam Mitra, Besmira Nushi, Dimitris Papailiopoulos, Olli Saarikivi, Shital Shah, Vaishnavi Shrivastava, Vibhav Vineet, Yue Wu, Safoora Yousefi, and Guoqing Zheng.
\newblock Phi-4-reasoning technical report.
\newblock \emph{arXiv preprint arXiv:2504.21318}, 2025.
\newblock URL \url{https://arxiv.org/abs/2504.21318}.

\bibitem[Face(2025)]{openr1}
Hugging Face.
\newblock {Open} {R1}: A fully open reproduction of {DeepSeek-R1}, January 2025.
\newblock URL \url{https://github.com/huggingface/open-r1}.

\bibitem[Wan et~al.(2025)Wan, Zhong, Yang, Chen, and Quan]{wan2025fusechat}
Fanqi Wan, Longguang Zhong, Ziyi Yang, Ruijun Chen, and Xiaojun Quan.
\newblock {FuseChat}: Knowledge fusion of chat models.
\newblock In \emph{Proceedings of the 2025 Conference on Empirical Methods in Natural Language Processing}, pages 21618--21642, Suzhou, China, November 2025. Association for Computational Linguistics.
\newblock \doi{10.18653/v1/2025.emnlp-main.1096}.
\newblock URL \url{https://aclanthology.org/2025.emnlp-main.1096/}.

\bibitem[Wen et~al.(2025)Wen, Cai, Xiao, He, An, Duan, Du, Liu, Tang, Lv, Zou, Deng, Jia, and Zhang]{wen2025lightr1}
Liang Wen, Yunke Cai, Fenrui Xiao, Xin He, Qi~An, Zhenyu Duan, Yimin Du, Junchen Liu, Lifu Tang, Xiaowei Lv, Haosheng Zou, Yongchao Deng, Shousheng Jia, and Xiangzheng Zhang.
\newblock {Light-R1}: Curriculum {SFT}, {DPO} and {RL} for long {COT} from scratch and beyond.
\newblock In \emph{Proceedings of the 63rd Annual Meeting of the Association for Computational Linguistics (Volume 6: Industry Track)}, pages 318--327, Vienna, Austria, July 2025. Association for Computational Linguistics.
\newblock \doi{10.18653/v1/2025.acl-industry.24}.
\newblock URL \url{https://aclanthology.org/2025.acl-industry.24/}.

\bibitem[Liu et~al.(2025{\natexlab{b}})Liu, Yang, Chen, Lee, Shoeybi, Catanzaro, and Ping]{liu2025acereason}
Zihan Liu, Zhuolin Yang, Yang Chen, Chankyu Lee, Mohammad Shoeybi, Bryan Catanzaro, and Wei Ping.
\newblock Acereason-nemotron 1.1: Advancing math and code reasoning through sft and rl synergy.
\newblock \emph{arXiv preprint arXiv:2506.13284}, 2025{\natexlab{b}}.
\newblock URL \url{https://arxiv.org/abs/2506.13284}.

\bibitem[Basant et~al.(2025)Basant, Khairnar, Paithankar, Khattar, Renduchintala, Malte, Bercovich, Hazare, Rico, Ficek, et~al.]{nvidia2025nemotronnano2}
Aarti Basant, Abhijit Khairnar, Abhijit Paithankar, Abhinav Khattar, Adithya Renduchintala, Aditya Malte, Akhiad Bercovich, Akshay Hazare, Alejandra Rico, Aleksander Ficek, et~al.
\newblock {NVIDIA} nemotron nano 2: An accurate and efficient hybrid mamba-transformer reasoning model.
\newblock \emph{arXiv preprint arXiv:2508.14444}, 2025.
\newblock URL \url{https://arxiv.org/abs/2508.14444}.

\bibitem[Labs(2025)]{bespoke_stratos}
Bespoke Labs.
\newblock Bespoke-stratos: The unreasonable effectiveness of reasoning distillation.
\newblock Bespoke Labs blog, 2025.
\newblock URL \url{https://www.bespokelabs.ai/blog/bespoke-stratos-the-unreasonable-effectiveness-of-reasoning-distillation}.
\newblock Accessed: 2025-01-22.

\bibitem[Gugger et~al.(2022)Gugger, Debut, Wolf, Schmid, Mueller, Mangrulkar, Sun, and Bossan]{accelerate}
Sylvain Gugger, Lysandre Debut, Thomas Wolf, Philipp Schmid, Zachary Mueller, Sourab Mangrulkar, Marc Sun, and Benjamin Bossan.
\newblock Accelerate: Training and inference at scale made simple, efficient and adaptable., 2022.
\newblock URL \url{https://github.com/huggingface/accelerate}.

\bibitem[Zhang et~al.(2025{\natexlab{b}})Zhang, Hariri, Zhong, Chaudhary, Sui, Hu, and Shrivastava]{Zhang2025DFloat11}
Tianyi Zhang, Mohsen Hariri, Shaochen Zhong, Vipin Chaudhary, Yang Sui, Xia Hu, and Anshumali Shrivastava.
\newblock 70\% size, 100\% accuracy: Lossless llm compression for efficient gpu inference via dynamic-length float.
\newblock In \emph{Advances in Neural Information Processing Systems (NeurIPS 2025)}, 2025{\natexlab{b}}.
\newblock URL \url{https://arxiv.org/abs/2504.11651}.

\end{thebibliography}
\bibliographystyle{unsrtnat}
\newpage

\setcounter{tocdepth}{3}
\addcontentsline{toc}{section}{Appendix}
\begingroup
\setlength{\parskip}{2pt}
\tableofcontents
\endgroup
\clearpage

\appendix

\section{Derivation of Bayesian Estimator and Uncertainty}
\label{app:Bayes_derivation}

As described in the main text, the Bayesian framework is built on two quantities. The first is $\mu(R)$, the average of $\bar{\pi}$ over the joint posterior for all the questions:
\begin{equation}\label{mudef}
    \mu(R) = \int_\Delta d\bm{\pi}_1 \cdots \int_\Delta d\bm{\pi}_M\,\bar\pi \prod_{\alpha=1}^M {\cal P}(\bm{\pi}_\alpha | \bm{R}_\alpha),
\end{equation}
where the integration region $\Delta$ is the probability simplex defined as the set of all possible $(C+1)$-dimensional vectors $\bm{p}$ such that $\sum_{k=0}^C p_k =1$. The second is the variance $\sigma^2(R)$ associated with our Bayesian estimator,
\begin{equation}\label{sigdef}
    \sigma^2(R) = \int_\Delta d\bm{\pi}_1 \cdots \int_\Delta d\bm{\pi}_M\,(\bar\pi - \mu(R))^2 \prod_{\alpha=1}^M {\cal P}(\bm{\pi}_\alpha | \bm{R}_\alpha).
\end{equation}

Our derivation of closed-form expressions for $\mu$ and $\sigma$ builds on the generalized ($C > 1$) and original ($C = 1$) Laplace rule of succession theory from \cite{jaynes2003probability}, recovering those results in the special case of a single question ($M = 1$). We start with Bayes' rule for each row of $R$:
\begin{equation}\label{bayes}
{\cal P}(\bm{\pi}_\alpha | \bm{R}_\alpha) = \frac{{\cal P}(\bm{R}_\alpha | \bm{\pi}_\alpha) {\cal P}(\bm{\pi}_\alpha)}{{\cal P}(\bm{R}_\alpha)}.
\end{equation}
The likelihood ${\cal P}(\bm{R}_\alpha | \bm{\pi}_\alpha)$ is a $(C+1)$-category multinomial distribution over $N$ trials, with the probability distribution function:
\begin{equation}\label{e4}
{\cal P}(\bm{R}_\alpha | \pi_\alpha) = \frac{N!}{n_{\alpha 0}! n_{\alpha 1}! \cdots n_{\alpha C}!} \prod_{k=0}^C \left(\pi_{\alpha k}\right)^{n_{\alpha k}},
\end{equation}
where $n_{\alpha k} = \sum_{i=1}^N \delta_{k, R_{\alpha i}}$, $\bm{n}_\alpha$ is the vector with elements $n_{\alpha k}$, and $\delta_{i,j}$ is the Kronecker delta.

The prior ${\cal P}(\bm{\pi}_\alpha)$ is chosen as the conjugate prior of the multinomial, a Dirichlet distribution ${\cal P}(\bm{\pi}_\alpha) \sim \text{Dir}(\bm{n}_\alpha^0)$, with concentration parameter vector $\bm{n}_\alpha^0 = (n_{\alpha 0}^0, \ldots, n_{\alpha C}^0)$.~\cite{hayden2025straightforward} A uniform prior (no prior knowledge) sets $n_{\alpha k}^0 = 1$ for all $k$. Prior information from an earlier $M \times D$ matrix $R^0$ (with $R^0_{\alpha i}$ as the category for the $i$th trial of the $\alpha$th question) can be incorporated as:
\begin{equation}
\label{e4b}
n^0_{\alpha k} = 1 + \sum_{i=1}^D \delta_{k, R^0_{\alpha i}}.
\end{equation}
The Dirichlet prior is:
\begin{equation}\label{e4c}
{\cal P}(\bm{\pi}_\alpha) = \frac{\Gamma(1 + C + D)}{\prod_{k=0}^C \Gamma(n^0_{\alpha k})} \prod_{k=0}^C \left(\pi_{\alpha k}\right)^{n^0_{\alpha k} - 1},
\end{equation}
where $\sum_{k=0}^C n^0_{\alpha k} = 1 + C + D$.

The normalization constant ${\cal P}(\bm{R}_\alpha)$ is:
\begin{equation}
{\cal P}(\bm{R}_\alpha) = \int_{\Delta} d\bm{p} \, {\cal P}(\bm{R}_\alpha | \bm{p}) {\cal P}(\bm{p}),
\end{equation}
and since the Dirichlet is the conjugate prior, the posterior is ${\cal P}(\bm{\pi}_\alpha | \bm{R}_\alpha) \sim \text{Dir}(\bm{\nu}_\alpha)$, with $\bm{\nu}_\alpha = \bm{n}_\alpha + \bm{n}^0_\alpha$. The posterior distribution is:
\begin{equation}\label{e6}
{\cal P}(\bm{\pi}_\alpha | \bm{R}_\alpha) = \frac{\Gamma(T)}{\prod_{k=0}^C \Gamma(\nu_{\alpha k})} \prod_{k=0}^C \left(\pi_{\alpha k}\right)^{\nu_{\alpha k} - 1},
\end{equation}
where $T \equiv \sum_{k=0}^C \nu_{\alpha k} = 1 + C + D + N$.

The moment generating function $\Phi(t) = \langle \exp(\bar{\pi} t) \rangle$ is:
\begin{equation}\label{e8}
\begin{split}
\Phi(t) &= \int_\Delta d\bm{\pi}_1 \cdots \int_\Delta d\bm{\pi}_M \exp\left(t \bar{\pi}\right) \prod_{\alpha=1}^M {\cal P}(\bm{\pi}_\alpha | \bm{R}_\alpha) \\
&= \prod_{\alpha=1}^M \int_\Delta d\bm{\pi}_\alpha \exp\left(\frac{t}{M} \sum_{k=0}^C w_k \pi_{\alpha k}\right) {\cal P}(\bm{\pi}_\alpha | \bm{R}_\alpha) \\
&= e^{t w_0} \prod_{\alpha=1}^M \int_\Delta d\bm{\pi}_\alpha \exp\left(t \sum_{k=1}^C s_k \pi_{\alpha k}\right) {\cal P}(\bm{\pi}_\alpha | \bm{R}_\alpha),
\end{split}
\end{equation}
where $s_k \equiv (w_k - w_0)/M$, and $\pi_{\alpha 0} = 1 - \sum_{k=1}^C \pi_{\alpha k}$.

Each integral is the moment-generating function for a Dirichlet distribution, expressed via the confluent Lauricella hypergeometric function $\Psi^{[C]}$:
\begin{equation}\label{e8b}
\Phi(t) = e^{t w_0} \prod_{\alpha=1}^M \Psi^{[C]}\left(\nu_{\alpha 1}, \ldots, \nu_{\alpha C}; T; ts_1, \ldots, ts_C\right),
\end{equation}
where
\begin{equation}
\Psi^{[C]}\left(\nu_{\alpha 1}, \ldots, \nu_{\alpha C}; T; ts_1, \ldots, ts_C\right) = \sum_{m_1=0}^\infty \cdots \sum_{m_C=0}^\infty \frac{(\nu_{\alpha 1})_{m_1} \cdots (\nu_{\alpha C})_{m_C} (ts_1)^{m_1} \cdots (ts_C)^{m_C}}{(T)_m m_1! \cdots m_C!},
\end{equation}
and $(x)_n$ is the Pochhammer symbol.

The moments are:
\begin{equation}\label{e9o}
\mu = \Phi'(0), \qquad \sigma^2 = \Phi''(0) - (\Phi'(0))^2.
\end{equation}
Expanding $\Psi^{[C]}$ to ${\cal O}(t^2)$:
\begin{equation}\label{e9}
\begin{split}
\Psi^{[C]} &= 1 + \frac{t}{T} \sum_{j=1}^C \nu_{\alpha j} s_j + \frac{t^2}{2 T (T+1)} \sum_{j=1}^C \nu_{\alpha j} (\nu_{\alpha j} + 1) s_j^2 \\
&\quad + \frac{t^2}{T (T+1)} \sum_{\ell=1}^C \sum_{m=\ell+1}^C \nu_{\alpha \ell} \nu_{\alpha m} s_\ell s_m + {\cal O}(t^3).
\end{split}
\end{equation}
Substituting into equation~\ref{e8b} and computing derivatives yields:
\begin{equation}
\label{e11}
\begin{split}
\mu &= w_0 + \frac{1}{M T} \sum_{\alpha=1}^M \sum_{j=0}^C \nu_{\alpha j} (w_j - w_0), \\
\sigma^2 &= \frac{1}{M^2 (T+1)} \sum_{\alpha=1}^M \left\{ \sum_{j=0}^C \frac{\nu_{\alpha j}}{T} (w_j - w_0)^2 - \left(\sum_{j=0}^C \frac{\nu_{\alpha j}}{T} (w_j - w_0)\right)^2 \right\}.
\end{split}
\end{equation}
The algorithm summarizing this calculation is shown in Algorithm~\ref{summary} in the main text.

\section{Proof of Equivalence of Bayesian and Average Rankings for Uniform Prior}
\label{app:bayes_avg_equiv}

For Bayesian estimators using a uniform prior (where $D=0$, $T = 1 + C + N$, $\nu_{\alpha k} = 1 + n_{\alpha k}$), the expression for the mean $\mu$ from equation~\ref{e11} simplifies as:
\begin{equation}
\begin{split}
\mu &= w_0 + \frac{1}{M (1 + C + N)} \sum_{\alpha=1}^M \sum_{j=0}^C (1 + n_{\alpha j})(w_j - w_0) \\
&= A + \frac{1}{M (1 + C + N)} \sum_{\alpha=1}^M \sum_{j=0}^C w_j n_{\alpha j},
\end{split}
\end{equation}
where the constant $A$ is given by
\begin{equation}
A = \frac{1}{1 + C + N} \sum_{j=0}^C w_j,
\end{equation}
and $\sum_{j=0}^C n_{\alpha j} = N$. Here, $\mu$ relates to a naive weighted average accuracy $a$ over the number of answers in each category,
\begin{equation}
a = \frac{1}{M N} \sum_{\alpha=1}^M \sum_{j=0}^C w_j n_{\alpha j},
\end{equation}
via
\begin{equation}
\label{eqn:Bayes_avg_equiv}
\mu = A + \frac{N}{1 + C + N} a.
\end{equation}
Note that in the binary case where $C=1$, $w_0=0$, $w_1=1$, the value of $a$ is just regular average accuracy \avg{N}. For categorical cases, it is just a weighted generalization of \avg{N}.

Since $A$ is constant across models and the prefactor $\frac{N}{1 + C + N}$ is positive, we see that if $\mu > \mu^\prime$, the corresponding values of $a$ and $a^\prime$ from the two methods must always give the same ranking, $a > a^\prime$. Additionally, in the limit of a large number of trials, $N \to \infty$, we see that $A \to 0$ and $\mu \approx a$, as expected.

This equivalence extends to uncertainty quantification. The relationship between the standard deviation of the average ($\sigma_{\mathrm{avg@}N}$) and the Bayesian standard deviation ($\sigma_{\mathrm{Bayes@}N}$ from equation~\ref{e11}) is

\begin{equation}
\label{eqn:Bayes_avg_sigma}
\sigma_{\mathrm{avg@}N} = \frac{1 + C + N}{N} \sigma_{\mathrm{Bayes@}N}.
\end{equation}
The Bayesian expression for $\sigma_{\mathrm{Bayes@}N}$ is valid for all $M$ and $N$, providing a reliable method to compute uncertainty in avg@$N$ without relying on the Central Limit Theorem.

\section{Runtime}

To see the asymptotic runtime and memory scaling let:

\[
\begin{aligned}
&M = \text{number of problems (rows)},\\
&N = \text{number of trials per problem (columns in } R),\\
&D = \text{number of prior outcomes per problem (columns in } R_0 \text{, which may be } 0),\\
&C+1 = \text{number of categories}.
\end{aligned}
\]

From Algorithm~1, the work is:

\[
\text{Two row-wise histograms: }
\mathcal{O}(MN) \ \text{for } R \;+\; \mathcal{O}(MD) \ \text{for } R_0,
\]

\[
\text{Posterior mean and variance on } \nu \in \mathbb{R}^{M \times (C+1)}:
\quad
\mathcal{O}\!\left(M(C+1)\right).
\]

So the \textbf{overall time complexity} is:

\[
\mathcal{O}\!\left(M(N + D + C)\right)
\]

i.e., linear in the number of entries in the result matrices and linear in the number of categories.

The \textbf{memory footprint} is likewise linear:

\[
\text{Store } R \text{ and (optionally) } R_0: \quad \mathcal{O}(MN + MD),
\]

\[
\text{Store per-row category counts and derived arrays (} \nu, \ \nu/T \text{): }
\quad \mathcal{O}(M(C+1)).
\]

Note that the evaluation consists of tallying counts and then plugging them into closed-form expressions for
\(\mu\) and \(\sigma\); no iterative optimization or Monte Carlo sampling is required.

\section{Categorical Evaluation}\label{app:sec:cat}

\subsection{Rubric-aware Bayes@N Evaluation of Reasoning Models}

As discussed in \Cref{theo:form,sec:exp:cat}, for each question $\alpha \in {1,\ldots,M}$, every attempt yields base signals such as \texttt{has\_box}, \texttt{is\_correct}, \texttt{token\_ratio}, \texttt{prompt\_bpt}, \texttt{completion\_bpt}, and verifier probabilities \texttt{compass\_context\_A}, \texttt{compass\_context\_B}, and \texttt{compass\_context\_C} for \emph{correct}, \emph{wrong}, and \emph{invalid/off-task}. Using thresholds and Boolean criteria, each attempt is mapped into one of $C+1$ categories under a chosen schema (e.g., \emph{Format Aware}, \emph{Conf-Wrong Penalty}, \emph{Efficiency-Adjusted}; Table~\ref{tab:cat_schema}). We instantiate categorical schemata and update posterior means via Dirichlet-multinomial inference, yielding metrics that preserve correctness while explicitly reflecting formatting, calibration, and efficiency.

\paragraph{Base signals}
All signals are directly obtainable from common LLM inference stacks such as Hugging Face transformers~\cite{wolf2019huggingface} and vLLM~\cite{kwon2023efficient}, via per-step scores/log-probs and termination metadata, and require no model-specific instrumentation; the verifier probabilities \texttt{compass\_context\_A}, \texttt{compass\_context\_B}, and \texttt{compass\_context\_C} are defined in \Cref{app:cat:rew}.

\begin{itemize}
  \item \textbf{has\_box}: 1 if a final boxed answer is present; else 0.
  \item \textbf{is\_correct}: 1 if the answer is correct; else 0.
  \item \textbf{token\_ratio}: completion tokens normalized by 32{,}768.
  \item \textbf{repeated\_pattern}: 0 if \texttt{finish\_reason} is \texttt{stop}; else 1 (degenerate output).
  \item \textbf{prompt\_bpt}: negative average prompt log-prob in bits/token (lower is better).
  \item \textbf{completion\_bpt}: negative average completion log-prob in bits/token (lower is better).
  \item \textbf{compass\_context\_A}: verifier contextual probability of \emph{correct}.
  \item \textbf{compass\_context\_B}: verifier contextual probability of \emph{wrong}.
  \item \textbf{compass\_context\_C}: verifier contextual probability of \emph{irrelevant/off-task}.
\end{itemize}

\paragraph{Reward models in evaluation.}\label{app:cat:rew}
While reward models are most familiar from fine-tuning (e.g., RLHF), we use one as a lightweight verifier to supply per-attempt label probabilities for
\[
\{compass\_context\_A, compass\_context\_B, compass\_context\_C\} = \{\text{correct}, \text{wrong}, \text{invalid/off-task}\}
\]
in evaluation. Concretely, we employ OpenCompass \texttt{CompassVerifier-3B} to produce probabilities and then apply \emph{contextual calibration} to obtain a more robust, prompt-stable label distribution: we evaluate next-token scores for the candidate labels at a fixed answer slot, subtract a content-free baseline logit $b_y$ from the task logit $s_y$ for each label $y$, and apply temperature scaling to yield calibrated probabilities
\[
p(y \mid x) = \mathrm{softmax}\!\left(\frac{s_y - b_y}{T}\right).
\]
This helps us mitigate saturation and the entanglement of formatting and confidence seen with last-token probabilities, and improves probability calibration for downstream rubric scoring.

\paragraph{Selected categorical schema.}

We define 12 schemata (Table~\ref{tab:cat_schema}) using the rubric variables (Table~\ref{app:tab:var}) derived from the base signals; here are two illustrative definitions (the others follow analogously):

\begin{itemize}
    \item \textbf{Format Aware}:
    \[
    \text{cat} =
    \begin{cases}
        0 & \text{invalid} \\
        1 & \text{wrong} \wedge \text{unboxed} \\
        2 & \text{wrong} \wedge \text{boxed} \\
        3 & \text{correct} \wedge \text{unboxed} \\
        4 & \text{correct} \wedge \text{boxed}
    \end{cases}
    \]

    \item \textbf{Conf-Wrong Penalty}:
    \[
    \text{cat} =
    \begin{cases}
        0 & \text{invalid} \\
        1 & \text{wrong}_{\text{high\_conf}} \\
        2 & \text{wrong} \wedge \text{low\_conf} \\
        3 & \text{correct}
    \end{cases}
    \]
\end{itemize}

Rubric weights $\mathbf{w}$ are chosen to reflect evaluation preferences. For example, \emph{Format Aware} might use $[0, 0, 1, 2, 3]$ to mildly reward formatting when correct and slightly penalize confidently wrong (via schema choice); \emph{Efficiency-Adjusted} can downweight verbose outputs among both correct and wrong categories.

\begin{itemize}
  \item \textbf{Exact Match} \; Correctness only; ignores formatting, confidence, and length.
  \item \textbf{Format Aware} \; Rewards boxed, well-formatted answers; distinguishes boxed/unboxed even when wrong.
  \item \textbf{Conf-Calibrated} \; Penalizes \emph{confidently wrong}; grades correct answers by confidence (low/mid/high).
  \item \textbf{OOD Robustness} \; Separates in-distribution vs. OOD prompts; checks correctness under both.
  \item \textbf{Strict Compliance} \; Requires boxed final answers; unboxed-correct is treated as non-compliant.
  \item \textbf{Conf-Wrong Penalty} \; Heavier penalty for wrong answers at high confidence; lighter when uncertain.
  \item \textbf{Verifier-Only} \; Uses verifier signals alone to rank; model-agnostic prob of the verifier.
  \item \textbf{Format+Confidence} \; Balanced composite over (boxed/unboxed) \(\times\) (low/high confidence) for both wrong and correct; emphasizes boxed, high-confidence correctness and penalizes confidently wrong.
  \item \textbf{Length-Robust} \; Isolates correctness irrespective of verbosity; does not penalize length.
  \item \textbf{Verifier Prob} \; Probes agreement with the verifier: flags wrong with high verifier \emph{A} as inconsistent and distinguishes under/over-confidence on correct.
  \item \textbf{Efficiency-Adjusted} \; Rewards short, correct completions; penalizes verbose outputs (especially when wrong).
  \item \textbf{Concision-High-Conf} \; Prefers concise, high-confidence correct answers; downweights verbose correctness.
\end{itemize}

\begin{table}[htbp]
  \centering
  \caption[Rubric variables and formulas used in categorical evaluation]{Rubric variables, decision formulas, and brief descriptions used to map each model attempt into discrete categories. Thresholds (\(\tau_{\text{high}}, \tau_{\text{low\_wrong}}, \tau_{\text{prompt}}\)) and length quantiles (\(\text{len\_p33}, \text{len\_p66}\)) are computed per dataset from observed bits-per-token and token-ratio statistics. Category \(0\) is reserved for invalid outputs (degenerate repetition or high verifier \(compass\_context\_C\)), and \(compass\_context\_A, compass\_context\_B, compass\_context\_C\) denote calibrated verifier probabilities for \emph{correct}, \emph{wrong}, and \emph{off-task}, respectively.}
  \label{app:tab:var}
  \begin{tabular}{@{}lll@{}}
    \toprule
    \textbf{Rubric variables } & \textbf{Formula} & \textbf{Description} \\
    \midrule
        invalid & $(\text{repeated\_pattern} = 1) \lor (compass\_context\_C \geq 0.5)$ & Category 0 reserved for invalid. \\
    correct & $(\text{is\_correct} \geq 0.5)$ & Boolean mask of correctness. \\
    wrong & $(\text{is\_correct} < 0.5)$ & Complement of correct. \\
    high\_conf & $(\text{completion\_bpt} \leq \tau_{\text{high}})$ & Confidence proxy \\
    low\_conf & $(\text{completion\_bpt} > \tau_{\text{high}})$ & Complement of high\_conf. \\
    wrong\_high\_conf & wrong $\land$ $(\text{completion\_bpt} \leq \tau_{\text{low\_wrong}})$ & Penalize confidently wrong. \\
    ood & $(\text{prompt\_bpt} \geq \tau_{\text{prompt}})$ & Out-of-distribution prompt. \\
    ind & $(\text{prompt\_bpt} < \tau_{\text{prompt}})$ & In-distribution prompt. \\
    economical & $(\text{token\_ratio} \leq \text{len\_p33})$ & Short completions. \\
    moderate & $(\text{len\_p33} < \text{token\_ratio} \leq \text{len\_p66})$ & Medium-length completions. \\
    verbose & $(\text{token\_ratio} > \text{len\_p66})$ & Long completions. \\
    boxed & $(\text{has\_box} \geq 0.5)$ & Answer is boxed. \\
    unboxed & $(\text{has\_box} < 0.5)$ & Answer is not boxed. \\
    A\_high & $(compass\_context\_A \geq 0.6)$ & Verifier confidence high. \\
    $\tau_{\text{high}}$ & 40th percentile of $\text{completion\_bpt}$ &  \\
    $\tau_{\text{low\_wrong}}$ & 60th percentile of $\text{completion\_bpt}$ among wrong items & \\
    $\tau_{\text{prompt}}$ & 90th percentile of $\text{prompt\_bpt}$ &  \\
    $\text{len\_p33},\,\text{len\_p66}$ & 33rd and 66th percentiles of $\text{token\_ratio}$ &  \\
    $\text{corr\_p33},\,\text{corr\_p66}$ & 33rd and 66th percentiles of $\text{completion\_bpt}$ correct items & \\
    \bottomrule
  \end{tabular}
\end{table}

\begin{table}[htbp]
  \centering
  \caption[Definitions of categorical evaluation schemata]{Definitions of the twelve categorical evaluation schemata used in our Dirichlet–multinomial framework. Each schema specifies decision rules over correctness, formatting (boxed/unboxed), confidence (via \(\text{completion\_bpt}\)), prompt distribution (in-distribution vs.\ OOD), output economy (via \(\text{token\_ratio}\)), and verifier signals \((A,B,C)\). These rules map every attempt into \(C{+}1\) discrete categories, enabling posterior means and credible intervals for any chosen weight vector \(\mathbf{w}\).}
  \label{tab:cat_schema}
  \begin{tabularx}{\textwidth}{lX}
    \toprule
    \textbf{Categorical Schema } & \textbf{Rubric} \\
\midrule
    Exact Match & 0 invalid; 1 wrong; 2 correct \\
    Format Aware & 0 invalid; 1 wrong $\land$ unboxed; 2 wrong $\land$ boxed; 3 correct $\land$ unboxed; 4 correct $\land$ boxed \\
    Conf-Calibrated & 0 invalid; 1 wrong $\land$ low\_conf; 2 wrong\_high\_conf; 3 correct $\land$ low\_conf; 4 correct $\land$ mid; 5 correct $\land$ high\_conf \\
    OOD Robustness & 0 invalid; 1 ood $\land$ wrong; 2 ind $\land$ wrong; 3 ood $\land$ correct; 4 ind $\land$ correct \\
    Strict Compliance & 0 invalid; 1 wrong $\lor$ (correct $\land$ unboxed); 2 correct $\land$ boxed \\
    Conf-Wrong Penalty & 0 invalid; 1 wrong\_high\_conf; 2 wrong $\land$ low\_conf; 3 correct \\
    Verifier-Only & 0 invalid; 1 high C; 2 high B; 3 A\_high \\
    Format+Confidence & 0 invalid; 1 wrong $\land$ unboxed; 2 wrong $\land$ boxed $\land$ low\_conf; 3 wrong $\land$ boxed $\land$ high\_conf; 4 correct $\land$ unboxed $\land$ low\_conf; 5 correct $\land$ unboxed $\land$ high\_conf; 6 correct $\land$ boxed $\land$ low\_conf; 7 correct $\land$ boxed $\land$ high\_conf \\
    Length-Robust & 0 invalid; 1 wrong; 2 correct \\
    Verifier Prob & 0 invalid; 1 wrong $\land$ A\_high; 2 wrong $\land$ $\lnot$ A\_high; 3 correct $\land$ $\lnot$ A\_high; 4 correct $\land$ A\_high \\
    Efficiency-Adjusted & 0 invalid; 1 wrong $\land$ economical; 2 wrong $\land$ moderate; 3 wrong $\land$ verbose; 4 correct $\land$ economical; 5 correct $\land$ moderate; 6 correct $\land$ verbose \\
    Concision-High-Conf & 0 invalid; 1 wrong; 2 correct $\land$ verbose; 3 correct $\land$ moderate; 4 correct $\land$ economical; 5 correct $\land$ economical $\land$ high\_conf \\
    \bottomrule
  \end{tabularx}
\end{table}

\subsection{Domain-agnostic rubric-aware Bayes@N}

The Bayesian construction is intentionally domain-agnostic: it applies whenever
model outputs can be mapped into a finite set of categories equipped with a
rubric. The evaluator specifies
\begin{enumerate}
  \item a mapping from raw outputs (and any side information) to categorical
        labels $R_{\alpha i} \in \{0,\dots,C\}$, and
  \item a weight vector $w$ that encodes how those categories are valued.
\end{enumerate}
Given these choices, Bayes@N returns the posterior mean $\mu(R)$ as a
rubric-aware point estimate, and $\sigma(R)$ as an uncertainty estimate, for
\emph{any} such categorical evaluation.

This viewpoint naturally covers subjective tasks. For instance:
\begin{itemize}
  \item In summarization, each response could be rated
        $\{\text{bad}, \text{okay}, \text{good}, \text{excellent}\}$ or by
        multi-criteria scores such as faithfulness, coverage, style, and
        harmful content. Each discrete level becomes a category index $k$, and
        $w_k$ reflects the importance of that level or criterion.
  \item In dialogue safety, categories might distinguish
        $\{\text{unsafe}, \text{borderline}, \text{safe}\}$, or finer-grained
        notions such as policy violations vs.\ merely over-cautious refusals.
\end{itemize}
Once the labels are available (from humans or an LLM-as-a-judge), Bayes@N
provides Bayesian estimates and credible intervals for any chosen rubric-based
score, reusing the same closed-form posterior as in the binary case.

Two aspects are particularly promising for future work in such subjective
domains:
\begin{enumerate}
  \item \textbf{Preference-based evaluation with rubrics.}
        When model comparisons are driven by preferences (either from human
        experts or LLM judges), each comparison can be converted into
        categorical labels over rubric dimensions (e.g., faithfulness,
        verbosity, harmfulness). A downstream weight vector $w$ can then fold
        these dimensions into a single scalar score that reflects
        application-specific trade-offs.
  \item \textbf{Transferring prior evidence across related tasks.}
        The optional prior matrix $R_0$ in Algorithm~1 lets us encode earlier
        outcome frequencies as a Dirichlet prior. For example, if a
        summarization system has been evaluated on a news dataset, the
        empirical category counts on that dataset can serve as prior counts
        when evaluating a closely related dataset. This allows stable rubric
        distributions to be reused across adjacent tasks or benchmark
        revisions, while still updating with new data.
\end{enumerate}
An important limitation in subjective settings is that Bayes@N does \emph{not}
resolve disagreement or bias in the rubric or labeling process itself. The
framework assumes a labeling scheme (from humans or an LLM-based judge) and a
weight vector $w$ are given; it then provides a statistically principled way to
aggregate those labels and quantify uncertainty. Designing good rubrics and
calibrating judges remain separate modeling decisions.

\section{\scorio{}}\label{sec:app:scorio}

Alongside this paper, we release \scorio{}, an open-source Python package that implements the evaluation framework presented in this work. \scorio{} provides a simple, unified API for computing \bayes{N}, \avg{N}, \pass{k}, and their credible intervals, enabling researchers to adopt principled Bayesian evaluation with minimal effort. The package is available on PyPI and its documentation is hosted at \url{https://scorio.readthedocs.io}.

\paragraph{Installation.} \scorio{} can be installed via:
\begin{lstlisting}[style=pythonstyle, numbers=none]
pip install scorio
\end{lstlisting}

\paragraph{Basic usage.}
All evaluation functions operate on a results matrix $R \in \{0, \dots, C\}^{M \times N}$, where $M$ is the number of problems, $N$ is the number of trials per problem, and $C+1$ is the number of outcome categories.
An optional weight vector $w$ of length $C+1$ maps each category to a score.
\Cref{lst:binary} shows binary evaluation using both the Bayesian estimator and \pass{k}.

\begin{lstlisting}[style=pythonstyle, caption={Binary evaluation with \scorio{}.}, label=lst:binary]
import numpy as np
from scorio import eval

# Binary outcomes: M=2 problems, N=5 trials each
R = np.array([[0, 1, 1, 0, 1],
              [1, 1, 0, 1, 1]])

# Bayesian evaluation (binary: w defaults to (0, 1))
mu, sigma = eval.bayes(R)
print(f"Bayes@5: mu={mu:.4f}, sigma={sigma:.4f}")

# Average accuracy
a, sigma_a = eval.avg(R)
print(f"avg@5:   mu={a:.4f}, sigma={sigma_a:.4f}")

# Pass@k
print(f"Pass@1 = {eval.pass_at_k(R, k=1):.4f}")
print(f"Pass@2 = {eval.pass_at_k(R, k=2):.4f}")
\end{lstlisting}

\paragraph{Credible intervals.}
Each estimator has a companion \texttt{\_ci} function that returns the posterior mean, standard deviation, and a credible interval, as shown in \Cref{lst:ci}.

\begin{lstlisting}[style=pythonstyle, caption={Computing credible intervals.}, label=lst:ci]
# 95% credible interval for Bayes@N
mu, sigma, lo, hi = eval.bayes_ci(R, confidence=0.95)
print(f"Bayes@5: {mu:.4f} [{lo:.4f}, {hi:.4f}]")

# 95% credible interval for Pass@k
mu, sigma, lo, hi = eval.pass_at_k_ci(R, k=1)
print(f"Pass@1:  {mu:.4f} [{lo:.4f}, {hi:.4f}]")
\end{lstlisting}

\paragraph{Categorical (rubric-based) evaluation.}
For graded outcomes with $C > 1$ categories, a weight vector $w$ specifies the score associated with each category.
\Cref{lst:categorical} illustrates evaluation under a three-level rubric ($C=2$) with partial credit.

\begin{lstlisting}[style=pythonstyle, caption={Categorical evaluation with a weighted rubric.}, label=lst:categorical]
# Graded outcomes: 0=incorrect, 1=partial, 2=correct
R = np.array([[0, 2, 1, 0, 2],
              [2, 1, 1, 2, 1]])

# Weight vector: incorrect=0, partial=0.5, correct=1
w = np.array([0.0, 0.5, 1.0])

mu, sigma = eval.bayes(R, w)
print(f"Bayes@5 (graded): mu={mu:.4f}, sigma={sigma:.4f}")

mu, sigma, lo, hi = eval.bayes_ci(R, w, confidence=0.95)
print(f"95% CrI: [{lo:.4f}, {hi:.4f}]")
\end{lstlisting}

\paragraph{Incorporating prior evidence.}
When prior evaluation data are available (e.g., from a previous benchmark or a pilot study), they can be passed as a prior matrix $R^0$ to inform the posterior, as described in \Cref{app:non_uni_prior}. \Cref{lst:prior} shows a minimal example in which a short pilot run is reused as prior evidence for a new evaluation.

\begin{lstlisting}[style=pythonstyle, caption={Using prior evidence.}, label=lst:prior]
# Prior outcomes from a pilot study (M=2, D=3 trials)
R0 = np.array([[1, 0, 1],
               [0, 1, 0]])

mu, sigma = eval.bayes(R, w=None, R0=R0)
print(f"Bayes@5 (with prior): mu={mu:.4f}, sigma={sigma:.4f}")
\end{lstlisting}

\Cref{tab:scorio_api} summarizes the main \scorio{} API.

\begin{table}[h]
\centering
\caption{Summary of the \scorio{} evaluation API. All functions accept a results matrix $R \in \{0,\dots,C\}^{M \times N}$. Functions with the \texttt{\_ci} suffix additionally return a credible interval.}
\label{tab:scorio_api}
\footnotesize
\begin{tabular}{@{}lll@{}}
\toprule
\textbf{Function} & \textbf{Returns} & \textbf{Description} \\
\midrule
\texttt{bayes(R, w, R0)} & $(\mu, \sigma)$ & Bayesian posterior mean and uncertainty \\
\texttt{bayes\_ci(R, w, R0)} & $(\mu, \sigma, \text{lo}, \text{hi})$ & \quad + credible interval \\
\midrule
\texttt{avg(R, w)} & $(a, \sigma_a)$ & Weighted average and uncertainty \\
\texttt{avg\_ci(R, w)} & $(a, \sigma_a, \text{lo}, \text{hi})$ & \quad + credible interval \\
\midrule
\texttt{pass\_at\_k(R, k)} & $p$ & \pass{k} estimate \\
\texttt{pass\_at\_k\_ci(R, k)} & $(\mu, \sigma, \text{lo}, \text{hi})$ & \quad + credible interval \\
\texttt{pass\_hat\_k(R, k)} & $p$ & Pass\^{}$k$ \\
\texttt{g\_pass\_at\_k\_tau(R, k, tau)} & $p$ & G-Pass@$k_{\tilde{\tau}}$ \\
\texttt{mg\_pass\_at\_k(R, k)} & $p$ & mG-Pass@$k$ \\
\bottomrule
\end{tabular}
\end{table}

\section{Extended Related Work}
\label{app:extended_related_work}

The evaluation of LLMs in generative reasoning tasks, under test-time scaling (e.g., via repeated sampling\cite{brown2024large}), has evolved to address the stochastic nature of inference and the need for robust measures of functional correctness. Early approaches relied on syntactic similarity metrics like BLEU \cite{papineni-etal-2002-bleu} and CodeBLEU \cite{ren2020codebleu}, which compare generated answers against reference solutions. However, these metrics often fail to capture semantic correctness in reasoning tasks, motivating metrics based on execution-validation or test-based validation \cite{kulal2019spoc, ren2020codebleu}. This limitation has shifted focus toward functional evaluation, where the generated solution is assessed via a ground truth to verify correctness\cite{kulal2019spoc, hendrycks2021measuring}. In this section, we review key functional metrics, focusing on those that leverage multiple samples to scale performance at inference time. These metrics form the basis to assess LLM capabilities but often overlook probabilistic uncertainty or consistency across samples, motivating our novel Bayesian framework.

\textbf{The Pass@$k$ metric}, originally introduced by \cite{kulal2019spoc,chen2021evaluating} for evaluating LLMs trained on code. It measures the probability that at least one of $k$ independently generated samples for a given problem passes all associated unit tests (i.e., by matching ground-truth answers or satisfying logical constraints), offering a practical estimate of a model's potential performance in solving a variety of complex tasks and problems. The unbiased estimator of Pass@$k$ is computed as:
\begin{equation}
\label{eqn:pass_at_k}
    \text{Pass@}k = \mathbb{E}_{\text{problems}} \left[ 1 - \frac{\binom{n - c}{k}}{\binom{n}{k}} \right],
\end{equation}
where $n$ is the total number of generated samples and $c$ is the total number of correct solutions within the $n$ trials. This estimator has smaller uncertainty in the limit of $n \gg k$, ensuring reliable approximations. However, due to computational costs, $k$ is often comparable to $n$ in practice, which can increase variance and weaken evaluation stability. The Pass@$k$ metric has been adapted beyond code to evaluate LLMs in various tasks requiring verifiable correctness, such as math, logic, and general reasoning \cite{hendrycks2021measuring,cobbe2021training,wang2022self,lewkowycz2022solving}.

\textbf{Pass\textasciicircum$k$}, introduced in \cite{yao2024taubench}, extends the Pass@$k$ metric to capture both the potential performance and the consistency of LLMs in reasoning tasks, where evaluating the reliability and stability of generated solutions is crucial. Pass\textasciicircum$k$ is defined as the probability that all $k$ trials are correct:

\begin{equation}
\label{eqn:pass_to_k}
    \text{Pass\textasciicircum} k = \mathbb{E}_{\text{problems}} \left[ \frac{\binom{c}{k}}{\binom{n}{k}} \right],
\end{equation}
where $c$ and $n$ retain the same meanings as in Pass@$k$. This metric assumes that all the trials are independent and uniformly distributed, approximating the binomial distribution with a hypergeometric distribution to account for sampling without replacement. By requiring all $k$ samples to be correct, Pass\textasciicircum$k$ provides a stringent measure of model consistency and stability.

To introduce flexibility, \citet{liu2024your} proposed \textbf{G-Pass@$k_{\tilde{\tau}}$}, which incorporates a tolerance threshold $\tilde{\tau} \in (0.0, 1.0]$:
\begin{equation}
\label{eqn:G_pass_at_k_tau}
\text{G-Pass@}k_{\tilde{\tau}} = \mathbb{E}_{\text{problems}} \left[ \sum_{j=\lceil \tau \cdot k \rceil}^{c} \frac{\binom{c}{j} \cdot \binom{n-c}{k-j}}{\binom{n}{k}} \right],
\end{equation}
where $\lceil \tau \cdot k \rceil$ is the smallest integer greater than or equal to $\tau \cdot k$. This formulation allows up to $k - \lceil \tau \cdot k \rceil$ incorrect solutions, balancing the assessment of potential with consistency. As a special case, Pass@$k$ corresponds to G-Pass@$k_{\tilde{\tau}}$ in the limit $\tau \to 0$.

Furthermore, \citet{liu2024your} introduced \textbf{mG-Pass@$k$}, an interpolated metric that integrates G-Pass@$k_{\tilde{\tau}}$ over $\tau \in [0.5, 1.0]$:
\begin{equation}
\label{eqn:mG_pass_at_k}
\text{mG-Pass@}k = 2 \int_{0.5}^{1.0} \text{G-Pass@}k_\tau  d\tau \approx \frac{2}{k} \sum_{i=\lceil 0.5 \cdot k \rceil + 1}^{k} \text{G-Pass@}k_{i/k},
\end{equation}
providing a more comprehensive measure that jointly reflects performance potential and reasoning stability.

These extended metrics have been applied to mathematical reasoning benchmarks such as LiveMathBench, MATH, and AIME, where they reveal substantial performance degradation of LLMs under stricter stability requirements.

\section{Experiment Setup and Reproducibility}\label{app:experiment}

\subsection{Metrics}

\paragraph{Kendall's Tau:}\label{app:kendall_tau}

Kendall's tau ($\tau$) \cite{10.1093/biomet/30.1-2.81} is a nonparametric rank correlation coefficient that quantifies the ordinal relationship between two ranked sets by evaluating the consistency in their orderings. For two rankings of $n$ items, it examines all unique pairs $(i, j)$ where $i < j$:

\begin{itemize}
    \item A pair is \emph{concordant} if the relative ordering of items $i$ and $j$ is the same in both rankings (both place $i$ before $j$ or vice versa).
    \item A pair is \emph{discordant} if the relative ordering is different.
    \item Pairs with ties in either ranking are neither concordant nor discordant.
\end{itemize}

Define $n_c$ as the number of concordant pairs, $n_d$ as the number of discordant pairs, and $n_0 = n(n-1)/2$ as the total number of unique pairs. Let $n_1$ represent the number of tied pairs in the first ranking, and $n_2$ similarly for the second ranking.
The two common variants are the following:

\begin{align}
    \text{Tau-a:}\quad \tau_a &= \frac{n_c - n_d}{n_0} \qquad\qquad\qquad\qquad \text{(no adjustment for ties)}, \\
    \text{Tau-b:}\quad \tau_b &= \frac{n_c - n_d}{\sqrt{(n_0 - n_1)(n_0 - n_2)}} \qquad \text{(adjusts for ties in both rankings)}.
\end{align}

Tau-a assumes no ties and may underestimate correlation when ties occur. Tau-b, which corrects for ties, is better suited for datasets with equivalent rankings.

In our implementation, we use \texttt{scipy.stats.kendalltau} with its default variant='b', which computes $\tau_b$ efficiently and handles ties appropriately. The coefficient ranges from $-1$ (perfect disagreement) to $+1$ (perfect agreement), with $0$ indicating no association. This metric provides a robust, distribution-free measure for comparing model performance rankings, particularly when ties reflect meaningful equivalences.

\paragraph{Convergence@$n$.} For a given bootstrap replicate, we measure convergence in terms of an \emph{exact ranking match}.
At each step $s \in \{1,\dots,N_{\max}\}$, we compute the ranking induced by the first $s$ trials and compare it to a gold-standard ranking (obtained from all $N_{\max}$ trials).
We then define
\[
s^\star \;=\; \min\Bigl\{ s \le N_{\max}\!-\!1 \Bigm|
\begin{array}{l}
\text{the ranking after $s$ trials matches the gold-standard ranking,} \\
\text{and remains unchanged after every subsequent trial}
\end{array}
\Bigr\},
\]
and refer to $s^\star$ as the convergence@$n$ value for that replicate.
If no such $s^\star \le N_{\max}$ exists, we declare that replicate to exhibit \emph{no convergence}.

\subsection{Models and Datasets}\label{app:exp}
\paragraph{Datasets.}
We evaluate on four math-reasoning test sets: \aimefour~\cite{MAA_AIME2024}, \aimefive~\cite{MAA_AIME2025}, \brumo~\cite{BrUMO_2025}, and \hmmt~\cite{HMMT_Feb2025}. AIME is administered by the Mathematical Association of America and consists of two sets of 15 integer-answer problems; we use the 2024 and 2025 problem sets. For \hmmt, we use the officially posted February 2025 contest set (algebra, geometry, number theory, and combinatorics). For \brumo, we use the published 2025 problem sets from the tournament archive.

\paragraph{Models.}
Unless noted otherwise, we run each generator with the provider-recommended chat template (DeepSeek/Qwen style when unspecified) and identical decoding settings (below) to minimize template-induced variance. The base model cohort includes 11 models (8 distinct models + 3 modes (low, medium, and high) of gpt-oss) as follows:
\skyicon~Sky-T1-32B-Flash~\cite{reduce_overthinking_2025} (reasoning-optimized “flash” variant tied to overthinking-reduction work),
\qwenicon~Qwen3-30B-A3B-Thinking-2507~\cite{qwen3technicalreport} (Qwen3 series, reasoning variant),
\dsicon~DeepSeek-R1-Distill-Qwen-1.5B~\cite{guo2025deepseek} (distilled reasoning model),
\gpticon~\mbox{gpt-oss-20b}~\cite{openai2025gptoss120bgptoss20bmodel} (OpenAI open-weight reasoning model; we use the default quantization, MXFP4, and, for prompting, rely on OpenAI Harmony, which defines three levels of reasoning effort),
\gairicon~LIMO-v2~\cite{ye2025limoreasoning} (data-efficient reasoning fine-tuned on curated traces),
\lgicon~EXAONE-4.0-1.2B~\cite{exaone-4.0} (hybrid non-reasoning/reasoning modes),
\nvidiaicon~OpenReasoning-Nemotron-1.5B~\cite{toshniwal2025genselect,moshkov2025aimo2winningsolutionbuilding,ahmad2025opencodereasoningiisimpletesttime,ahmad2025opencodereasoning} (open-weight small reasoning model),
\openicon~OpenThinker2-32B~\cite{guha2025openthoughtsdatarecipesreasoning} and \openicon~OpenThinker3-1.5B~\cite{guha2025openthoughtsdatarecipesreasoning} (trained on OpenThoughts2/3 data recipes).

To investigate the effect of the number of models required to reach a stable ranking with and without credible intervals, in addition to the 11 above-mentioned models, we extend the evaluation to 20 models in total (17 + 3):
\microsoft~Phi-4-reasoning and \microsoft~Phi-4-reasoning-plus~\cite{abdin2025phi4reasoning} (14B small language models with supervised “teachable” reasoning traces and an RL-enhanced variant),
\openr~OpenR1-Distill-7B~\cite{openr1} (an open 7B distillation of DeepSeek-R1 using fully public data),
\fuse~FuseO1-DeepSeekR1-QwQ-SkyT1-Flash-32B-Preview~\cite{wan2025fusechat} (System-II “long-short” reasoning fusion of DeepSeek-R1, QwQ, and Sky-T1-32B-Flash),
\qihoo~Light-R1-14B-DS~\cite{wen2025lightr1} (a Qwen2.5-based long-chain-of-thought model further improved with GRPO-style reinforcement learning),
\nvidiaicon~AceReason-Nemotron-1.1-7B~\cite{liu2025acereason} (7B NVIDIA Nemotron math/code model trained on OpenMathReasoning/OpenCodeReasoning data),
\nvidiaicon~NVIDIA-Nemotron-Nano-9B-v2~\cite{nvidia2025nemotronnano2} (a hybrid Mamba-Transformer “Nano 2” model with controllable reasoning mode),
\qwenicon~Qwen3-4B-Thinking-2507~\cite{qwen3technicalreport} (4B “thinking” variant of Qwen3 with scaled reasoning depth), and
\bespoke~Bespoke-Stratos-7B~\cite{bespoke_stratos} (Qwen2.5-7B student obtained via DeepSeek-R1-based reasoning distillation on Bespoke-Stratos-17k).

For verification, we additionally use \compassicon~CompassVerifier-3B~\cite{CompassVerifier}, a lightweight answer verifier suitable for outcome reward and equivalence checking.

\begin{table}[t]
\centering
\small
\begin{tabular}{c l l}
\hline
ID & Model & Short name \\
\hline
1  & \dsicon~DeepSeek-R1-Distill-Qwen-1.5B                         & DS-R1-Qwen              \\
2  & \gairicon~LIMO-v2                                             & LIMO-v2                 \\
3  & \openicon~OpenThinker2-32B                                    & OpenThinker2            \\
4  & \openicon~OpenThinker3-1.5B                                   & OpenThinker3            \\
5  & \qwenicon~Qwen3-30B-A3B-Thinking-2507                         & Qwen3-Thinking          \\
6  & \skyicon~Sky-T1-32B-Flash                                     & Sky-T1-Flash            \\
7  & \gpticon~gpt-oss-20b\_high                                    & gpt-oss-high            \\
8  & \gpticon~gpt-oss-20b\_low                                     & gpt-oss-low             \\
9  & \gpticon~gpt-oss-20b\_medium                                  & gpt-oss-medium          \\
10 & \lgicon~EXAONE-4.0-1.2B                                       & EXAONE-4.0              \\
11 & \nvidiaicon~OpenReasoning-Nemotron-1.5B                       & OR-Nemotron             \\
12 & \microsoft~Phi-4-reasoning                                    & Phi-4                   \\
13 & \microsoft~Phi-4-reasoning-plus                               & Phi-4-plus              \\
14 & \openr~OpenR1-Distill-7B                                      & OR1-Distill             \\
15 & \fuse~FuseO1-DeepSeekR1-QwQ-SkyT1-Flash-32B-Preview           & FuseO1-DS-QwQ-SkyT1     \\
16 & \qihoo~Light-R1-14B-DS                                        & Light-R1-DS             \\
17 & \nvidiaicon~AceReason-Nemotron-1.1-7B                         & AR-Nemotron             \\
18 & \nvidiaicon~NVIDIA-Nemotron-Nano-9B-v2                        & NVIDIA-Nemotron         \\
19 & \qwenicon~Qwen3-4B-Thinking-2507                              & Qwen3-4B                \\
20 & \bespoke~Bespoke-Stratos-7B                                   & Bespoke                 \\
\hline
\end{tabular}
\caption{Mapping between model IDs, full model names, and the shortened names used in figures and legends. Corresponding subsets are listed in Tables~\ref{app:tab:comb5}, \ref{app:tab:comb10}, and \ref{app:tab:comb15}.}
\label{app:tab:modelid}
\end{table}

\paragraph{Prompting.}
For most models, we follow the provider-recommended DeepSeek/Qwen-style prompt:
\emph{``Please reason step by step, and put your final answer within \texttt{\textbackslash boxed\{\}}.''} For \gpticon~\texttt{gpt-oss-20b}, we instead use the OpenAI Harmony prompt template, which provides three levels of reasoning effort. For \nvidiaicon~\texttt{OpenReasoning-Nemotron-1.5B}, we adopt the task-specific prompt:
\emph{``Solve the following math problem. Make sure to put the answer (and only the answer) inside \texttt{\textbackslash boxed\{\}}.''}

\subsection{Reproducibility}

\textbf{Sampling setup.} All trials use top-$p$ sampling with temperature~$0.6$, $p=0.95$, batch size~$1$, and seeds $1234$--$1313$. We perform $N=80$ trials per dataset $\times$ model.

\textbf{Verifier.} We use \compassicon~\texttt{CompassVerifier-3B} as a reward model. During evaluation, we leverage the model's scores on prompts generated by other models to create categorical schemas. We rely on the \texttt{Transformers}~\cite{wolf2019huggingface} and \texttt{Accelerate}~\cite{accelerate} libraries. To maximize throughput, we enable FlashAttention kernels~\cite{Dao2022FlashAttention} and adopt the \texttt{DFloat11} format~\cite{Zhang2025DFloat11}.

\textbf{Serving stack.} Token generation is served with \texttt{vLLM} (PagedAttention)~\cite{kwon2023efficient}, and models are loaded in \texttt{bf16} unless the release requires MXFP4 (e.g., \texttt{gpt-oss}). We record log-probabilities for both the input prompt and generated tokens, and cap \texttt{max\_tokens} at $32{,}768$.

\textbf{Hardware.} All runs execute on clusters with $8\times$ NVIDIA H200~(141GB).

\subsection*{Computational Cost and Token Statistics}
\label{sec:compute_stats}

\begin{figure}[htbp]
    \centering
    \includegraphics[width=\linewidth]{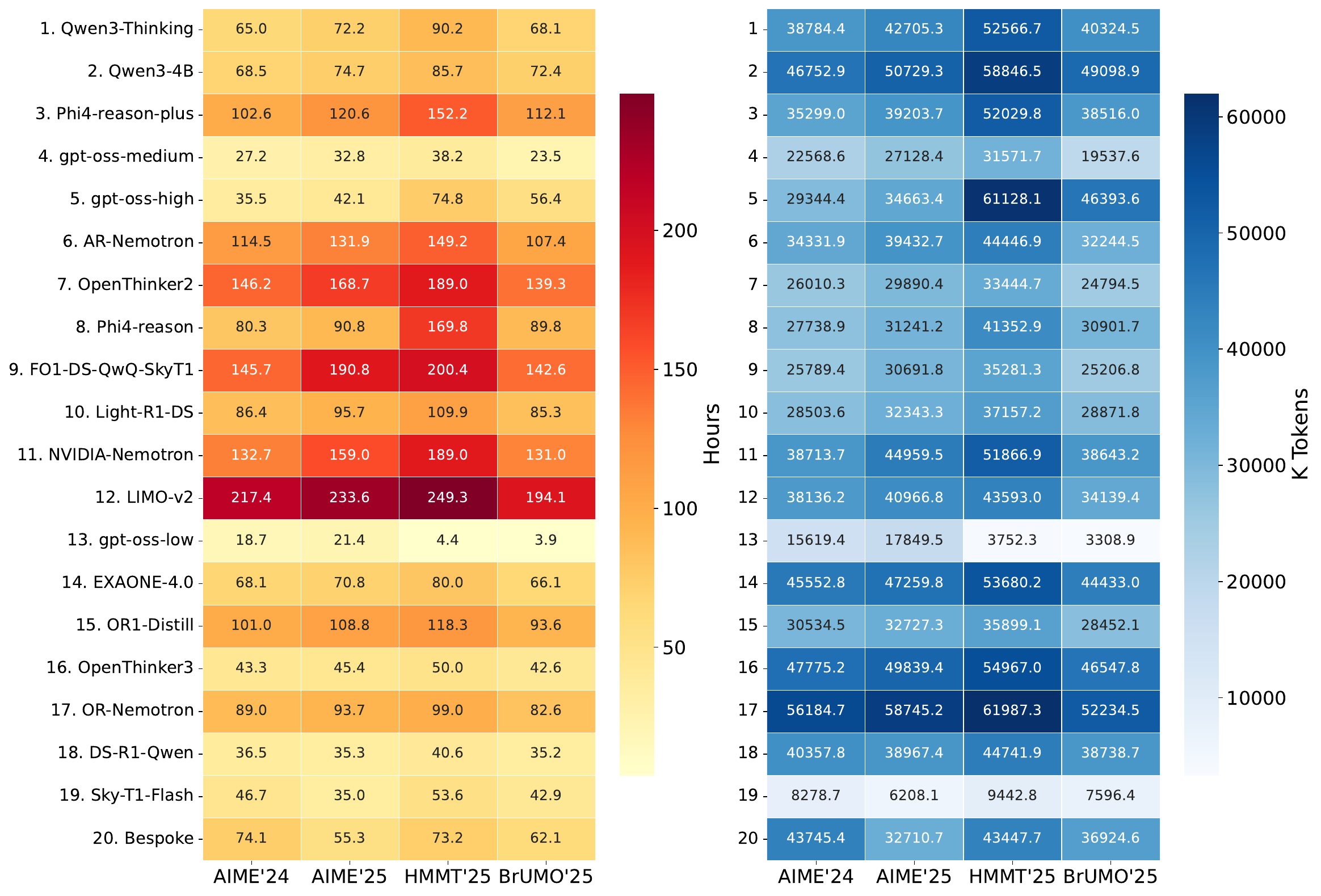}
    \caption{\textbf{Computational cost analysis.}
    (\textbf{Left}) Total inference time in hours aggregated over 80 trials and 30 questions per benchmark (2{,}400 inference runs per cell).
    (\textbf{Right}) Total number of completion tokens (in thousands) generated across the same runs.
    Models are ordered by overall performance (best to worst, top to bottom).}
    \label{fig:compcost}
\end{figure}

Across all tasks, we evaluated 20 models with 80 trials per model and 30 questions per benchmark, yielding a total of 192{,}000 independent inference runs. This required 7{,}445 GPU-hours ($\sim$310 GPU-days) and generated 2.96B tokens (2{,}963{,}318{,}176) in total (see \Cref{fig:compcost} for details).

\paragraph{Task-level computational cost.}

\begin{table}[h]
\centering
\begin{tabular}{lrr}
\toprule
\textbf{Task} & \textbf{Inference Time (hours)} & \textbf{Completion Tokens (M)} \\
\midrule
AIME'24  & 1{,}699.4 & 680.0 \\
AIME'25  & 1{,}878.4 & 728.3 \\
HMMT'25  & 2{,}216.5 & 851.2 \\
BrUMO'25 & 1{,}650.9 & 666.9 \\
\midrule
\textbf{TOTAL} & \textbf{7{,}445.2} & \textbf{2{,}926.4} \\
\bottomrule
\end{tabular}
\caption{Task-level computational cost aggregated over 20 models, 80 trials, 4 tasks, and 30 questions per task. Token counts correspond to \emph{completion} tokens only.}
\label{tab:computational_cost}
\end{table}

\hmmt~is the most expensive benchmark in terms of GPU time (2{,}217 GPU-hours), while \brumo~is the least expensive (1{,}651 GPU-hours). \Cref{fig:compcost} provides a complementary visualization of these patterns, showing inference time and completion-token usage across models and tasks.

\paragraph{Token breakdown.}

Aggregating across all tasks and models, the total number of tokens (prompt + completion) is 2.96B. The breakdown is:

\begin{itemize}
    \item Prompt tokens: 37M (1.2\%)
    \item Completion tokens: 2.93B (98.8\%)
    \item Average per query: 15{,}434 tokens
\end{itemize}

\paragraph{GPU-hours by model efficiency.}The 20 model configurations varied substantially in computational efficiency:
\begin{itemize}
    \item Most efficient: \texttt{gpt-oss-20b-low} (48.4 GPU-hours for 9{,}600 queries)
    \item Least efficient: \texttt{LIMO-v2} (894.3 GPU-hours for 9{,}600 queries)
    \item Average per query over all models: 139.6 seconds ($\sim$2.3 minutes)
\end{itemize}

\section{Convergence}\label{app:conv}

\begin{figure}[htbp]
  \centering
  \includegraphics[width=1.0\linewidth]
{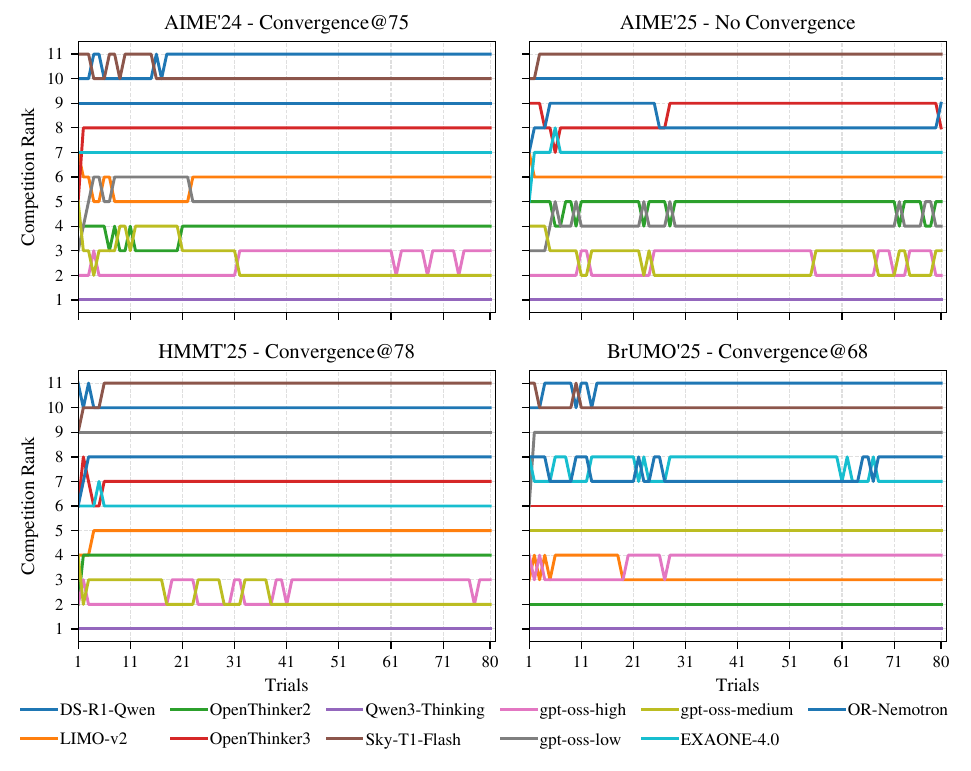}
  \caption{\textbf{Worst-case bootstrap rank trajectories}. Each line shows the ranking of a model as trials are added (11 models in total). Convergence is defined as the minimal $N$ after which the ranking remains unchanged. (a) \aimefour{}: converges at $N=75$. (b) \aimefive{}: no convergence observed within $80$ trials. (c) \hmmt{}: converges at $N=78$. (d) \brumo{}: converges at $N=68$.}
  \label{fig:combined_rankings_1}
\end{figure}

\begin{figure}[htbp]
  \centering
  \includegraphics[width=1.0\linewidth]
{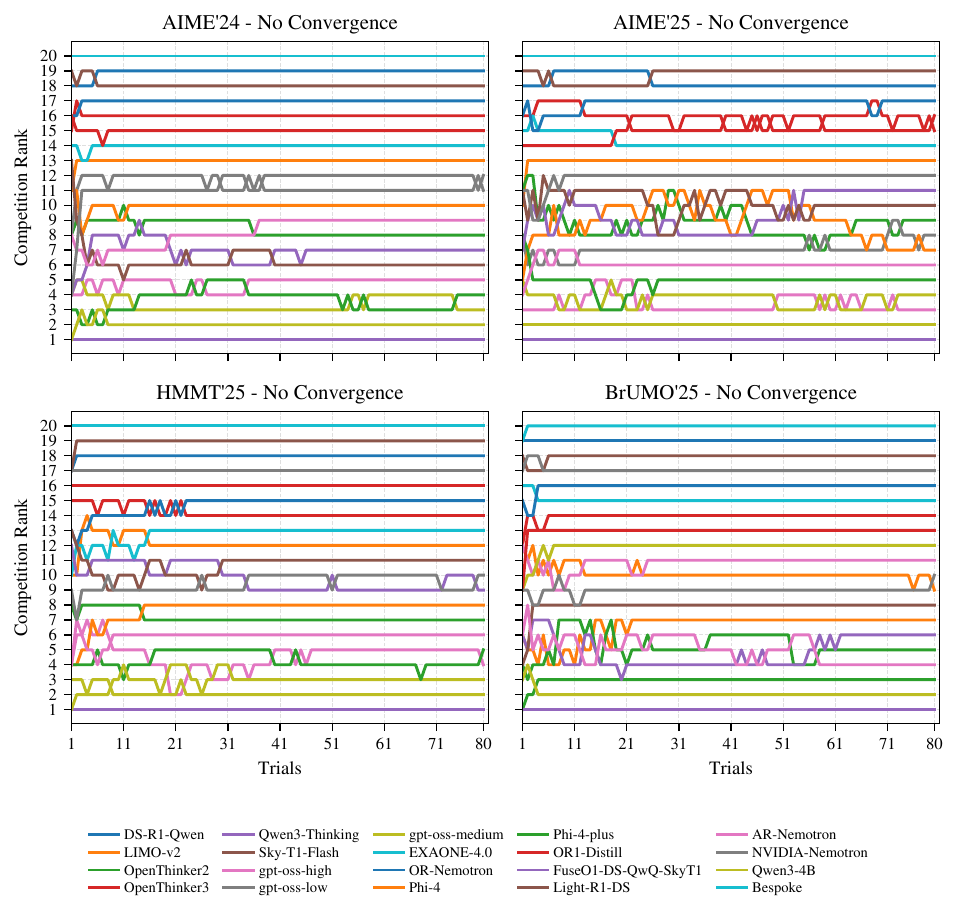}
  \caption{\textbf{Worst-case bootstrap rank trajectories}. Each line shows the ranking of a model as trials are added (20 models in total). Convergence is defined as the minimal $N$ after which the ranking remains unchanged. There is at least one \emph{no convergence} replicate among the $10^5$ bootstrapped replications.}
  \label{fig:combined_rankings_2}
\end{figure}

\begin{figure}[htbp]
  \centering
    \includegraphics[width=\linewidth]
  {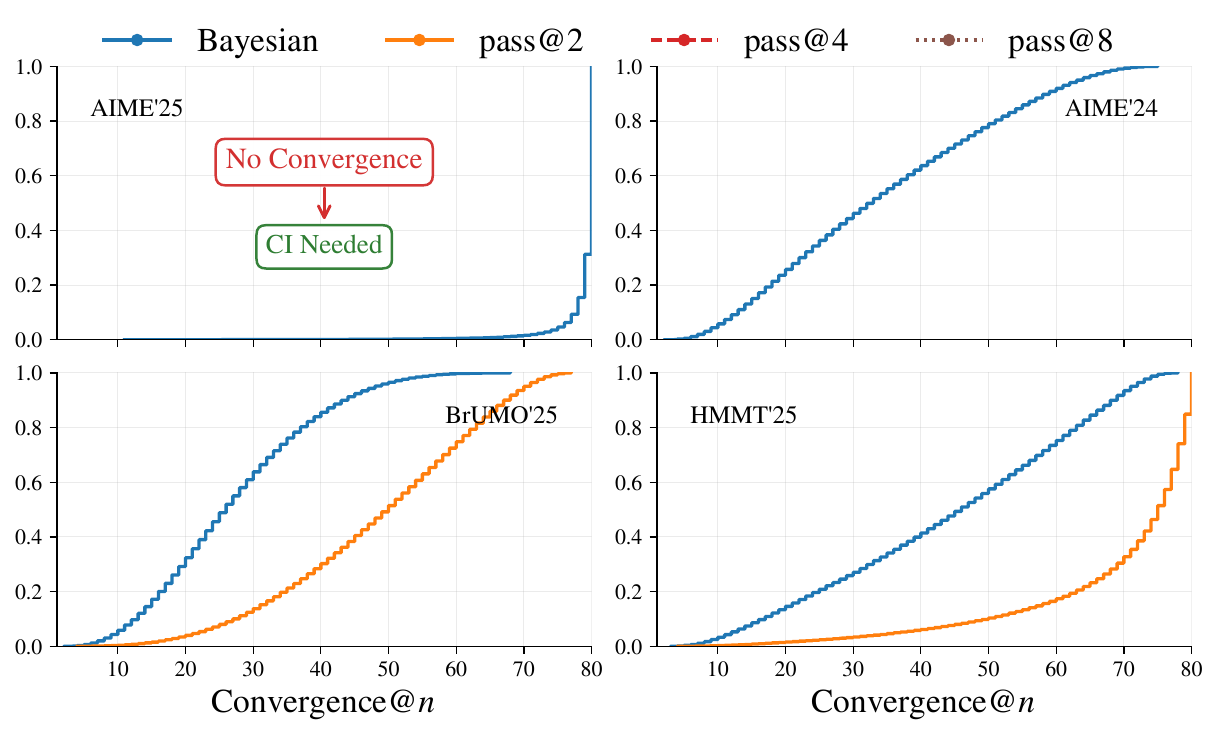}
  \caption{\textbf{CDF of convergence@$n$.} Complementing the PMFs in \Cref{fig:showcase}, these CDFs plot $P(k \!\le\! n)$ for the convergence threshold $k$ across \aimefour, \aimefive, \hmmt, and \brumo. Steeper and earlier rises indicate faster convergence. \bayes{N} accumulates mass with fewer trials than Pass@2/4/8, and on \aimeboth\ the Pass curves do not reach 1 by $N_{\max} = 80$. Greater convergence suggests that credible intervals should be reported for the evaluation tasks.}
  \label{fig:app:cdf}
\end{figure}

While \Cref{fig:showcase} shows the PMF of convergence@$n$, \Cref{fig:app:cdf} shows the corresponding cumulative distribution functions (CDFs). For \pass{4} and \pass{8}, there is no convergence, as the figure shows no CDFs associated with them. The CDFs are computed using the same bootstrap replicates as in \Cref{fig:showcase}. The distribution of convergence@$n$ is computed using the result matrices $R$ from the first 11 models (\Cref{app:tab:modelid}). Among the $10^5$ replications, \Cref{fig:combined_rankings_1} shows the worst-case scenarios in which convergence@$n$ attains its maximum value. As discussed in \Cref{par:wc}, convergence@$n$ depends on the number of models $L$: as $L$ increases, convergence@$n$ grows. When we extend the pool of LLMs from 11 to 20 models, convergence@$n$ reaches \emph{no convergence} for all datasets (see \Cref{fig:combined_rankings_2}).

\begin{figure}[t]
  \centering
  \includegraphics[width=\linewidth]{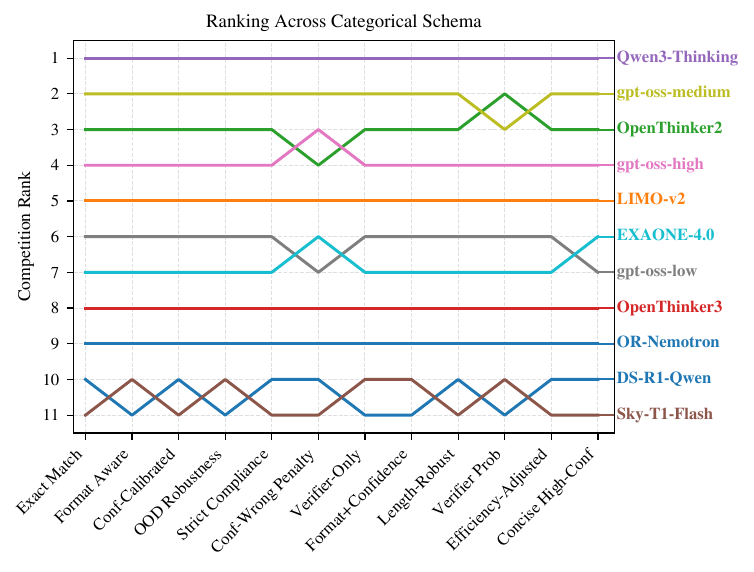}
\caption{Sensitivity of model rankings to the categorical scoring schema. For each schema variant (x-axis; see \cref{tab:cat_schema}), models are assigned a competition rank (y-axis; 1 = best). Colored trajectories track each model’s rank as the rubric changes, highlighting rank stability and crossovers.}
  \label{fig:cat:rank}
\end{figure}

To complement the worst-case trajectories discussed in \Cref{par:wc} and shown in \Cref{fig:combined_rankings_1,fig:combined_rankings_2}, we provide additional details on the construction of the model subsets and the resulting convergence behavior. \Cref{app:tab:modelid} lists the pool of 20 LLMs used in this analysis, together with the shortened identifiers that appear throughout the figures and tables. From this pool we construct 50 subsets of 5 models, 20 subsets of 10 models, and 20 subsets of 15 models, as summarized in \Cref{app:tab:comb5,app:tab:comb10,app:tab:comb15}. Each row in these tables corresponds to one subset, indicating which models are included and reporting, under each task, the convergence@$n$ metric computed \emph{without} a credible interval; each entry is the mean over \(10^{5}\) bootstrap replicates. Thus, the tables make explicit how convergence@$n$ depends not only on the task but also on the particular mixture of models being compared. Aggregating across all subsets and replicates, \Cref{fig:conv-comb} then visualizes the distribution of convergence@$n$ as a function of the number of models \(L\), confirming the trend anticipated in the main text: as \(L\) grows from 5 to 15 and ultimately to the full set of 20 LLMs, the required number of trials increases and non-convergence becomes common, indicating that rank-based evaluation methods such as \avg{N} and the \pass{k} family become increasingly unreliable without an accompanying Bayesian uncertainty quantification such as \bayes{N}.

\begin{table}[htbp]
\centering
\caption{\textbf{5-model combinations}. Matrix showing model presence across the 50 evaluated combinations.
Values under each task report the convergence$@n$ metric computed without a credible interval; each value is the mean of $10^5$ bootstrapped samples.
Model identifiers are listed in Table~\ref{app:tab:modelid}.}
\label{app:tab:comb5}
\small
\setlength{\tabcolsep}{3pt}
\begin{tabular}{c|cccccccccccccccccccc|rrrr}
\toprule
\textbf{Comb.} & \textbf{1} & \textbf{2} & \textbf{3} & \textbf{4} & \textbf{5} & \textbf{6} & \textbf{7} & \textbf{8} & \textbf{9} & \textbf{10} & \textbf{11} & \textbf{12} & \textbf{13} & \textbf{14} & \textbf{15} & \textbf{16} & \textbf{17} & \textbf{18} & \textbf{19} & \textbf{20} & \textbf{AIME'24} & \textbf{AIME'25} & \textbf{HMMT'25} & \textbf{BrUMO'25} \\
\midrule
\textbf{1} &  &  &  & \cellcolor{blue!15}4 &  &  &  &  &  &  &  &  & \cellcolor{blue!15}13 & \cellcolor{blue!15}14 &  &  &  & \cellcolor{blue!15}18 & \cellcolor{blue!15}19 &  & \cellcolor{green!40}\textbf{13.5} & \cellcolor{red!30}\textbf{69.0} & \cellcolor{green!40}\textbf{11.9} & \cellcolor{green!20}\textbf{8.8} \\
\textbf{2} & \cellcolor{blue!15}1 & \cellcolor{blue!15}2 &  &  &  &  &  &  &  &  &  &  &  & \cellcolor{blue!15}14 &  &  &  & \cellcolor{blue!15}18 &  & \cellcolor{blue!15}20 & \cellcolor{green!20}\textbf{3.0} & \cellcolor{orange!30}\textbf{38.4} & \cellcolor{green!40}\textbf{13.3} & \cellcolor{red!30}\textbf{59.3} \\
\textbf{3} & \cellcolor{blue!15}1 & \cellcolor{blue!15}2 &  & \cellcolor{blue!15}4 &  &  &  &  &  &  &  &  & \cellcolor{blue!15}13 &  &  &  &  &  &  & \cellcolor{blue!15}20 & \cellcolor{green!20}\textbf{1.5} & \cellcolor{green!20}\textbf{3.1} & \cellcolor{green!20}\textbf{2.7} & \cellcolor{green!20}\textbf{2.7} \\
\textbf{4} &  &  &  &  &  &  &  &  &  &  & \cellcolor{blue!15}11 &  & \cellcolor{blue!15}13 & \cellcolor{blue!15}14 &  &  &  & \cellcolor{blue!15}18 & \cellcolor{blue!15}19 &  & \cellcolor{green!40}\textbf{10.9} & \cellcolor{red!30}\textbf{65.0} & \cellcolor{green!40}\textbf{12.6} & \cellcolor{green!20}\textbf{5.0} \\
\textbf{5} & \cellcolor{blue!15}1 &  &  &  &  &  &  &  &  &  & \cellcolor{blue!15}11 &  & \cellcolor{blue!15}13 &  &  &  & \cellcolor{blue!15}17 &  &  & \cellcolor{blue!15}20 & \cellcolor{green!20}\textbf{2.5} & \cellcolor{green!20}\textbf{5.5} & \cellcolor{green!20}\textbf{3.7} & \cellcolor{red!30}\textbf{50.2} \\
\textbf{6} & \cellcolor{blue!15}1 &  &  &  &  &  &  &  &  & \cellcolor{blue!15}10 &  &  &  & \cellcolor{blue!15}14 &  &  & \cellcolor{blue!15}17 & \cellcolor{blue!15}18 &  &  & \cellcolor{green!40}\textbf{11.4} & \cellcolor{green!40}\textbf{15.3} & \cellcolor{green!20}\textbf{7.9} & \cellcolor{green!20}\textbf{3.1} \\
\textbf{7} & \cellcolor{blue!15}1 &  &  &  &  &  &  &  &  & \cellcolor{blue!15}10 &  & \cellcolor{blue!15}12 & \cellcolor{blue!15}13 &  &  &  &  &  & \cellcolor{blue!15}19 &  & \cellcolor{green!40}\textbf{11.0} & \cellcolor{green!20}\textbf{3.3} & \cellcolor{green!40}\textbf{12.5} & \cellcolor{green!40}\textbf{11.1} \\
\textbf{8} & \cellcolor{blue!15}1 &  &  & \cellcolor{blue!15}4 &  &  &  &  &  & \cellcolor{blue!15}10 &  &  &  &  &  &  & \cellcolor{blue!15}17 & \cellcolor{blue!15}18 &  &  & \cellcolor{green!20}\textbf{5.9} & \cellcolor{green!40}\textbf{13.1} & \cellcolor{green!20}\textbf{9.8} & \cellcolor{green!20}\textbf{6.3} \\
\textbf{9} &  &  &  &  &  &  &  &  &  &  & \cellcolor{blue!15}11 & \cellcolor{blue!15}12 &  &  &  &  & \cellcolor{blue!15}17 &  & \cellcolor{blue!15}19 & \cellcolor{blue!15}20 & \cellcolor{orange!50}\textbf{44.3} & \cellcolor{green!20}\textbf{2.8} & \cellcolor{green!20}\textbf{5.1} & \cellcolor{green!20}\textbf{9.4} \\
\textbf{10} & \cellcolor{blue!15}1 & \cellcolor{blue!15}2 &  &  &  &  &  &  &  &  &  &  & \cellcolor{blue!15}13 &  &  &  & \cellcolor{blue!15}17 & \cellcolor{blue!15}18 &  &  & \cellcolor{green!20}\textbf{6.8} & \cellcolor{orange!30}\textbf{38.8} & \cellcolor{green!40}\textbf{14.6} & \cellcolor{red!30}\textbf{68.4} \\
\textbf{11} &  &  &  &  &  &  &  &  &  & \cellcolor{blue!15}10 & \cellcolor{blue!15}11 &  &  & \cellcolor{blue!15}14 &  &  & \cellcolor{blue!15}17 &  & \cellcolor{blue!15}19 &  & \cellcolor{green!40}\textbf{10.4} & \cellcolor{red!30}\textbf{66.9} & \cellcolor{green!20}\textbf{5.3} & \cellcolor{green!40}\textbf{16.6} \\
\textbf{12} &  &  &  &  &  &  &  &  &  &  & \cellcolor{blue!15}11 &  & \cellcolor{blue!15}13 & \cellcolor{blue!15}14 &  &  & \cellcolor{blue!15}17 &  &  & \cellcolor{blue!15}20 & \cellcolor{green!20}\textbf{3.7} & \cellcolor{red!30}\textbf{65.1} & \cellcolor{green!20}\textbf{5.4} & \cellcolor{red!30}\textbf{50.3} \\
\textbf{13} &  &  &  & \cellcolor{blue!15}4 &  &  &  &  &  & \cellcolor{blue!15}10 & \cellcolor{blue!15}11 &  &  &  &  &  &  & \cellcolor{blue!15}18 & \cellcolor{blue!15}19 &  & \cellcolor{green!20}\textbf{9.2} & \cellcolor{red!30}\textbf{72.0} & \cellcolor{green!40}\textbf{17.2} & \cellcolor{green!40}\textbf{17.7} \\
\textbf{14} & \cellcolor{blue!15}1 & \cellcolor{blue!15}2 &  &  &  &  &  &  &  &  &  & \cellcolor{blue!15}12 & \cellcolor{blue!15}13 &  &  &  &  &  &  & \cellcolor{blue!15}20 & \cellcolor{green!20}\textbf{2.9} & \cellcolor{green!20}\textbf{4.4} & \cellcolor{green!20}\textbf{5.8} & \cellcolor{green!40}\textbf{11.5} \\
\textbf{15} &  &  &  & \cellcolor{blue!15}4 &  &  &  &  &  &  & \cellcolor{blue!15}11 & \cellcolor{blue!15}12 &  & \cellcolor{blue!15}14 &  &  &  &  &  & \cellcolor{blue!15}20 & \cellcolor{green!40}\textbf{12.9} & \cellcolor{red!30}\textbf{79.0} & \cellcolor{green!40}\textbf{15.0} & \cellcolor{green!20}\textbf{8.4} \\
\textbf{16} &  & \cellcolor{blue!15}2 &  &  &  &  &  &  &  & \cellcolor{blue!15}10 & \cellcolor{blue!15}11 &  &  &  &  &  &  & \cellcolor{blue!15}18 & \cellcolor{blue!15}19 &  & \cellcolor{green!20}\textbf{4.7} & \cellcolor{orange!50}\textbf{41.1} & \cellcolor{yellow!30}\textbf{27.4} & \cellcolor{red!30}\textbf{60.3} \\
\textbf{17} & \cellcolor{blue!15}1 & \cellcolor{blue!15}2 &  & \cellcolor{blue!15}4 &  &  &  &  &  &  &  &  &  &  &  &  & \cellcolor{blue!15}17 &  &  & \cellcolor{blue!15}20 & \cellcolor{green!20}\textbf{1.8} & \cellcolor{green!20}\textbf{3.2} & \cellcolor{green!20}\textbf{3.1} & \cellcolor{green!20}\textbf{2.3} \\
\textbf{18} & \cellcolor{blue!15}1 & \cellcolor{blue!15}2 &  & \cellcolor{blue!15}4 &  &  &  &  &  &  &  &  &  &  &  &  & \cellcolor{blue!15}17 & \cellcolor{blue!15}18 &  &  & \cellcolor{green!20}\textbf{6.2} & \cellcolor{orange!30}\textbf{38.3} & \cellcolor{green!40}\textbf{14.3} & \cellcolor{red!30}\textbf{59.4} \\
\textbf{19} & \cellcolor{blue!15}1 & \cellcolor{blue!15}2 &  &  &  &  &  &  &  &  &  & \cellcolor{blue!15}12 &  &  &  &  & \cellcolor{blue!15}17 &  & \cellcolor{blue!15}19 &  & \cellcolor{orange!50}\textbf{44.3} & \cellcolor{green!20}\textbf{4.2} & \cellcolor{green!20}\textbf{8.1} & \cellcolor{green!40}\textbf{11.1} \\
\textbf{20} & \cellcolor{blue!15}1 &  &  &  &  &  &  &  &  & \cellcolor{blue!15}10 & \cellcolor{blue!15}11 &  &  &  &  &  &  & \cellcolor{blue!15}18 &  & \cellcolor{blue!15}20 & \cellcolor{green!20}\textbf{1.8} & \cellcolor{green!40}\textbf{13.1} & \cellcolor{green!20}\textbf{7.8} & \cellcolor{green!40}\textbf{15.0} \\
\textbf{21} & \cellcolor{blue!15}1 &  &  &  &  &  &  &  &  & \cellcolor{blue!15}10 &  & \cellcolor{blue!15}12 &  &  &  &  &  & \cellcolor{blue!15}18 &  & \cellcolor{blue!15}20 & \cellcolor{green!20}\textbf{5.4} & \cellcolor{green!20}\textbf{4.1} & \cellcolor{green!40}\textbf{16.4} & \cellcolor{green!20}\textbf{3.1} \\
\textbf{22} &  & \cellcolor{blue!15}2 &  &  &  &  &  &  &  &  &  & \cellcolor{blue!15}12 &  & \cellcolor{blue!15}14 &  &  & \cellcolor{blue!15}17 &  &  & \cellcolor{blue!15}20 & \cellcolor{orange!50}\textbf{44.3} & \cellcolor{green!20}\textbf{4.1} & \cellcolor{green!20}\textbf{7.9} & \cellcolor{green!20}\textbf{9.8} \\
\textbf{23} &  &  &  & \cellcolor{blue!15}4 &  &  &  &  &  &  & \cellcolor{blue!15}11 &  & \cellcolor{blue!15}13 &  &  &  & \cellcolor{blue!15}17 &  & \cellcolor{blue!15}19 &  & \cellcolor{green!40}\textbf{14.7} & \cellcolor{red!30}\textbf{71.2} & \cellcolor{green!40}\textbf{19.7} & \cellcolor{red!30}\textbf{50.9} \\
\textbf{24} & \cellcolor{blue!15}1 & \cellcolor{blue!15}2 &  & \cellcolor{blue!15}4 &  &  &  &  &  &  &  &  &  & \cellcolor{blue!15}14 &  &  &  & \cellcolor{blue!15}18 &  &  & \cellcolor{green!20}\textbf{8.4} & \cellcolor{red!30}\textbf{72.3} & \cellcolor{green!40}\textbf{14.0} & \cellcolor{red!30}\textbf{59.8} \\
\textbf{25} &  &  &  & \cellcolor{blue!15}4 &  &  &  &  &  & \cellcolor{blue!15}10 &  & \cellcolor{blue!15}12 &  &  &  &  &  & \cellcolor{blue!15}18 &  & \cellcolor{blue!15}20 & \cellcolor{green!20}\textbf{6.3} & \cellcolor{green!40}\textbf{13.6} & \cellcolor{green!40}\textbf{17.1} & \cellcolor{green!20}\textbf{7.0} \\
\textbf{26} & \cellcolor{blue!15}1 &  &  & \cellcolor{blue!15}4 &  &  &  &  &  &  &  &  &  &  &  &  &  & \cellcolor{blue!15}18 & \cellcolor{blue!15}19 & \cellcolor{blue!15}20 & \cellcolor{green!20}\textbf{1.4} & \cellcolor{green!20}\textbf{2.9} & \cellcolor{green!20}\textbf{2.0} & \cellcolor{green!20}\textbf{1.8} \\
\textbf{27} &  &  &  &  &  &  &  &  &  & \cellcolor{blue!15}10 &  &  &  & \cellcolor{blue!15}14 &  &  & \cellcolor{blue!15}17 &  & \cellcolor{blue!15}19 & \cellcolor{blue!15}20 & \cellcolor{green!20}\textbf{9.5} & \cellcolor{green!40}\textbf{15.3} & \cellcolor{green!20}\textbf{2.5} & \cellcolor{green!20}\textbf{5.2} \\
\textbf{28} &  &  &  & \cellcolor{blue!15}4 &  &  &  &  &  &  &  &  &  &  &  &  & \cellcolor{blue!15}17 & \cellcolor{blue!15}18 & \cellcolor{blue!15}19 & \cellcolor{blue!15}20 & \cellcolor{green!20}\textbf{5.4} & \cellcolor{green!20}\textbf{1.6} & \cellcolor{green!20}\textbf{3.3} & \cellcolor{green!20}\textbf{4.5} \\
\textbf{29} &  &  &  &  &  &  &  &  &  & \cellcolor{blue!15}10 &  & \cellcolor{blue!15}12 &  &  &  &  & \cellcolor{blue!15}17 & \cellcolor{blue!15}18 &  & \cellcolor{blue!15}20 & \cellcolor{orange!50}\textbf{45.1} & \cellcolor{green!20}\textbf{4.1} & \cellcolor{green!40}\textbf{17.5} & \cellcolor{green!20}\textbf{8.4} \\
\textbf{30} &  & \cellcolor{blue!15}2 &  &  &  &  &  &  &  & \cellcolor{blue!15}10 & \cellcolor{blue!15}11 &  &  &  &  &  & \cellcolor{blue!15}17 & \cellcolor{blue!15}18 &  &  & \cellcolor{green!20}\textbf{7.3} & \cellcolor{orange!50}\textbf{41.1} & \cellcolor{yellow!30}\textbf{27.5} & \cellcolor{red!30}\textbf{60.3} \\
\textbf{31} &  &  & \cellcolor{blue!15}3 &  & \cellcolor{blue!15}5 &  & \cellcolor{blue!15}7 &  & \cellcolor{blue!15}9 &  &  &  &  &  &  & \cellcolor{blue!15}16 &  &  &  &  & \cellcolor{orange!30}\textbf{39.6} & \cellcolor{red!30}\textbf{73.6} & \cellcolor{orange!30}\textbf{38.1} & \cellcolor{green!40}\textbf{13.1} \\
\textbf{32} &  &  & \cellcolor{blue!15}3 &  &  & \cellcolor{blue!15}6 &  & \cellcolor{blue!15}8 &  &  &  &  &  &  & \cellcolor{blue!15}15 & \cellcolor{blue!15}16 &  &  &  &  & \cellcolor{orange!50}\textbf{48.0} & \cellcolor{red!30}\textbf{71.4} & \cellcolor{orange!30}\textbf{32.9} & \cellcolor{green!40}\textbf{18.6} \\
\textbf{33} &  &  &  &  & \cellcolor{blue!15}5 & \cellcolor{blue!15}6 & \cellcolor{blue!15}7 & \cellcolor{blue!15}8 &  &  &  &  &  &  &  & \cellcolor{blue!15}16 &  &  &  &  & \cellcolor{green!40}\textbf{19.5} & \cellcolor{orange!30}\textbf{39.5} & \cellcolor{green!20}\textbf{1.6} & \cellcolor{green!40}\textbf{10.4} \\
\textbf{34} &  &  & \cellcolor{blue!15}3 &  & \cellcolor{blue!15}5 &  &  & \cellcolor{blue!15}8 & \cellcolor{blue!15}9 &  &  &  &  &  &  & \cellcolor{blue!15}16 &  &  &  &  & \cellcolor{yellow!30}\textbf{23.9} & \cellcolor{red!30}\textbf{67.8} & \cellcolor{green!20}\textbf{6.5} & \cellcolor{green!20}\textbf{3.3} \\
\textbf{35} &  &  &  &  &  & \cellcolor{blue!15}6 & \cellcolor{blue!15}7 & \cellcolor{blue!15}8 & \cellcolor{blue!15}9 &  &  &  &  &  &  & \cellcolor{blue!15}16 &  &  &  &  & \cellcolor{orange!30}\textbf{35.7} & \cellcolor{red!30}\textbf{73.1} & \cellcolor{orange!30}\textbf{36.9} & \cellcolor{green!40}\textbf{17.1} \\
\textbf{36} &  &  & \cellcolor{blue!15}3 &  & \cellcolor{blue!15}5 & \cellcolor{blue!15}6 & \cellcolor{blue!15}7 & \cellcolor{blue!15}8 &  &  &  &  &  &  &  &  &  &  &  &  & \cellcolor{green!40}\textbf{10.2} & \cellcolor{red!30}\textbf{61.1} & \cellcolor{green!20}\textbf{2.3} & \cellcolor{green!40}\textbf{10.2} \\
\textbf{37} &  &  &  &  & \cellcolor{blue!15}5 &  & \cellcolor{blue!15}7 &  & \cellcolor{blue!15}9 &  &  &  &  &  & \cellcolor{blue!15}15 & \cellcolor{blue!15}16 &  &  &  &  & \cellcolor{orange!50}\textbf{47.3} & \cellcolor{red!30}\textbf{74.7} & \cellcolor{orange!50}\textbf{47.4} & \cellcolor{green!40}\textbf{16.6} \\
\textbf{38} &  &  & \cellcolor{blue!15}3 &  & \cellcolor{blue!15}5 &  & \cellcolor{blue!15}7 & \cellcolor{blue!15}8 & \cellcolor{blue!15}9 &  &  &  &  &  &  &  &  &  &  &  & \cellcolor{yellow!30}\textbf{29.6} & \cellcolor{red!30}\textbf{75.4} & \cellcolor{orange!30}\textbf{37.1} & \cellcolor{green!40}\textbf{12.1} \\
\textbf{39} &  &  &  &  & \cellcolor{blue!15}5 & \cellcolor{blue!15}6 &  &  & \cellcolor{blue!15}9 &  &  &  &  &  & \cellcolor{blue!15}15 & \cellcolor{blue!15}16 &  &  &  &  & \cellcolor{orange!30}\textbf{35.6} & \cellcolor{orange!50}\textbf{47.0} & \cellcolor{yellow!30}\textbf{28.6} & \cellcolor{green!40}\textbf{10.1} \\
\textbf{40} &  &  & \cellcolor{blue!15}3 &  &  & \cellcolor{blue!15}6 &  & \cellcolor{blue!15}8 & \cellcolor{blue!15}9 &  &  &  &  &  &  & \cellcolor{blue!15}16 &  &  &  &  & \cellcolor{yellow!30}\textbf{23.9} & \cellcolor{red!30}\textbf{67.8} & \cellcolor{green!20}\textbf{6.4} & \cellcolor{green!40}\textbf{11.0} \\
\textbf{41} &  &  & \cellcolor{blue!15}3 &  &  & \cellcolor{blue!15}6 & \cellcolor{blue!15}7 & \cellcolor{blue!15}8 &  &  &  &  &  &  & \cellcolor{blue!15}15 &  &  &  &  &  & \cellcolor{orange!30}\textbf{36.0} & \cellcolor{red!30}\textbf{64.3} & \cellcolor{green!40}\textbf{10.9} & \cellcolor{green!40}\textbf{13.9} \\
\textbf{42} &  &  &  &  &  & \cellcolor{blue!15}6 & \cellcolor{blue!15}7 & \cellcolor{blue!15}8 &  &  &  &  &  &  & \cellcolor{blue!15}15 & \cellcolor{blue!15}16 &  &  &  &  & \cellcolor{orange!50}\textbf{40.5} & \cellcolor{red!30}\textbf{59.1} & \cellcolor{yellow!30}\textbf{28.6} & \cellcolor{green!40}\textbf{15.8} \\
\textbf{43} &  &  & \cellcolor{blue!15}3 &  & \cellcolor{blue!15}5 & \cellcolor{blue!15}6 &  &  &  &  &  &  &  &  & \cellcolor{blue!15}15 & \cellcolor{blue!15}16 &  &  &  &  & \cellcolor{orange!50}\textbf{47.8} & \cellcolor{red!30}\textbf{60.1} & \cellcolor{orange!30}\textbf{32.9} & \cellcolor{green!40}\textbf{14.7} \\
\textbf{44} &  &  & \cellcolor{blue!15}3 &  & \cellcolor{blue!15}5 &  & \cellcolor{blue!15}7 &  & \cellcolor{blue!15}9 &  &  &  &  &  & \cellcolor{blue!15}15 &  &  &  &  &  & \cellcolor{orange!50}\textbf{43.0} & \cellcolor{red!30}\textbf{72.5} & \cellcolor{orange!30}\textbf{39.1} & \cellcolor{green!40}\textbf{16.0} \\
\textbf{45} &  &  &  &  & \cellcolor{blue!15}5 &  & \cellcolor{blue!15}7 & \cellcolor{blue!15}8 & \cellcolor{blue!15}9 &  &  &  &  &  & \cellcolor{blue!15}15 &  &  &  &  &  & \cellcolor{orange!30}\textbf{31.3} & \cellcolor{red!30}\textbf{72.6} & \cellcolor{orange!30}\textbf{36.9} & \cellcolor{green!40}\textbf{12.1} \\
\textbf{46} &  &  & \cellcolor{blue!15}3 &  & \cellcolor{blue!15}5 & \cellcolor{blue!15}6 &  & \cellcolor{blue!15}8 &  &  &  &  &  &  &  & \cellcolor{blue!15}16 &  &  &  &  & \cellcolor{green!40}\textbf{19.8} & \cellcolor{red!30}\textbf{67.8} & \cellcolor{green!20}\textbf{6.0} & \cellcolor{green!40}\textbf{11.0} \\
\textbf{47} &  &  & \cellcolor{blue!15}3 &  & \cellcolor{blue!15}5 & \cellcolor{blue!15}6 &  & \cellcolor{blue!15}8 &  &  &  &  &  &  & \cellcolor{blue!15}15 &  &  &  &  &  & \cellcolor{orange!30}\textbf{33.0} & \cellcolor{red!30}\textbf{64.3} & \cellcolor{green!40}\textbf{10.3} & \cellcolor{green!40}\textbf{13.9} \\
\textbf{48} &  &  & \cellcolor{blue!15}3 &  &  & \cellcolor{blue!15}6 & \cellcolor{blue!15}7 &  & \cellcolor{blue!15}9 &  &  &  &  &  &  & \cellcolor{blue!15}16 &  &  &  &  & \cellcolor{orange!30}\textbf{39.6} & \cellcolor{red!30}\textbf{73.6} & \cellcolor{orange!30}\textbf{38.1} & \cellcolor{green!40}\textbf{13.1} \\
\textbf{49} &  &  &  &  & \cellcolor{blue!15}5 & \cellcolor{blue!15}6 & \cellcolor{blue!15}7 &  &  &  &  &  &  &  & \cellcolor{blue!15}15 & \cellcolor{blue!15}16 &  &  &  &  & \cellcolor{orange!30}\textbf{39.9} & \cellcolor{orange!50}\textbf{47.0} & \cellcolor{yellow!30}\textbf{28.6} & \cellcolor{green!40}\textbf{10.5} \\
\textbf{50} &  &  & \cellcolor{blue!15}3 &  &  & \cellcolor{blue!15}6 & \cellcolor{blue!15}7 & \cellcolor{blue!15}8 &  &  &  &  &  &  &  & \cellcolor{blue!15}16 &  &  &  &  & \cellcolor{yellow!30}\textbf{29.8} & \cellcolor{red!30}\textbf{67.8} & \cellcolor{green!20}\textbf{6.7} & \cellcolor{green!40}\textbf{11.2} \\
\bottomrule
\end{tabular}
\end{table}

\begin{table}[htbp]
\centering
\caption{\textbf{10-model combinations}. Matrix showing model presence across the 20 evaluated combinations.
Values under each task report the convergence$@n$ metric computed without a credible interval; each value is the mean of $10^5$ bootstrapped samples.
Model identifiers are listed in Table~\ref{app:tab:modelid}.}
\label{app:tab:comb10}
\small
\setlength{\tabcolsep}{3pt}
\begin{tabular}{c|cccccccccccccccccccc|rrrr}
\toprule
\textbf{Comb.} & \textbf{1} & \textbf{2} & \textbf{3} & \textbf{4} & \textbf{5} & \textbf{6} & \textbf{7} & \textbf{8} & \textbf{9} & \textbf{10} & \textbf{11} & \textbf{12} & \textbf{13} & \textbf{14} & \textbf{15} & \textbf{16} & \textbf{17} & \textbf{18} & \textbf{19} & \textbf{20} & \textbf{AIME'24} & \textbf{AIME'25} & \textbf{HMMT'25} & \textbf{BrUMO'25} \\
\midrule
\textbf{1} & \cellcolor{blue!15}1 & \cellcolor{blue!15}2 & \cellcolor{blue!15}3 & \cellcolor{blue!15}4 & \cellcolor{blue!15}5 & \cellcolor{blue!15}6 &  &  &  &  &  &  &  & \cellcolor{blue!15}14 &  & \cellcolor{blue!15}16 &  &  & \cellcolor{blue!15}19 & \cellcolor{blue!15}20 & \cellcolor{yellow!30}\textbf{28.2} & \cellcolor{red!30}\textbf{71.9} & \cellcolor{yellow!30}\textbf{23.3} & \cellcolor{orange!30}\textbf{35.4} \\
\textbf{2} & \cellcolor{blue!15}1 & \cellcolor{blue!15}2 &  & \cellcolor{blue!15}4 &  &  &  &  &  &  & \cellcolor{blue!15}11 &  &  & \cellcolor{blue!15}14 &  & \cellcolor{blue!15}16 & \cellcolor{blue!15}17 & \cellcolor{blue!15}18 & \cellcolor{blue!15}19 & \cellcolor{blue!15}20 & \cellcolor{green!40}\textbf{16.4} & \cellcolor{red!30}\textbf{79.0} & \cellcolor{orange!50}\textbf{43.2} & \cellcolor{red!30}\textbf{60.4} \\
\textbf{3} &  &  & \cellcolor{blue!15}3 &  & \cellcolor{blue!15}5 & \cellcolor{blue!15}6 & \cellcolor{blue!15}7 & \cellcolor{blue!15}8 & \cellcolor{blue!15}9 &  &  & \cellcolor{blue!15}12 & \cellcolor{blue!15}13 &  & \cellcolor{blue!15}15 & \cellcolor{blue!15}16 &  &  &  &  & \cellcolor{red!30}\textbf{68.4} & \cellcolor{red!30}\textbf{79.3} & \cellcolor{red!30}\textbf{71.3} & \cellcolor{red!30}\textbf{73.5} \\
\textbf{4} &  & \cellcolor{blue!15}2 &  &  &  &  & \cellcolor{blue!15}7 & \cellcolor{blue!15}8 & \cellcolor{blue!15}9 &  & \cellcolor{blue!15}11 & \cellcolor{blue!15}12 & \cellcolor{blue!15}13 &  &  & \cellcolor{blue!15}16 & \cellcolor{blue!15}17 & \cellcolor{blue!15}18 &  &  & \cellcolor{red!30}\textbf{76.5} & \cellcolor{red!30}\textbf{79.0} & \cellcolor{red!30}\textbf{70.3} & \cellcolor{red!30}\textbf{74.8} \\
\textbf{5} & \cellcolor{blue!15}1 & \cellcolor{blue!15}2 &  & \cellcolor{blue!15}4 &  &  &  &  &  & \cellcolor{blue!15}10 & \cellcolor{blue!15}11 &  &  & \cellcolor{blue!15}14 &  &  & \cellcolor{blue!15}17 & \cellcolor{blue!15}18 & \cellcolor{blue!15}19 & \cellcolor{blue!15}20 & \cellcolor{green!40}\textbf{18.1} & \cellcolor{red!30}\textbf{79.0} & \cellcolor{orange!30}\textbf{30.6} & \cellcolor{red!30}\textbf{60.9} \\
\textbf{6} & \cellcolor{blue!15}1 & \cellcolor{blue!15}2 &  & \cellcolor{blue!15}4 &  &  &  &  &  &  & \cellcolor{blue!15}11 & \cellcolor{blue!15}12 & \cellcolor{blue!15}13 &  &  & \cellcolor{blue!15}16 & \cellcolor{blue!15}17 &  & \cellcolor{blue!15}19 & \cellcolor{blue!15}20 & \cellcolor{orange!50}\textbf{47.9} & \cellcolor{red!30}\textbf{72.8} & \cellcolor{orange!30}\textbf{30.5} & \cellcolor{red!30}\textbf{70.0} \\
\textbf{7} &  &  & \cellcolor{blue!15}3 &  & \cellcolor{blue!15}5 & \cellcolor{blue!15}6 & \cellcolor{blue!15}7 & \cellcolor{blue!15}8 &  &  &  & \cellcolor{blue!15}12 & \cellcolor{blue!15}13 & \cellcolor{blue!15}14 & \cellcolor{blue!15}15 & \cellcolor{blue!15}16 &  &  &  &  & \cellcolor{red!30}\textbf{56.6} & \cellcolor{red!30}\textbf{78.3} & \cellcolor{red!30}\textbf{68.7} & \cellcolor{red!30}\textbf{73.5} \\
\textbf{8} & \cellcolor{blue!15}1 &  & \cellcolor{blue!15}3 & \cellcolor{blue!15}4 & \cellcolor{blue!15}5 & \cellcolor{blue!15}6 & \cellcolor{blue!15}7 & \cellcolor{blue!15}8 & \cellcolor{blue!15}9 &  &  &  &  &  & \cellcolor{blue!15}15 &  &  &  & \cellcolor{blue!15}19 &  & \cellcolor{orange!50}\textbf{46.3} & \cellcolor{red!30}\textbf{75.8} & \cellcolor{orange!50}\textbf{49.4} & \cellcolor{orange!30}\textbf{38.6} \\
\textbf{9} & \cellcolor{blue!15}1 & \cellcolor{blue!15}2 &  & \cellcolor{blue!15}4 &  &  &  &  &  & \cellcolor{blue!15}10 & \cellcolor{blue!15}11 & \cellcolor{blue!15}12 & \cellcolor{blue!15}13 &  &  & \cellcolor{blue!15}16 &  & \cellcolor{blue!15}18 & \cellcolor{blue!15}19 &  & \cellcolor{yellow!30}\textbf{20.8} & \cellcolor{red!30}\textbf{74.4} & \cellcolor{orange!50}\textbf{48.1} & \cellcolor{red!30}\textbf{73.0} \\
\textbf{10} &  &  &  &  & \cellcolor{blue!15}5 &  & \cellcolor{blue!15}7 & \cellcolor{blue!15}8 & \cellcolor{blue!15}9 &  & \cellcolor{blue!15}11 & \cellcolor{blue!15}12 & \cellcolor{blue!15}13 &  &  &  & \cellcolor{blue!15}17 & \cellcolor{blue!15}18 & \cellcolor{blue!15}19 &  & \cellcolor{red!30}\textbf{76.5} & \cellcolor{red!30}\textbf{78.9} & \cellcolor{red!30}\textbf{70.1} & \cellcolor{red!30}\textbf{53.6} \\
\textbf{11} & \cellcolor{blue!15}1 & \cellcolor{blue!15}2 &  &  &  &  &  &  &  &  & \cellcolor{blue!15}11 & \cellcolor{blue!15}12 & \cellcolor{blue!15}13 & \cellcolor{blue!15}14 &  & \cellcolor{blue!15}16 & \cellcolor{blue!15}17 &  & \cellcolor{blue!15}19 & \cellcolor{blue!15}20 & \cellcolor{orange!50}\textbf{47.6} & \cellcolor{red!30}\textbf{67.5} & \cellcolor{yellow!30}\textbf{27.2} & \cellcolor{red!30}\textbf{70.0} \\
\textbf{12} & \cellcolor{blue!15}1 & \cellcolor{blue!15}2 & \cellcolor{blue!15}3 &  & \cellcolor{blue!15}5 & \cellcolor{blue!15}6 &  &  &  &  &  &  & \cellcolor{blue!15}13 & \cellcolor{blue!15}14 & \cellcolor{blue!15}15 & \cellcolor{blue!15}16 & \cellcolor{blue!15}17 &  &  &  & \cellcolor{red!30}\textbf{51.3} & \cellcolor{red!30}\textbf{60.3} & \cellcolor{orange!50}\textbf{40.6} & \cellcolor{red!30}\textbf{71.9} \\
\textbf{13} &  &  & \cellcolor{blue!15}3 &  & \cellcolor{blue!15}5 &  & \cellcolor{blue!15}7 & \cellcolor{blue!15}8 & \cellcolor{blue!15}9 &  &  & \cellcolor{blue!15}12 & \cellcolor{blue!15}13 &  &  & \cellcolor{blue!15}16 &  & \cellcolor{blue!15}18 &  & \cellcolor{blue!15}20 & \cellcolor{red!30}\textbf{75.9} & \cellcolor{red!30}\textbf{79.3} & \cellcolor{red!30}\textbf{71.5} & \cellcolor{red!30}\textbf{63.6} \\
\textbf{14} & \cellcolor{blue!15}1 & \cellcolor{blue!15}2 & \cellcolor{blue!15}3 & \cellcolor{blue!15}4 &  &  &  &  &  & \cellcolor{blue!15}10 & \cellcolor{blue!15}11 & \cellcolor{blue!15}12 & \cellcolor{blue!15}13 &  &  & \cellcolor{blue!15}16 &  &  & \cellcolor{blue!15}19 &  & \cellcolor{yellow!30}\textbf{29.0} & \cellcolor{red!30}\textbf{76.4} & \cellcolor{red!30}\textbf{50.8} & \cellcolor{red!30}\textbf{66.0} \\
\textbf{15} &  & \cellcolor{blue!15}2 & \cellcolor{blue!15}3 & \cellcolor{blue!15}4 & \cellcolor{blue!15}5 & \cellcolor{blue!15}6 & \cellcolor{blue!15}7 & \cellcolor{blue!15}8 & \cellcolor{blue!15}9 &  &  &  &  &  & \cellcolor{blue!15}15 &  &  &  & \cellcolor{blue!15}19 &  & \cellcolor{orange!50}\textbf{45.0} & \cellcolor{red!30}\textbf{75.8} & \cellcolor{orange!50}\textbf{49.6} & \cellcolor{orange!30}\textbf{38.8} \\
\textbf{16} & \cellcolor{blue!15}1 & \cellcolor{blue!15}2 &  & \cellcolor{blue!15}4 &  &  &  &  &  &  & \cellcolor{blue!15}11 & \cellcolor{blue!15}12 &  & \cellcolor{blue!15}14 &  &  & \cellcolor{blue!15}17 & \cellcolor{blue!15}18 & \cellcolor{blue!15}19 & \cellcolor{blue!15}20 & \cellcolor{orange!50}\textbf{46.4} & \cellcolor{red!30}\textbf{79.0} & \cellcolor{yellow!30}\textbf{26.3} & \cellcolor{red!30}\textbf{60.4} \\
\textbf{17} & \cellcolor{blue!15}1 & \cellcolor{blue!15}2 &  & \cellcolor{blue!15}4 &  &  & \cellcolor{blue!15}7 & \cellcolor{blue!15}8 & \cellcolor{blue!15}9 & \cellcolor{blue!15}10 & \cellcolor{blue!15}11 &  &  &  &  &  &  & \cellcolor{blue!15}18 & \cellcolor{blue!15}19 &  & \cellcolor{red!30}\textbf{70.3} & \cellcolor{red!30}\textbf{78.8} & \cellcolor{red!30}\textbf{51.8} & \cellcolor{red!30}\textbf{61.9} \\
\textbf{18} &  &  &  &  & \cellcolor{blue!15}5 & \cellcolor{blue!15}6 & \cellcolor{blue!15}7 & \cellcolor{blue!15}8 & \cellcolor{blue!15}9 & \cellcolor{blue!15}10 &  & \cellcolor{blue!15}12 & \cellcolor{blue!15}13 &  &  & \cellcolor{blue!15}16 &  & \cellcolor{blue!15}18 &  &  & \cellcolor{red!30}\textbf{75.9} & \cellcolor{red!30}\textbf{78.9} & \cellcolor{red!30}\textbf{70.1} & \cellcolor{red!30}\textbf{63.9} \\
\textbf{19} &  &  & \cellcolor{blue!15}3 &  & \cellcolor{blue!15}5 & \cellcolor{blue!15}6 &  &  &  & \cellcolor{blue!15}10 & \cellcolor{blue!15}11 & \cellcolor{blue!15}12 & \cellcolor{blue!15}13 &  & \cellcolor{blue!15}15 & \cellcolor{blue!15}16 &  & \cellcolor{blue!15}18 &  &  & \cellcolor{orange!50}\textbf{49.2} & \cellcolor{red!30}\textbf{68.4} & \cellcolor{red!30}\textbf{72.8} & \cellcolor{red!30}\textbf{73.5} \\
\textbf{20} & \cellcolor{blue!15}1 & \cellcolor{blue!15}2 & \cellcolor{blue!15}3 &  & \cellcolor{blue!15}5 &  & \cellcolor{blue!15}7 &  & \cellcolor{blue!15}9 &  & \cellcolor{blue!15}11 &  & \cellcolor{blue!15}13 &  &  & \cellcolor{blue!15}16 &  &  & \cellcolor{blue!15}19 &  & \cellcolor{red!30}\textbf{66.6} & \cellcolor{red!30}\textbf{75.6} & \cellcolor{red!30}\textbf{70.2} & \cellcolor{orange!30}\textbf{38.2} \\
\bottomrule
\end{tabular}
\end{table}

\begin{table}[htbp]
\centering
\caption{\textbf{15-model combinations}. Matrix showing model presence across the 20 evaluated combinations.
Values under each task report the convergence$@n$ metric computed without a credible interval; each value is the mean of $10^5$ bootstrapped samples.
Model identifiers are listed in Table~\ref{app:tab:modelid}.}
\label{app:tab:comb15}
\small
\setlength{\tabcolsep}{3pt}
\begin{tabular}{c|cccccccccccccccccccc|rrrr}
\toprule
\textbf{Comb.} & \textbf{1} & \textbf{2} & \textbf{3} & \textbf{4} & \textbf{5} & \textbf{6} & \textbf{7} & \textbf{8} & \textbf{9} & \textbf{10} & \textbf{11} & \textbf{12} & \textbf{13} & \textbf{14} & \textbf{15} & \textbf{16} & \textbf{17} & \textbf{18} & \textbf{19} & \textbf{20} & \textbf{AIME'24} & \textbf{AIME'25} & \textbf{HMMT'25} & \textbf{BrUMO'25} \\
\midrule
\textbf{1} & \cellcolor{blue!15}1 & \cellcolor{blue!15}2 & \cellcolor{blue!15}3 & \cellcolor{blue!15}4 & \cellcolor{blue!15}5 & \cellcolor{blue!15}6 &  &  &  &  &  & \cellcolor{blue!15}12 & \cellcolor{blue!15}13 & \cellcolor{blue!15}14 & \cellcolor{blue!15}15 & \cellcolor{blue!15}16 & \cellcolor{blue!15}17 & \cellcolor{blue!15}18 & \cellcolor{blue!15}19 & \cellcolor{blue!15}20 & \cellcolor{red!30}\textbf{59.3} & \cellcolor{red!30}\textbf{76.1} & \cellcolor{red!30}\textbf{72.9} & \cellcolor{red!30}\textbf{77.1} \\
\textbf{2} & \cellcolor{blue!15}1 & \cellcolor{blue!15}2 & \cellcolor{blue!15}3 & \cellcolor{blue!15}4 & \cellcolor{blue!15}5 & \cellcolor{blue!15}6 &  &  &  & \cellcolor{blue!15}10 & \cellcolor{blue!15}11 &  &  & \cellcolor{blue!15}14 & \cellcolor{blue!15}15 & \cellcolor{blue!15}16 & \cellcolor{blue!15}17 & \cellcolor{blue!15}18 & \cellcolor{blue!15}19 & \cellcolor{blue!15}20 & \cellcolor{red!30}\textbf{51.3} & \cellcolor{red!30}\textbf{79.1} & \cellcolor{red!30}\textbf{71.7} & \cellcolor{red!30}\textbf{67.6} \\
\textbf{3} &  & \cellcolor{blue!15}2 &  &  & \cellcolor{blue!15}5 & \cellcolor{blue!15}6 & \cellcolor{blue!15}7 & \cellcolor{blue!15}8 & \cellcolor{blue!15}9 &  &  & \cellcolor{blue!15}12 & \cellcolor{blue!15}13 & \cellcolor{blue!15}14 & \cellcolor{blue!15}15 & \cellcolor{blue!15}16 & \cellcolor{blue!15}17 & \cellcolor{blue!15}18 & \cellcolor{blue!15}19 & \cellcolor{blue!15}20 & \cellcolor{red!30}\textbf{76.6} & \cellcolor{red!30}\textbf{79.0} & \cellcolor{red!30}\textbf{76.4} & \cellcolor{red!30}\textbf{77.1} \\
\textbf{4} &  & \cellcolor{blue!15}2 & \cellcolor{blue!15}3 &  &  & \cellcolor{blue!15}6 & \cellcolor{blue!15}7 & \cellcolor{blue!15}8 & \cellcolor{blue!15}9 &  & \cellcolor{blue!15}11 & \cellcolor{blue!15}12 & \cellcolor{blue!15}13 & \cellcolor{blue!15}14 & \cellcolor{blue!15}15 & \cellcolor{blue!15}16 & \cellcolor{blue!15}17 & \cellcolor{blue!15}18 &  & \cellcolor{blue!15}20 & \cellcolor{red!30}\textbf{76.7} & \cellcolor{red!30}\textbf{79.5} & \cellcolor{red!30}\textbf{76.5} & \cellcolor{red!30}\textbf{77.1} \\
\textbf{5} & \cellcolor{blue!15}1 &  & \cellcolor{blue!15}3 & \cellcolor{blue!15}4 & \cellcolor{blue!15}5 & \cellcolor{blue!15}6 & \cellcolor{blue!15}7 & \cellcolor{blue!15}8 & \cellcolor{blue!15}9 &  &  & \cellcolor{blue!15}12 & \cellcolor{blue!15}13 & \cellcolor{blue!15}14 & \cellcolor{blue!15}15 & \cellcolor{blue!15}16 &  &  & \cellcolor{blue!15}19 & \cellcolor{blue!15}20 & \cellcolor{red!30}\textbf{68.5} & \cellcolor{red!30}\textbf{79.6} & \cellcolor{red!30}\textbf{71.7} & \cellcolor{red!30}\textbf{73.9} \\
\textbf{6} & \cellcolor{blue!15}1 &  & \cellcolor{blue!15}3 & \cellcolor{blue!15}4 & \cellcolor{blue!15}5 & \cellcolor{blue!15}6 & \cellcolor{blue!15}7 & \cellcolor{blue!15}8 & \cellcolor{blue!15}9 &  & \cellcolor{blue!15}11 &  &  & \cellcolor{blue!15}14 & \cellcolor{blue!15}15 & \cellcolor{blue!15}16 & \cellcolor{blue!15}17 & \cellcolor{blue!15}18 & \cellcolor{blue!15}19 &  & \cellcolor{red!30}\textbf{73.1} & \cellcolor{red!30}\textbf{79.7} & \cellcolor{red!30}\textbf{72.7} & \cellcolor{red!30}\textbf{53.2} \\
\textbf{7} & \cellcolor{blue!15}1 & \cellcolor{blue!15}2 &  & \cellcolor{blue!15}4 &  &  & \cellcolor{blue!15}7 & \cellcolor{blue!15}8 & \cellcolor{blue!15}9 & \cellcolor{blue!15}10 & \cellcolor{blue!15}11 & \cellcolor{blue!15}12 & \cellcolor{blue!15}13 & \cellcolor{blue!15}14 &  &  & \cellcolor{blue!15}17 & \cellcolor{blue!15}18 & \cellcolor{blue!15}19 & \cellcolor{blue!15}20 & \cellcolor{red!30}\textbf{76.5} & \cellcolor{red!30}\textbf{79.9} & \cellcolor{red!30}\textbf{70.3} & \cellcolor{red!30}\textbf{69.0} \\
\textbf{8} & \cellcolor{blue!15}1 & \cellcolor{blue!15}2 &  & \cellcolor{blue!15}4 & \cellcolor{blue!15}5 &  & \cellcolor{blue!15}7 & \cellcolor{blue!15}8 & \cellcolor{blue!15}9 & \cellcolor{blue!15}10 & \cellcolor{blue!15}11 & \cellcolor{blue!15}12 & \cellcolor{blue!15}13 &  &  &  & \cellcolor{blue!15}17 & \cellcolor{blue!15}18 & \cellcolor{blue!15}19 & \cellcolor{blue!15}20 & \cellcolor{red!30}\textbf{76.5} & \cellcolor{red!30}\textbf{79.6} & \cellcolor{red!30}\textbf{70.3} & \cellcolor{red!30}\textbf{68.9} \\
\textbf{9} & \cellcolor{blue!15}1 & \cellcolor{blue!15}2 & \cellcolor{blue!15}3 &  & \cellcolor{blue!15}5 &  & \cellcolor{blue!15}7 & \cellcolor{blue!15}8 & \cellcolor{blue!15}9 & \cellcolor{blue!15}10 & \cellcolor{blue!15}11 & \cellcolor{blue!15}12 & \cellcolor{blue!15}13 &  &  &  & \cellcolor{blue!15}17 & \cellcolor{blue!15}18 & \cellcolor{blue!15}19 & \cellcolor{blue!15}20 & \cellcolor{red!30}\textbf{76.5} & \cellcolor{red!30}\textbf{79.3} & \cellcolor{red!30}\textbf{71.5} & \cellcolor{red!30}\textbf{69.9} \\
\textbf{10} & \cellcolor{blue!15}1 & \cellcolor{blue!15}2 & \cellcolor{blue!15}3 & \cellcolor{blue!15}4 &  &  & \cellcolor{blue!15}7 & \cellcolor{blue!15}8 & \cellcolor{blue!15}9 & \cellcolor{blue!15}10 & \cellcolor{blue!15}11 & \cellcolor{blue!15}12 & \cellcolor{blue!15}13 &  &  &  & \cellcolor{blue!15}17 & \cellcolor{blue!15}18 & \cellcolor{blue!15}19 & \cellcolor{blue!15}20 & \cellcolor{red!30}\textbf{76.5} & \cellcolor{red!30}\textbf{79.7} & \cellcolor{red!30}\textbf{71.5} & \cellcolor{red!30}\textbf{69.9} \\
\textbf{11} &  &  & \cellcolor{blue!15}3 & \cellcolor{blue!15}4 & \cellcolor{blue!15}5 & \cellcolor{blue!15}6 & \cellcolor{blue!15}7 & \cellcolor{blue!15}8 & \cellcolor{blue!15}9 &  & \cellcolor{blue!15}11 & \cellcolor{blue!15}12 & \cellcolor{blue!15}13 & \cellcolor{blue!15}14 & \cellcolor{blue!15}15 & \cellcolor{blue!15}16 & \cellcolor{blue!15}17 & \cellcolor{blue!15}18 &  &  & \cellcolor{red!30}\textbf{76.7} & \cellcolor{red!30}\textbf{79.9} & \cellcolor{red!30}\textbf{76.5} & \cellcolor{red!30}\textbf{75.8} \\
\textbf{12} &  &  &  &  & \cellcolor{blue!15}5 & \cellcolor{blue!15}6 & \cellcolor{blue!15}7 & \cellcolor{blue!15}8 & \cellcolor{blue!15}9 &  & \cellcolor{blue!15}11 & \cellcolor{blue!15}12 & \cellcolor{blue!15}13 & \cellcolor{blue!15}14 & \cellcolor{blue!15}15 & \cellcolor{blue!15}16 & \cellcolor{blue!15}17 & \cellcolor{blue!15}18 & \cellcolor{blue!15}19 & \cellcolor{blue!15}20 & \cellcolor{red!30}\textbf{76.6} & \cellcolor{red!30}\textbf{79.3} & \cellcolor{red!30}\textbf{76.4} & \cellcolor{red!30}\textbf{75.8} \\
\textbf{13} & \cellcolor{blue!15}1 & \cellcolor{blue!15}2 & \cellcolor{blue!15}3 & \cellcolor{blue!15}4 & \cellcolor{blue!15}5 & \cellcolor{blue!15}6 & \cellcolor{blue!15}7 & \cellcolor{blue!15}8 & \cellcolor{blue!15}9 & \cellcolor{blue!15}10 &  &  &  & \cellcolor{blue!15}14 & \cellcolor{blue!15}15 & \cellcolor{blue!15}16 &  &  & \cellcolor{blue!15}19 & \cellcolor{blue!15}20 & \cellcolor{red!30}\textbf{54.4} & \cellcolor{red!30}\textbf{78.7} & \cellcolor{red!30}\textbf{55.8} & \cellcolor{orange!50}\textbf{40.3} \\
\textbf{14} & \cellcolor{blue!15}1 & \cellcolor{blue!15}2 & \cellcolor{blue!15}3 & \cellcolor{blue!15}4 & \cellcolor{blue!15}5 & \cellcolor{blue!15}6 &  &  &  & \cellcolor{blue!15}10 & \cellcolor{blue!15}11 & \cellcolor{blue!15}12 & \cellcolor{blue!15}13 & \cellcolor{blue!15}14 & \cellcolor{blue!15}15 & \cellcolor{blue!15}16 &  &  & \cellcolor{blue!15}19 & \cellcolor{blue!15}20 & \cellcolor{red!30}\textbf{50.9} & \cellcolor{red!30}\textbf{79.2} & \cellcolor{red!30}\textbf{55.6} & \cellcolor{red!30}\textbf{73.9} \\
\textbf{15} & \cellcolor{blue!15}1 & \cellcolor{blue!15}2 & \cellcolor{blue!15}3 & \cellcolor{blue!15}4 & \cellcolor{blue!15}5 & \cellcolor{blue!15}6 &  &  &  &  & \cellcolor{blue!15}11 & \cellcolor{blue!15}12 & \cellcolor{blue!15}13 & \cellcolor{blue!15}14 & \cellcolor{blue!15}15 & \cellcolor{blue!15}16 & \cellcolor{blue!15}17 & \cellcolor{blue!15}18 & \cellcolor{blue!15}19 &  & \cellcolor{red!30}\textbf{59.4} & \cellcolor{red!30}\textbf{79.2} & \cellcolor{red!30}\textbf{73.0} & \cellcolor{red!30}\textbf{77.1} \\
\textbf{16} & \cellcolor{blue!15}1 & \cellcolor{blue!15}2 & \cellcolor{blue!15}3 & \cellcolor{blue!15}4 &  & \cellcolor{blue!15}6 &  &  &  & \cellcolor{blue!15}10 & \cellcolor{blue!15}11 & \cellcolor{blue!15}12 & \cellcolor{blue!15}13 &  & \cellcolor{blue!15}15 & \cellcolor{blue!15}16 & \cellcolor{blue!15}17 & \cellcolor{blue!15}18 & \cellcolor{blue!15}19 & \cellcolor{blue!15}20 & \cellcolor{red!30}\textbf{59.3} & \cellcolor{red!30}\textbf{77.0} & \cellcolor{red!30}\textbf{73.0} & \cellcolor{red!30}\textbf{77.1} \\
\textbf{17} & \cellcolor{blue!15}1 & \cellcolor{blue!15}2 & \cellcolor{blue!15}3 & \cellcolor{blue!15}4 &  & \cellcolor{blue!15}6 &  &  &  & \cellcolor{blue!15}10 & \cellcolor{blue!15}11 & \cellcolor{blue!15}12 & \cellcolor{blue!15}13 & \cellcolor{blue!15}14 &  & \cellcolor{blue!15}16 & \cellcolor{blue!15}17 & \cellcolor{blue!15}18 & \cellcolor{blue!15}19 & \cellcolor{blue!15}20 & \cellcolor{red!30}\textbf{52.8} & \cellcolor{red!30}\textbf{79.2} & \cellcolor{red!30}\textbf{57.0} & \cellcolor{red!30}\textbf{74.9} \\
\textbf{18} & \cellcolor{blue!15}1 & \cellcolor{blue!15}2 & \cellcolor{blue!15}3 & \cellcolor{blue!15}4 &  & \cellcolor{blue!15}6 &  &  &  & \cellcolor{blue!15}10 & \cellcolor{blue!15}11 & \cellcolor{blue!15}12 & \cellcolor{blue!15}13 & \cellcolor{blue!15}14 & \cellcolor{blue!15}15 &  & \cellcolor{blue!15}17 & \cellcolor{blue!15}18 & \cellcolor{blue!15}19 & \cellcolor{blue!15}20 & \cellcolor{red!30}\textbf{55.9} & \cellcolor{red!30}\textbf{79.2} & \cellcolor{red!30}\textbf{71.4} & \cellcolor{red!30}\textbf{75.3} \\
\textbf{19} & \cellcolor{blue!15}1 & \cellcolor{blue!15}2 & \cellcolor{blue!15}3 & \cellcolor{blue!15}4 & \cellcolor{blue!15}5 & \cellcolor{blue!15}6 & \cellcolor{blue!15}7 & \cellcolor{blue!15}8 & \cellcolor{blue!15}9 & \cellcolor{blue!15}10 &  &  &  & \cellcolor{blue!15}14 & \cellcolor{blue!15}15 & \cellcolor{blue!15}16 & \cellcolor{blue!15}17 & \cellcolor{blue!15}18 &  &  & \cellcolor{red!30}\textbf{73.1} & \cellcolor{red!30}\textbf{78.8} & \cellcolor{red!30}\textbf{72.7} & \cellcolor{red!30}\textbf{66.5} \\
\textbf{20} & \cellcolor{blue!15}1 &  & \cellcolor{blue!15}3 & \cellcolor{blue!15}4 & \cellcolor{blue!15}5 & \cellcolor{blue!15}6 & \cellcolor{blue!15}7 & \cellcolor{blue!15}8 & \cellcolor{blue!15}9 &  &  &  &  & \cellcolor{blue!15}14 & \cellcolor{blue!15}15 & \cellcolor{blue!15}16 & \cellcolor{blue!15}17 & \cellcolor{blue!15}18 & \cellcolor{blue!15}19 & \cellcolor{blue!15}20 & \cellcolor{red!30}\textbf{73.1} & \cellcolor{red!30}\textbf{78.7} & \cellcolor{red!30}\textbf{72.7} & \cellcolor{red!30}\textbf{53.2} \\
\bottomrule
\end{tabular}
\end{table}
\end{document}